\documentclass[]{article}

\usepackage[nonatbib, preprint]{neurips_data_2024}
\usepackage{graphicx}

\usepackage{math}
\usepackage[utf8]{inputenc} 
\usepackage[T1]{fontenc}    
\usepackage{hyperref}       
\usepackage{url}            
\usepackage{booktabs}       
\usepackage{amsfonts}       
\usepackage{nicefrac}       
\usepackage{microtype}      
\usepackage{wrapfig}        
\usepackage{xcolor, colortbl}         
\definecolor{mypurple}{rgb}{0.6, 0.59, 0.91}
\definecolor{mygreen}{rgb}{0.694, 0.816, 0.584} 
\definecolor{myorange}{rgb}{0.882, 0.678, 0.522} 
\definecolor{mydarkpurple}{rgb}{0.443, 0.1, 0.6}
\definecolor{mydarkgreen}{rgb}{0.369, 0.506, 0.247}
\definecolor{mydarkorange}{rgb}{0.722, 0.376, 0.161}

\usepackage{scalerel}       
\usepackage{marvosym}     
\usepackage{enumitem}
\usepackage{wrapfig}
\usepackage{algorithm}
\usepackage{algorithmic}

\usepackage{eqparbox}
\usepackage{dirtree}
\usepackage{todonotes}
\usepackage{subcaption}

\usepackage{multirow, multicol}

\def\emoji{\scaleobj{0.038}{\includegraphics{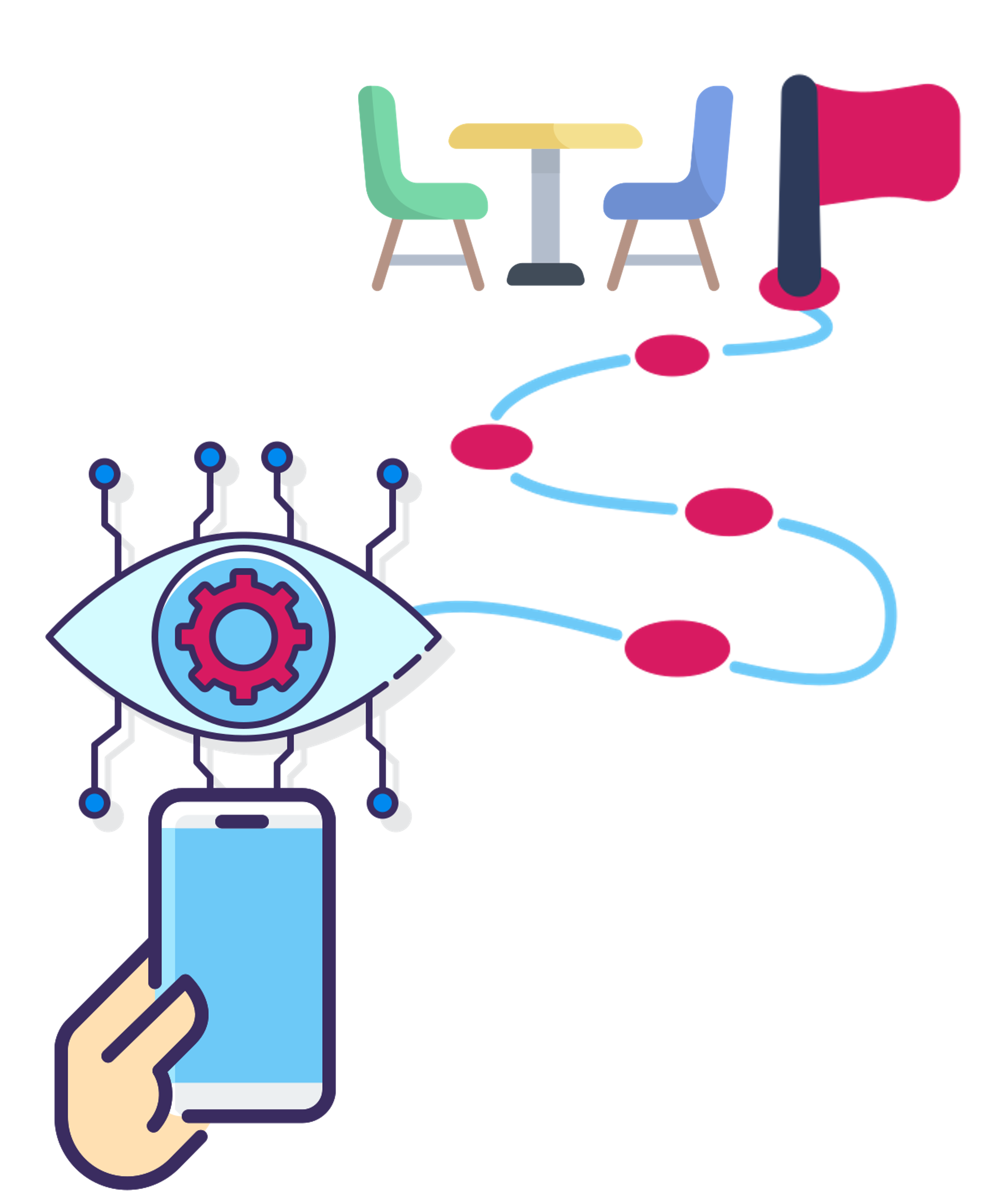}}}

\usepackage[most]{tcolorbox}
\newtcolorbox{applebox}[1]{
    enhanced,
    colback=white,
    colframe=gray!20,
    fonttitle=\bfseries,
    coltitle=black,
    colbacktitle=white,
    left=6pt,
    right=6pt,
    top=2pt,
    bottom=2pt,
    boxsep=5pt,
    boxrule=1.5pt,
    arc=2mm,
    title=#1, %
    borderline={0pt}{0pt}{white},
    fonttitle=\small\bfseries,
    attach boxed title to top left={yshift=-2mm, xshift=3mm},
    boxed title style={boxrule=0pt, colback=gray!30, frame hidden, left=2pt, right=2pt},
}

\title{
    \begin{tabular}{c l}%
        \multirow{2}{*}{\protect\emoji{}} & \textbf{\texttt{NaVIP}}: An Image-Centric Indoor \underline{Navi}gation\\%
        & ~~~~~Solution for \underline{V}isually \underline{I}mpaired \underline{P}eople \\%
    \end{tabular}%
}
\author{
    {\normalfont\begin{tabular}{c c c c}
        \textbf{Jun Yu}$^{\,1}$ & \textbf{Yifan Zhang}$^{\,1}$ & \textbf{Badrinadh Aila}$^{\,1}$ & \textbf{Vinod Namboodiri}$^{\,1, 2}$ \\
        \multicolumn{4}{c}{\texttt{\{juy220, yiz521, baa223, vin423\}@lehigh.edu}}\\
    \end{tabular}}\\
    $^1$ \normalfont Department of Computer Science and Engineering, Lehigh University.\quad{} \\ 
    $^2$ \normalfont Department of Community and Population Health, Lehigh University.\quad{} \\
    \url{https://accesslab180.github.io/navip.github.io}
}

\begin{document}
\setlength{\aboverulesep}{0pt}
\setlength{\belowrulesep}{0pt}
\setlength{\extrarowheight}{.75ex}
\maketitle
\vspace{-15pt}
\begin{figure}[h]
    \centering
    \includegraphics[width=0.9\linewidth]{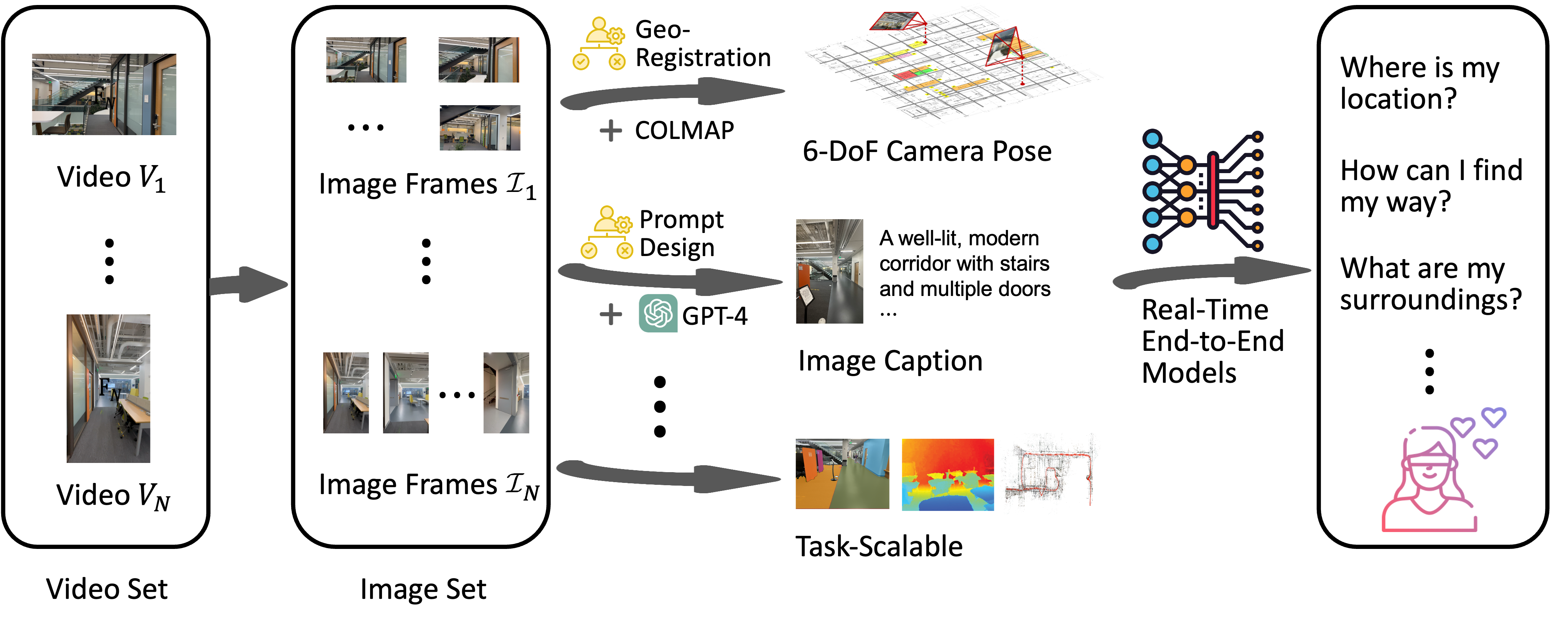}
    \vspace{-5pt}
    \caption{Illustration of pipelines for purely image-based indoor navigation. We collect videos and extract image frames as data sources. Each image is annotated with: 1) 6-DoF camera poses aligned with the floor plan, 2) indoor points-of-interest (PoIs), and 3) visual descriptions that assist visually impaired people (VIPs) in understanding their surroundings. We highlight the task scalability of this solution, facilitated by its end-to-end training and inference using simple image forward pass.}
    \label{fig:framework}
\end{figure}
\begin{abstract}
Indoor navigation is challenging due to the absence of satellite positioning. This challenge is manifold greater for Visually Impaired People (VIPs) who lack the ability to get information from wayfinding signage. Other sensor signals (e.g., Bluetooth and LiDAR) can be used to create turn-by-turn navigation solutions with position updates for users. Unfortunately, these solutions require tags to be installed all around the environment or the use of fairly expensive hardware. Moreover, these solutions require a high degree of manual involvement that raises costs, thus hampering scalability. We propose an image dataset and associated image-centric solution called \textbf{\texttt{NaVIP}} towards visual intelligence that is infrastructure-free and task-scalable, and can assist VIPs in understanding their surroundings. Specifically, we start by curating large-scale phone camera data in a four-floor research building, with $300$K images, to lay the foundation for creating an image-centric indoor navigation and exploration solution for inclusiveness. Every image is labelled with precise 6DoF camera poses, details of indoor PoIs, and descriptive captions to assist VIPs. We benchmark on two main aspects: 1) positioning system and 2) exploration support, prioritizing training scalability and real-time inference, to validate the prospect of image-based solution towards indoor navigation. The dataset, code, and model checkpoints are made publicly available at \href{https://github.com/junfish/VIP_Navi}{https://github.com/junfish/VIP\_Navi}. 
\end{abstract}

\section{Introduction}

\label{sec:intro}
Indoor navigation~\cite{huang2010survey, kunhoth2020indoor, volkovich2024indoor} remains a complex challenge due to the unavailability of satellite signals, such as GPS, for indoor localization. The dynamic nature of indoor environment scenes further complicates the positioning. While Wi-Fi-based indoor positioning systems (WIPS)~\cite{cypriani2009open, wandell2023cost} offer some promise by utilizing signal strength measurements from multiple access points (APs), their effectiveness can vary. To improve positioning accuracy and stability, deploying bluetooth low energy (BLE) beacons~\cite{zhuang2016smartphone, canton2017bluetooth, szyc2023bluetooth} within buildings has been explored. These beacons emit audio signals that, when received by mobile devices, can aid in determining the user locations. However, this solution necessitates extensive tag hardware deployment and ongoing maintenance and sometimes interferes with other signals, often making it impractical due to resistance from building managers. Similar technologies, such as radio frequency identification (RFID)~\cite{kim2020review} and ultra-wideband (UWB)~\cite{che2023indoor}, have been refined to mitigate the inconvenience and further enhance accuracy; however, these also come with the same disadvantages of requiring tag installations and maintenance. Other alternatives like magnetic and acoustic positioning offer infrastructure-free solutions but lack robustness against the spatial and temporal variations inherent in indoor settings. Dead reckoning~\cite{geng2021smartphone, yan2023deep} and SLAM~\cite{teed2021droid, macario2022comprehensive} encounter challenges with error accumulation in positioning during navigation and also demonstrate limited robustness to environmental changes. Current marker-based methods offer cost-effective solutions but struggle with the accuracy of positioning when tracking visual landmarks (e.g., fiducial markers~\cite{fiala2009designing, kalaitzakis2021fiducial}).

All the technologies above have their place in indoor navigation solutions, often complementing each other within hybrid systems~\cite{de2017study, dong2018vinav, feng2023adaptive} that aim to leverage the strengths of each approach while mitigating their individual disadvantages. While the above mentioned options may already serve sighted individuals with a strong sense of direction, the challenges of indoor navigation is much greater for visually impaired people (VIPs)~\cite{manjari2020survey, el2021systematic, kuriakose2022tools} who cannot independently explore unfamiliar indoor environments due to lack of access to existing wayfinding signage and the inability to create a mental map of complex layouts. For VIPs, challenges of independent indoor navigation is more than just an inconvenience; it can mean the difference between venturing into unknown indoor spaces or just avoiding them altogether. In summary, current indoor navigation technologies struggle to provide effective solutions, a necessity for VIPs, for dynamic environments where (1) accurate and real-time positioning, (2) scalability across varying scene sizes, and (3) understanding of the surroundings, are all required simultaneously. To address these concerns, we are pioneering the application of visual intelligence towards indoor navigation. However, our initial and most significant challenge is the absence of a large real-world dataset derived directly from indoor navigation scenarios. These scenarios involve rapidly changing scenes as users move, and continuously evolving environments over days. Therefore, we curated a image-based indoor navigation dataset to facilitate the straightforward implementation of feasible applications. To further support this, we also ensured that the dataset collection process is scalable and adaptable to changes in building layouts and scene sizes.

The key task in such a purely image-based navigation system is positioning, simply put as determining your camera location from a single image. Contemporary methods for this camera localization task exhibit different trade-offs among hardware resources, prediction accuracy, inference time, and algorithm robustness. For example, state-of-the-art (SOTA) accuracy in camera localization is achieved by constructing a 3D model of the scene using either sparse feature points~\cite{schonberger2016structure, sarlin2019coarse} or dense reconstruction~\cite{torii201524, taira2018inloc}. The camera pose is then estimated through geometric calculations~\cite{fischler1981random, lepetit2009ep} or 2D-3D matching~\cite{sattler2016efficient}. However, these methods typically require significant memory overhead and have slow processing speeds, making them impractical for integration into a navigation system. On the other hand, absolute pose regression (APR)~\cite{kendall2015posenet, musallam2022leveraging, chen2023refinement} can determine the camera pose with a single forward propagation in deep neural networks using query images. This line of research offers significant advantages in inference speed and can be easily deployed in thin client applications due to its minimal memory footprint. It achieves this while only incurring an acceptable level of accuracy loss in positioning, which is suitable for real-time navigation purposes. Additionally, the APR decision process is aligned with other well-established computer vision (CV) tasks in the era of deep learning, such as image recognition~\cite{krizhevsky2012imagenet, He_2016_CVPR}, semantic segmentation~\cite{long2015fully, kirillov2023segment}, and depth estimation~\cite{eigen2014depth, yang2024depth}. Considering task scalability is crucial for the development of an inclusive and accessible indoor navigation system. In this paper, we not only benchmark APR but also demonstrate how image captioning can assist VIPs in navigating indoor environments independently for their own needs~\cite{plikynas2020indoor, jain2023want}. Without the burden of human-crafted annotations, this is achieved by leveraging recent advancements in pre-trained foundation models (PFMs)~\cite{bommasani2021opportunities, zhou2023comprehensive}, such as Segment Anything Model (SAM)~\cite{kirillov2023segment} and GPT-4~\cite{achiam2023gpt}.

The main contribution of this paper is the creation of an image dataset for an indoor building and the demonstration of our image-centric solution using this dataset to assist VIPs in navigating and exploring dynamically changing indoor environments. Using only  a commodity-off-the-shelf (COTS)  phone camera and minimizing human involvement in ground-truth annotations, we demonstrate our solution to be both practical and highly adaptable. Our benchmarks utilize deep representation learning for its robustness to varying scenarios, scalability across different scene sizes, and, most importantly, its simple forward pass inference in an end-to-end manner, responding to query images in just a few milliseconds. In \S\ref{subsec:camera_relocalization}, we benchmark APR methods (e.g., representative PoseNet~\cite{kendall2015posenet}) on \textbf{\texttt{NaVIP}}, demonstrating their ability to pinpoint phone-captured images on a floor plan with sub-meter accuracy when applied to actual building layouts. \S\ref{subsec:image_captioning} explores the potential of image captioning techniques to support the exploration needs of VIPs, illustrating the broader applicability of this image-centric approach to addressing the diverse needs of individuals with disabilities. With the release of \textbf{\texttt{NaVIP}}, we hope to spark further interest within the academic community to develop image-centric solutions for challenges such as indoor navigation and exploration, thus realizing some important human and societal benefits of AI.

\section{Related Work}
\label{sec:related}
\subsection{Indoor Navigation} 
Current full-fledged \textit{indoor navigation systems}~\cite{huang2010survey, kunhoth2020indoor, el2021indoor, khan2022recent} employ a set of technologies including WiFi fingerprints, BLE beacons, magnetic fields, and IMUs, either individually or in combination. These sensor technologies are specifically engineered to address the most challenging aspect of indoor navigation---accurate and robust positioning. 
\cite{shi2018design} proposed a hybrid multi-sensor fusion system for indoor localization using WiFi and LiDAR.
ViNav~\cite{dong2018vinav}, named vision-based navigation system, relies on a combination of WiFi fingerprints, dead reckoning, and SfM 3D point clouds to localize user positions.
ASSIST~\cite{nair2018assist} is a personalized system using multimodal sensors and high-level semantic information, with efficacy tested on a blind and visually impaired (BVI) group.
\cite{lin2023wi} proposed data-driven methods similar to ours but trained their localization model using annotated Wi-Fi observations instead of images. \cite{albraheem2023hybrid} explored the combination of radio and visible light communication-based positioning technologies.

Recent advancements in \textit{vision-and-language navigation (VLN)}~\cite{anderson2018vision, wu2023vision, gu2022vision} has facilitated the translation of natural language instructions into practical actions for embodied agents, leveraging their visual perceptions. This development is instrumental in assisting users who can command such agents for indoor applications, such as search-and-rescue missions~\cite{tadokoro2009rescue}. However, most VLN methods~\cite{anderson2019chasing, qi2020reverie, lin2022multimodal, li2023kerm, huo2023geovln, liu2023bird, lin2023learning, kuo2023structure, qiao2023vln} are largely limited to controlled, simulated 3D environments~\cite{koh2021pathdreamer}, which significantly narrows the applicability in real-world settings. Interactive VLN~\cite{Chi_Shen_Eric_Kim_Hakkani-tur_2020, burns2022dataset, krantz2023iterative}, synchronizing human feedback to adapt to new environment, can explore unknown command feasibility, but struggles in scenarios where the oracle is disabled. While trajectory-instruction generation~\cite{fried2018speaker, agarwal2019visual, tan2019learning, fu2020counterfactual, gu2022vision, dou2022foam, wang2022counterfactual, li2023improving, wang2023scaling} that synthesizes new language instructions can alleviate data scarsity, their efficacy still lags behind precious oracle instructions~\cite{zhao2021evaluation}. 

\subsection{Visual Localization}
The problem of visual localization, aka camera pose estimation, is fundamental in many CV applications except navigation, such as augmented reality, SLAM system, and autonomous driving.

\textit{Indirect localization} casts the camera pose as a query frame retrieval problem~\cite{sattler2012image, radenovic2016cnn, sarlin2019coarse}. Traditional methods~\cite{galvez2011real} are dependent on the quality of feature detection and matching, and often require manual tuning to effectively retrieve the most similar images stored in a database or to interpolate the camera pose from top retrieved images. In contrast, deep learning approaches~\cite{sarlin2019coarse, germain2019sparse, li2020hierarchical, wang2024hscnet++} utilize hierarchical pipelines that establish 2D-3D correspondences more efficiently and robustly. Despite achieving SOTA accuracy, these methods, which often incorporate PnP and RANSAC processes, tend to be slower by an order of magnitude ranging from 10 to 100 times compared to APR methods. Furthermore, although relative pose regression (RPR)~\cite{laskar2017camera, balntas2018relocnet, ding2019camnet, turkoglu2021visual, chen2021wide, sinha2023sparsepose} also adopts camera pose regression, the inherent retrieval phase continues to hinder inference efficiency.

\textit{Direct localization} can instantly relocalize the camera pose from the query image~\cite{shotton2013scene, guzman2014multi, valentin2015exploiting, kendall2015posenet, kendall2017geometric, shavit2021learning}, thus requiring a smaller memory footprint compared to indirect methods. Feature-based matching methods~\cite{lowe2004distinctive, arth2009wide, irschara2009structure, li2010location, sarlin2021back} always perform global localization by matching the feature points of query images with a 3D point cloud reconstruction. APR methods~\cite{kendall2015posenet, kendall2016modelling, kendall2017geometric, walch2017image, melekhov2017image, cai2019hybrid, chen2021direct, brahmbhatt2018geometry, shavit2021learning, shavit2022camera, chen2022dfnet, chen2023refinement} are capable of directly regressing the camera pose from a single image input in milliseconds, with minimal loss of accuracy. Although these methods have historically struggled to generalize to new camera poses~\cite{sattler2019understanding}, recent advancements in novel view synthesis (NVS)~\cite{mildenhall2021nerf, martin2021nerf, chen2021direct, liu2023nerf, chen2023refinement} could alleviate this burden via synthesizing new images from random viewpoints as data augmentation. 
Recent \textit{floorplan localization}~\cite{karkus2018particle, chen2024f} proposed to directly predict and localize with a single image in a floor plan.

\subsection{CV and Beyond}
CV has achieved the most predominant achievement among generic applications over the past few deep learning years, such as object detection~\cite{zhao2019object, liu2020deep, 10028728}, pose estimation~\cite{xiang2017posecnn, wang2019densefusion, hu2020single}, semantic segmentation~\cite{long2015fully, kirillov2023segment}, depth estimation~\cite{eigen2014depth, yang2024depth}, etc. Recent PFMs~\cite{bommasani2021opportunities, zhou2023comprehensive} provide a  powerful base that can achieve zero-shot performance. For example, SAM~\cite{kirillov2023segment} enables zero-shot generalization in the detection and segmentation of unfamiliar subjects. Depth Anything~\cite{yang2024depth} leverages unlabeled data to achieve SOTA performance in monocular depth estimation across previously unseen datasets. More recently, multimodal large language models (MLLMs), such as GPT-4(o)~\cite{achiam2023gpt} and Gemini~\cite{team2023gemini}, have been instrumental in providing accurate image descriptions and facilitating various user-adaptive tasks including image-to-text and image-to-audio conversions. Although there have been significant advances in computer vision, research dedicated to enhancing assistive and accessibility technologies remains limited~\cite{gonzalez2024investigating, shukurov2024improve}. For VIPs, the project VizWiz\footnote{https://vizwiz.org/}~\cite{bigham2010vizwiz, burton2012crowdsourcing, gurari2018vizwiz, chen2022grounding, reynolds2024salient} pioneers the first Visual Question Answering (VQA) datasets that originate directly from VIPs and are tailored to benefit them. Recently, applications such as Be My Eyes and Seeing AI utilize a generative AI-powered virtual volunteer\footnote{https://openai.com/index/be-my-eyes/} to interpret and describe images for their BVI users. Our research, however, distinguishes itself by focusing on customizing distilled model and providing tailored navigation assistance, thereby enhancing the support offered to VIPs.

\section{NaVIP}
\label{sec:navip}
\subsection{Data collection and annotation}
This section presents the methods used for collecting and annotating our large-scale image dataset. To make it easily extend this process to other buildings, we prioritize reducing human labour. Within our workflow, human involvement is confined to recording videos, pinpointing 5--10 images from each video to the floor plan, and designing prompts for GPT-4. We also release the data preprocessing and annotating code to further alleviate the burdens associated with data collection and annotation.
\subsubsection{Collection} 
\label{subsubsec:collection}
To facilitate the adaptation of this data-driven, vision-based indoor navigation solution across various buildings in the future, data collection is streamlined to minimize human effort and allow for automation through a basic robot that does not require specialized design. We hereby consider the video recording by smartphone built-in cameras and subsequent extraction of image frames. In this study, we collected approximately 400 videos, each lasting 2--4 minutes, within the Health, Science, and Technology (HST) Building\footnote{HST is a hub designed to encourage collaboration among faculty and students across disciplines, which is the largest building Lehigh has ever built. For more information, please click \href{https://www2.lehigh.edu/news/new-health-science-and-technology-building-a-hub-for-interdisciplinary-research}{this link}.}. By setting the frame extraction interval between 0.2 and 0.3 seconds, we obtained a big dataset comprising around $300$K images. The simplicity of our data collection allows for its easy way of expansion, either through additional video recordings or by narrowing the time intervals for frame extraction.
    
To enhance the robustness of algorithms developed from this dataset, videos were captured using four distinct smartphone models, in both landscape and portrait orientations. These recordings were made in two different holding ways: by human hand and using a smartphone gimbal stabilizer. To comprehensively capture the variability of indoor environments, recordings spanned various times of the day--sunrise, morning, noon, afternoon, sunset, and evening--and were conducted from December 2023 through June 2024. Videos recorded after April are designated as testing set to ensure that models developed using this dataset exhibit generalizability. For additional statistics and instructions regarding the data, please refer to \textbf{Appendix A}. We acknowledge that despite our efforts to align this dataset closely with real-world scenarios, the dataset biases still exist and may be captured by models~\cite{torralba2011unbiased, liu2024decade}. For example, our dataset lacks the coverage of the fall season.

\begin{wrapfigure}[16]{r}{0.58\textwidth}
\vspace{-21pt}
\begin{minipage}{0.58\textwidth}
\begin{algorithm}[H]
\scriptsize
\caption{~Camera Pose (6DoF) Annotation}
\label{alg:algorithm_6DoF}
\begin{algorithmic}[1]
\REQUIRE Video set $\mathcal{V} = \{V_1, \cdots, V_N\}$ by mobile cameras.
\ENSURE Geo-registered ground-truth of camera poses (positions and orientations) on the floor plan for extracted video frames.
\STATE Initialize frame interval $\Delta f$. \COMMENT{$\Delta f\approx16$.}
\FOR{$n = 1, \cdots, N$} 
    \STATE {
        Sample the image sequences $\mathcal{I}_n =\{I_1, \cdots, I_{M_n}\}$ from video $V_n$ every $\Delta f$ frames.
        }
    \REPEAT
        \STATE Run SfM algorithm on image set $\mathcal{I}$. \COMMENT{Use COLMAP.}
        \IF {Point clouds align} 
            \STATE Obtain camera pose for image sequences $\mathcal{I}$. 
        \ELSE 
            \STATE Sample more images $\mathcal{I}'$ at breaking points. \COMMENT{$\Delta f' = 1/3\Delta f$}
            \STATE Data augmentations: $\mathcal{I} = \mathcal{I} \cup \mathcal{I}'$.
        \ENDIF
    \UNTIL Point clouds align
    \STATE Geo-register the camera pose on the floor plan. \COMMENT{See Figure~\ref{fig:geo_registrationb}.}
\ENDFOR
\end{algorithmic}
\end{algorithm}
\end{minipage}
\end{wrapfigure}

\subsubsection{Camera Pose}
\label{subsubsec:camera_pose}
There are multiple ways to annotate a camera with its accurate shooting position. Most methods require professional equipment~\cite{shotton2013scene, walch2017image, taira2018inloc}, e.g., the NavVis VLX 3D scanner backpack, or involve complex optimization pipelines that are either closed-source or difficult to replicate~\cite{valentin2016learning, sarlin2022lamar, huang2023360loc}.
We conclude the summary of camera relocalization datasets, both indoors and outdoors, in Table~\ref{tab:pose-dataset}. 
It is noteworthy that only the Cambridge~\cite{kendall2015posenet} relied on ubiquitous device and open-source software to obtain the 6-DoF ground-truth of camera pose and achieved a 10 dm error level outdoors. Based on this, we emphasize the importance of low-cost and replicable methods for obtaining ground-truth data for training, especially when our users include low-income groups with health disparities. We employ the active open-source project COLMAP\footnote{COLMAP is available at \url{https://github.com/colmap/colmap} and is licensed under the BSD-3-Clause.} to reconstruct the 3D scene and determine the camera pose for each image in our collections. The algorithm for 6DoF annotation process is presented in Algorithm~\ref{alg:algorithm_6DoF}. 

\begin{figure}[t]
    \centering
    \includegraphics[width=0.88\linewidth, height=0.38\linewidth]{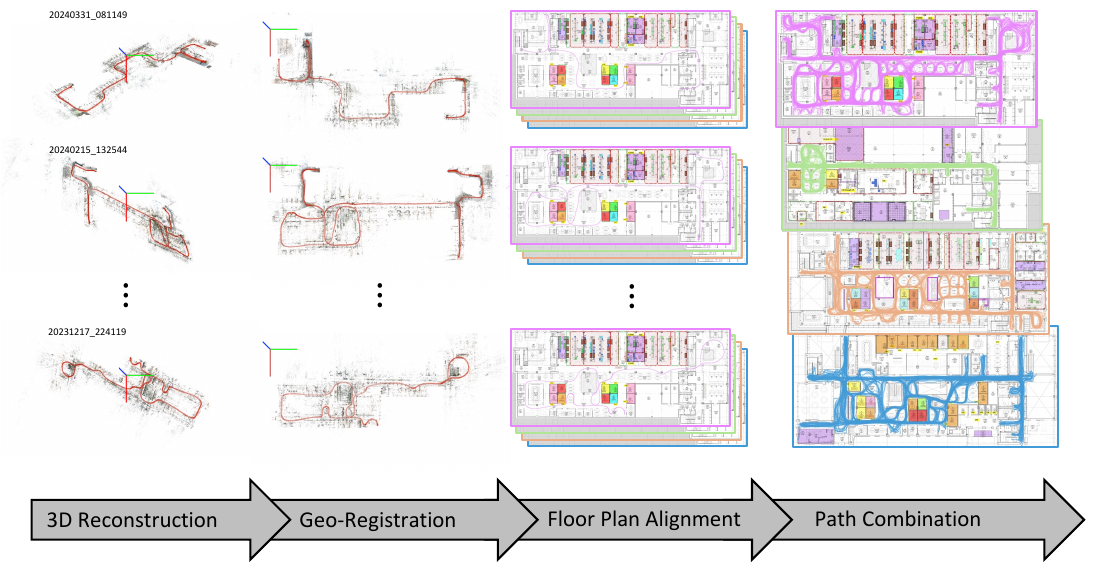}
    \caption{This illustration outlines the steps for converting camera poses from initial coordinates to a unified floor plan world coordinate system: 1) Independent 3D scene construction for images from each video using COLMAP, executed in parallel; 2) Pinpointing several anchor images from each video to the floor plan and geo-registering camera poses using these anchors; 3) Aligning and validating the entire path onto the floor plan; 4) Combining all images from different paths for unified training purposes.}
    \label{fig:geo_registrationb}
\end{figure}

As discussed in \S~\ref{subsubsec:collection}, to capture the dynamic variations in scene evolution over time, we independently collected approximately 400 videos. These videos were recorded by various collectors using different gestures and devices at diverse times. Consequently, the world coordinates for each video are isolated within the COLMAP 3D reconstructions; they are not geo-registered and cannot be integrated into a unified world coordinate system that would align with a floor plan for navigation purposes. We introduced minimal human labor to accurately pinpoint 5--10 images for each video to the floor plan. This accuracy is achieved as the 3D points of the scene and each image are visualized using the COLMAP GUI. Additionally, we utilized embedded geo-registration function to transform the world coordinates. Figure~\ref{fig:geo_registrationb} shows the process of geo-registring all camera poses from different paths into a unified world coordinate system, ensuring alignment with the floor plan.

\begin{table*}[t]
    \centering
    \scriptsize
    \caption{Summary of camera relocalization datasets. This table exclusively compares datasets that are publicly available.}
    \label{tab:pose-dataset}
    \scalebox{0.75}{
    \begin{tabular}{llllllll}
    \toprule
        Dataset & Environment & Device
        & \# Train / Test & Resolution & Scope  & Ground Truth Tool & Error Level  \\
    \toprule
        Dubrovnik 6K~\cite{li2012worldwide} & Outdoor & -- 
        & $6$K / $0.8$K & --- & $1.5\times 1.5$ km$^2$ & SIFT Matching & $\sim 10$ m \\
    \midrule
        7-Scenes~\cite{shotton2013scene} & Indoor & Kinect depth camera 
        & $16$K / $17$K & $640\times 480$ & $4\times 3$ m$^2$ & KinectFusion~\cite{newcombe2011kinectfusion} & $<10$ cm\\
    \midrule
        Cambridge~\cite{kendall2015posenet} & Outdoor & Smartphone camera 
        & $8.4$K / $4.8$K & $1920\times 1080$ & $500\times 100$ m$^2$ & MVS~\cite{furukawa2010towards} & $<10$ dm \\
    \midrule
        12-Scenes~\cite{valentin2016learning} & Indoor & Structure.io depth sensor with iPad 
        & $240$K / $6.7$K & $1296\times968$ & 80 m$^3$ & VoxelHashing~\cite{niessner2013real} & $\sim10$ cm\\
    \midrule
        TUM-LSI~\cite{walch2017image} & Indoor & NavVis M3 trolley & $875$/$220$ & $4592 \times 3448$ & $5575$ m$^2$ & SLAM & $<10$ cm \\
    \midrule
        InLoc~\cite{taira2018inloc} & Indoor & Faro 3D laser scanner 
        & $10$K / $0.4$K & $1600\times 1,200$ & $186$ m$^2$ & LiDAR + Manual & $\sim 10$ dm \\
    \midrule
        \multirow{2}{*}{LaMAR~\cite{sarlin2022lamar}} & \multirow{1.2}{*}{Indoor \&} & \multirow{1.2}{*}{Microsoft Hololens 2, Smartphone,}
        & \multirow{2}{*}{---} & \multirow{2}{*}{$640\times 480$} & \multirow{2}{*}{45,000 m$^2$} & \multirow{1.2}{*}{LiDAR + SfM +} & \multirow{2}{*}{$<10$ cm}\\
        & \multirow{-1.15}{*}{Outdoor} & \multirow{-1.15}{*}{iPad, NavVis M6 or VLX backpack} & & & & \multirow{-1.15}{*}{VIO~\cite{koide2023general}} & \\
    \midrule
        360Loc~\cite{huang2023360loc} & Indoor \& Outdoor & Velodyne lidar with $360^\circ$ camera 
        & $9.3$K / -- & $6144\times 3072$ & $105\times 70$ m$^2$ & LiDAR + VIO & $<10$ cm \\
    \midrule
        \textbf{\texttt{NaVIP}} (Ours) & Indoor & Smartphone camera 
        & $212$K / $88$K & $3840\times 2160$ & $40 \times 90$ m$^2$ & COLMAP~\cite{schonberger2016structure} & $<10$ cm \\
        \bottomrule
    \end{tabular}
    }
\end{table*}

\subsection{PoIs}
\begin{wrapfigure}[13]{r}{0.64\textwidth}
    \centering
    \includegraphics[width=0.95\linewidth, height=0.46\linewidth]{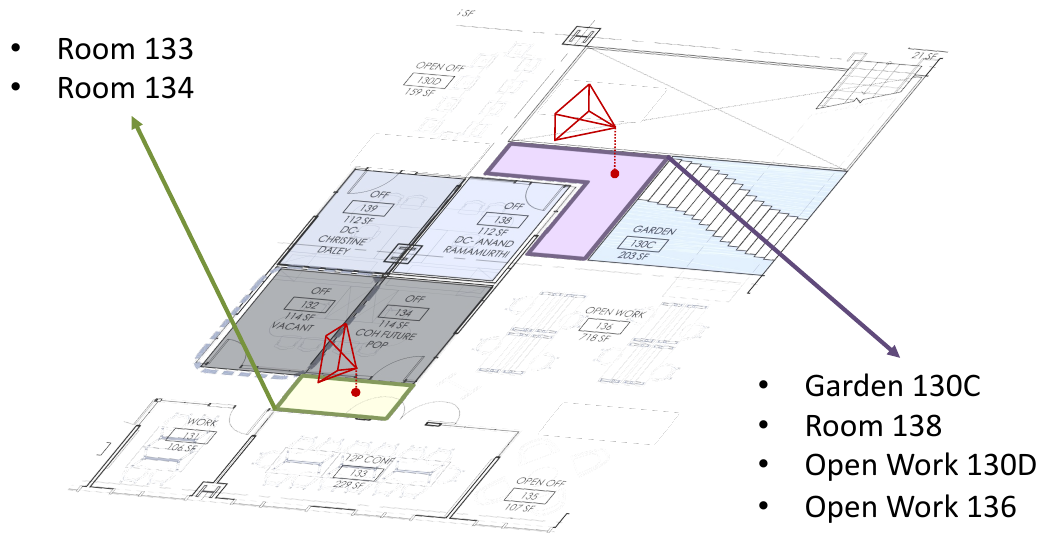}
    \caption{Floor plan dependent PoIs can be reported simultaneously as the camera pose is determined.}
    \label{fig:PoI}
\end{wrapfigure}
By aligning all images to the floor plan, annotating nearby points-of-interest (PoIs) becomes straightforward. As illustrated in Figure \ref{fig:PoI}, we can mark pixel-level PoI labels for each public area. Predicted images at specific points can then provide feedback on these PoIs to users, depending on their camera positions and orientations. We release the detailed PoIs for each pixel of the HST floor plans to support the development of applications.

\subsection{Captions for VIPs}
We leverage recent MLLMs, e.g., OpenAI GPT-4 and Google Gemini, to generate tailored image descriptions that meet the specific needs of VIPs. Our prompt design considers both the capabilities of these models and the feedback from VIP volunteers. For more details on our explorations of prompting MLLMs and output examples of image descriptions, please refer to \textbf{Appendix B}. We offer three types of image descriptions for each image:
\begin{itemize}[itemsep=-1pt, leftmargin=5mm]
\item A concise description suitable for general use.
\item A detailed description specifically designed for individuals with vision impairments (including those with low vision or acquired blindness).
\item Another detailed description tailored for individuals who have been blind since birth.
\end{itemize}

\section{Benchmarking}
\label{sec:benchmark}
To evaluate the generalization ability across ever-changing indoor environments, we utilize images captured prior to April 15th as training dataset and images captured after May 1st as testing dataset throughout the benchmarking process.
\subsection{Camera Relocalization}
\label{subsec:camera_relocalization}
\textbf{Preliminary.} For clarity and consistency, we represent the camera pose in 3D space by a 2-element tuple $[\xbold, \qbold]$ according to \cite{kendall2015posenet}, where $\xbold\in\mathbb{R}^3$ defines the position of camera center in 3D Cartesian coordinates and $\qbold\in\mathbb{R}^4$ is the unit quaternion encoding its orientation. Other variations focus on orientation representations, with BranchPoseNet~\cite{wu2017delving} employing an Euler angle representation and MapNet~\cite{mapnet2018} using a logarithm of the unit quaternion.

Our purpose is to directly regress the camera pose $[\hat{\xbold}, \hat{\qbold}]$ from a single monocular image $I$ using the trained function $f$. The standard objective loss function is defined as follows:
\begin{equation}
    \Lmcal(I) = {\|\hat{\xbold} - \xbold\|}_2 + \beta \cdot {\left\|\hat{\qbold} - \frac{\qbold}{{\|\qbold\|}_2}\right\|}_2,
    \label{eq:posenet_loss}
\end{equation}
where $[\xbold, \qbold] = f(I)$ represents the values predicted by our models. Notably, $\beta$ is a hyperparameter introduced to balance the learning scales between position and orientation. Learnable PoseNet~\cite{kendall2017geometric} captures homoscedastic uncertainty~\cite{kendall2018multi} between two tasks and omit $\beta$ as
\begin{equation}
    \Lmcal(I) = e^{-s_x} \cdot {\|\hat{\xbold} - \xbold\|}_2 + e^{-s_q} \cdot {\left\|\hat{\qbold} - \frac{\qbold}{{\|\qbold\|}_2}\right\|}_2 + s_x + s_q,
    \label{eq:learnable_posenet_loss}
\end{equation}
where $x_x$ and $s_q$ are both learnable and only an approximate initial guess is required.
\begin{table}[t]
    \centering
    \caption{Mean and median errors$\downarrow$ of models across four floors (Basement, Lower Level, Level 1, and Level 2) in our dataset. Benchmarking is limited to models with publicly available code.}
    \label{tab:apr_benchmarking}
    \scalebox{0.55}{
        \begin{tabular}{ccccccccccccc}
        \toprule
            \multirow{2}{*}{Model} & \multirow{2}{*}{Backbone} & \multicolumn{2}{c}{Basement} & \multicolumn{2}{c}{Lower Level} & \multicolumn{2}{c}{Level 1} & \multicolumn{2}{c}{Level 2} \\
            \cmidrule(lr){3-4} \cmidrule(lr){5-6} \cmidrule(lr){7-8} \cmidrule(lr){9-10} \cmidrule(lr){11-12}
             & & Mean & Median & Mean & Median & Mean & Median & Mean & Median \\
        \midrule
            \multirow{2}{*}{PoseNet~\cite{kendall2015posenet}} & ResNet-34 & 0.52m, 5.50$^\circ$ & 0.42m, 4.52$^\circ$ & 0.96m, 7.14$^\circ$ & 0.71m, 5.70$^\circ$ & 0.91m, 7.65$^\circ$ & 0.52m, 6.21$^\circ$ & 1.12m, 6.78$^\circ$ & 0.72m, 5.99$^\circ$\\
             & MobileNet-V3 & 0.60m, 5.78$^\circ$ & 0.52m, 5.01$^\circ$ & 0.95m, 7.10$^\circ$ & 0.69m, 5.59$^\circ$ & 0.90m, 7.94$^\circ$ & 0.50m, 6.34$^\circ$ & 1.11m, 6.91$^\circ$ & 0.73m, 6.04$^\circ$ \\
        \midrule
            Bayesian & ResNet-34 & 0.57m, 6.88$^\circ$ & 0.51m, 5.70$^\circ$ & 1.02m, 8.34$^\circ$ & 0.75m, 6.04$^\circ$ & 0.98m, 9.04$^\circ$ & 0.61m, 7.38$^\circ$ & 1.09m, 7.34$^\circ$ & 0.71m, 6.51$^\circ$ \\
            PoseNet~\cite{kendall2016modelling} & MobileNet-V3 & 0.62m, 6.99$^\circ$ & 0.60m, 5.98$^\circ$ & 1.05m, 8.73$^\circ$ & 0.79m, 6.48$^\circ$ & 1.01m, 8.93$^\circ$ & 0.72m, 6.76$^\circ$ & 1.09m, 7.52$^\circ$ & 0.72m, 6.58$^\circ$ \\
        \midrule
            LSTM- & ResNet-34 & 0.49m, 6.20$^\circ$ & 0.42m, 5.13$^\circ$ & 0.92m, 7.43$^\circ$ & 0.74m, 5.97$^\circ$ & 0.81m, 7.54$^\circ$ & 0.51m, 5.49$^\circ$ & 0.92m, 6.17$^\circ$ & 0.59m, 4.73$^\circ$ \\
            PoseNet~\cite{walch2017image} & MobileNet-V3 & 0.52m, 6.49$^\circ$ & 0.41m, 6.24$^\circ$ & 0.94m, 7.83$^\circ$ & 0.73m, 5.99$^\circ$ & 0.84m, 7.62$^\circ$ & 0.52m, 5.73$^\circ$ & 0.95m, 6.21$^\circ$ & 0.60m, 4.41$^\circ$ \\
        \midrule
            Learnable & ResNet-34 & 0.39m, 3.41$^\circ$ & 0.33m, \textbf{2.19$^\circ$} & 0.76m, 3.24$^\circ$ & 0.50m, 3.19$^\circ$ & 0.70m, 4.30$^\circ$ & 0.46m, 3.84$^\circ$ & 0.87m, 3.90$^\circ$ & 0.49m, \textbf{2.41$^\circ$} \\
             PoseNet~\cite{kendall2017geometric} & MobileNet-V3 & 0.42m, 3.90$^\circ$ & 0.34m, 2.81$^\circ$ & 0.78m, \textbf{3.20$^\circ$} & 0.53m, 3.14$^\circ$ & 0.75m, 4.76$^\circ$ & 0.47m, 3.90$^\circ$ & 0.91m, 4.74$^\circ$ & 0.46m, 3.82$^\circ$ \\
        \midrule
            Geometric & ResNet-34 & 0.41m, 4.45$^\circ$ & 0.33m, 3.01$^\circ$ & 0.80m, 3.41$^\circ$ & 0.55m, 3.24$^\circ$ & 0.73m, 4.28$^\circ$ & 0.47m, 3.87$^\circ$ & 0.90m, 5.21$^\circ$ & 0.57m, 3.43$^\circ$ \\
            PoseNet~\cite{kendall2017geometric} & MobileNet-V3 & 0.46m, 3.89$^\circ$ & 0.39m, 2.80$^\circ$ & 0.83m, 3.40$^\circ$ & 0.56m, 3.19$^\circ$ & 0.77m, 4.91$^\circ$ & 0.49m, 4.19$^\circ$ & 0.94m, 5.32$^\circ$ & 0.58m, 3.49$^\circ$ \\
        \midrule
            Hourglass & ResNet-34 & 0.45m, 5.95$^\circ$ & 0.34m, 4.91$^\circ$ & 0.88m, 5.47$^\circ$ & 0.62m, 4.79$^\circ$ & 0.81m, 6.23$^\circ$ & 0.49m, 4.02$^\circ$ & 0.85m, 6.87$^\circ$ & 0.49m, 3.96$^\circ$ \\
            PoseNet~\cite{melekhov2017image} & MobileNet-V3 & 0.53m, 6.18$^\circ$ & 0.48m, 6.03$^\circ$ & 0.89m, 5.86$^\circ$ & 0.60m, 4.79$^\circ$ & 0.83m, 6.48$^\circ$ & 0.50m, 4.18$^\circ$ & 0.94m, 8.31$^\circ$ & 0.52m, 4.37$^\circ$ \\
        \midrule
            BranchNet- & ResNet-34 & 0.43m, 5.99$^\circ$ & 0.32m, 4.72$^\circ$ & 0.79m, 4.00$^\circ$ & 0.48m, 3.25$^\circ$ & 0.94m, 9.21$^\circ$ & 0.65m, 6.27$^\circ$ & 1.04m, 7.26$^\circ$ & 0.78m, 6.13$^\circ$ \\
            Euler6~\cite{wu2017delving} & MobileNet-V3 & 0.50m, 6.23$^\circ$ & 0.41m, 4.82$^\circ$ & 0.84m, 4.31$^\circ$ & 0.51m, 3.36$^\circ$ & 0.98m, 9.74$^\circ$ & 0.66m, 6.40$^\circ$ & 1.16m, 8.17$^\circ$ & 0.82m, 7.10$^\circ$ \\
        \midrule
            \multirow{2}{*}{MapNet~\cite{brahmbhatt2018geometry}} & ResNet-34 & 0.39m, 3.41$^\circ$ & 0.31m, 2.71$^\circ$ & 0.70m, 3.75$^\circ$ & 0.42m, 3.10$^\circ$ & 0.65m, 4.58$^\circ$ & 0.44m, 3.90$^\circ$ & 0.82m, 5.43$^\circ$ & 0.48m, 3.97$^\circ$ \\
             & MobileNet-V3 & 0.40m, \textbf{3.37$^\circ$} & 0.29m, 2.65$^\circ$ & 0.74m, 3.76$^\circ$ & 0.43m, 3.22$^\circ$ & 0.66m, 4.79$^\circ$ & 0.41m, 4.21$^\circ$ & 0.86m, 4.27$^\circ$ & 0.50m, 3.21$^\circ$ \\
        \midrule
            \multirow{2}{*}{MSPN~\cite{blanton2020extending}} & ResNet-34 & 0.39m, 3.40$^\circ$ & 0.30m, 2.68$^\circ$ & 0.68m, 3.70$^\circ$ & 0.42m, 3.13$^\circ$ & 0.69m, 5.17$^\circ$ & 0.47m, 3.96$^\circ$ & 0.91m, \textbf{3.71$^\circ$} & 0.48m, 2.53$^\circ$ \\
             & MobileNet-V3 & 0.41m, 3.38$^\circ$ & 0.30m, 2.58$^\circ$ & 0.73m, 3.94 & 0.45m, 3.27$^\circ$ & 0.71m, 5.32$^\circ$ & 0.48m, 3.87$^\circ$ & 0.92m, 4.14$^\circ$ & 0.49m, 3.04$^\circ$ \\
        \midrule
        Direct- & ResNet-34 & 0.35m, 3.94$^\circ$ & 0.27m, 3.10$^\circ$ & 0.69m, 3.77$^\circ$ & 0.44m, 2.69$^\circ$ & 0.63m, 4.74$^\circ$ & 0.41m, 3.90$^\circ$ & 0.84m, 3.94$^\circ$ & 0.46m, 2.97$^\circ$ \\
        PoseNet~\cite{chen2021direct} & MobileNet-V3 & 0.39m, 3.49$^\circ$ & 0.28m, 2.88$^\circ$ & 0.70m, 3.71$^\circ$ & 0.45m, 2.71$^\circ$ & 0.61m, 4.82$^\circ$ & 0.42m, 3.89$^\circ$ & 0.88m, 4.10$^\circ$ & 0.45m, 3.04$^\circ$ \\
        \midrule
            MS- & ResNet-34 & \textbf{0.35m}, 4.47$^\circ$ & \textbf{0.26m}, 3.78$^\circ$ & 0.63m, 3.74$^\circ$ & 0.43m, 2.50$^\circ$ & \textbf{0.60m}, 4.66$^\circ$ & \textbf{0.41m}, 3.84$^\circ$ & 0.83m, 4.21$^\circ$ & 0.43m, 2.76$^\circ$ \\
            Transformer~\cite{shavit2021learning} & MobileNet-V3 & 0.44m, 5.26 & 0.36m, 4.19$^\circ$ & 0.65m, 3.89$^\circ$ & \textbf{0.41m, 2.43$^\circ$} & 0.61m, \textbf{4.06$^\circ$} & 0.45m, \textbf{3.17$^\circ$} & 0.85m, 4.18$^\circ$ & 0.43m, 2.96$^\circ$ \\
        \midrule
            \multirow{2}{*}{PAE~\cite{shavit2022camera}} & ResNet-34 & 0.40m, 3.99$^\circ$ & 0.27m, 2.87$^\circ$ & 0.71m, 3.79$^\circ$ & 0.45m, 2.57$^\circ$ & 0.71m, 5.23$^\circ$ & 0.49m, 4.15$^\circ$ & 0.96m, 4.73$^\circ$ & 0.51m, 3.25$^\circ$\\
             & MobileNet-V3 & 0.41m, 4.10$^\circ$ & 0.27m, 2.93$^\circ$ & 0.73m, 3.80$^\circ$ & 0.47m, 2.63$^\circ$ & 0.71m, 5.23$^\circ$ & 0.49m, 4.15$^\circ$ & 0.98m, 4.69$^\circ$ & 0.52m, 3.09$^\circ$ \\
        \midrule
            \multirow{2}{*}{DFNet~\cite{chen2022dfnet}} & ResNet-34 & 0.36m, 3.49$^\circ$ & 0.28m, 2.58$^\circ$ & \textbf{0.60m}, 3.80$^\circ$ & 0.43m, 2.78$^\circ$ & 0.64m, 4.82$^\circ$ & 0.43m, 3.85$^\circ$ & 0.81m, 4.37$^\circ$ & \textbf{0.43m}, 2.81$^\circ$\\
             & MobileNet-V3 & 0.40m, 3.56$^\circ$ & 0.34m, 3.16$^\circ$ & 0.61m, 3.72$^\circ$ & 0.42m, 2.55$^\circ$ & 0.67m, 5.01$^\circ$ & 0.47m, 3.91$^\circ$ & \textbf{0.80m}, 4.26$^\circ$ & 0.45m, 2.90$^\circ$ \\
        \bottomrule
        \end{tabular}
    }
\end{table}

\textbf{Settings.} To ensure a fair comparison in benchmarking camera relocalization, we employed identical CNN architectures—ResNet-34 \cite{He_2016_CVPR} and MobileNet-V3 \cite{howard2019searching}—across 13 models ranging from the pioneering PoseNet \cite{kendall2015posenet} to its latest advancements \cite{shavit2022camera, chen2022dfnet}. All the above models were trained for 200 epochs. Specifically, for the PoseNet-series models, we utilized the loss function defined in Eq.(\ref{eq:posenet_loss}), setting $\beta$ to $e^5$ across all four floors\footnote{Specifically tuning $\beta$ for each floor may yield better results.}. In the Bayesian PoseNet, uncertainty is incorporated only before the layers with randomly initialized weights, following~\cite{kendall2016modelling}. LSTM-PoseNet models have the hidden size of all LSTM units set at 256 to achieve optimal results \cite{walch2017image}. Learnable PoseNet starts with initial guesses of $s_x$ and $s_q$ set to $0$ and $-5$, respectively, for all scenes~\cite{kendall2017geometric}, and then Geometric PoseNet continues training these models using geometric reprojection data to balance positional and rotational errors per image. We employed feature map concatenation in ResNet-34 and element-wise summation in MobileNet-V3 to combine features from the front layers and achieve the optimal results in Hourglass PoseNet~\cite{melekhov2017image}. For BranchNet-Euler6~\cite{wu2017delving}, the networks are split before the final convolutional block to facilitate multi-task learning. MapNet~\cite{brahmbhatt2018geometry} configurations avoid updating model weights with unlabeled data to maintain fairness in APR comparisons.
Both MapNet and MSPN~\cite{blanton2020extending} adopt log-quaternion for rotation representation.
Direct-PoseNet maintains a direct matching ratio of photometric difference to the loss function Eq.(\ref{eq:posenet_loss}) at $3/7$~\cite{chen2021direct}. MS-Transformer~\cite{shavit2021learning} replaces only the convolutional backbones to extract activation maps while preserving the encoder-decoder architecture of Transformers. PAE embeds $\xbold$ and $\qbold$ using Fourier Features with expanded dimensions of 12 ($L=6$ in \cite[Eq.(5)]{shavit2022camera}). Finally, DFNet results are confined to single-frame APR~\cite{chen2022dfnet}. For more details of learning settings, please refer to \textbf{Appendix C} and the publicly available code.

\begin{wrapfigure}[12]{r}{0.3\textwidth}
\vspace{-5pt}
        \centering
        \includegraphics[width = 0.99\linewidth, height = 0.5\linewidth]{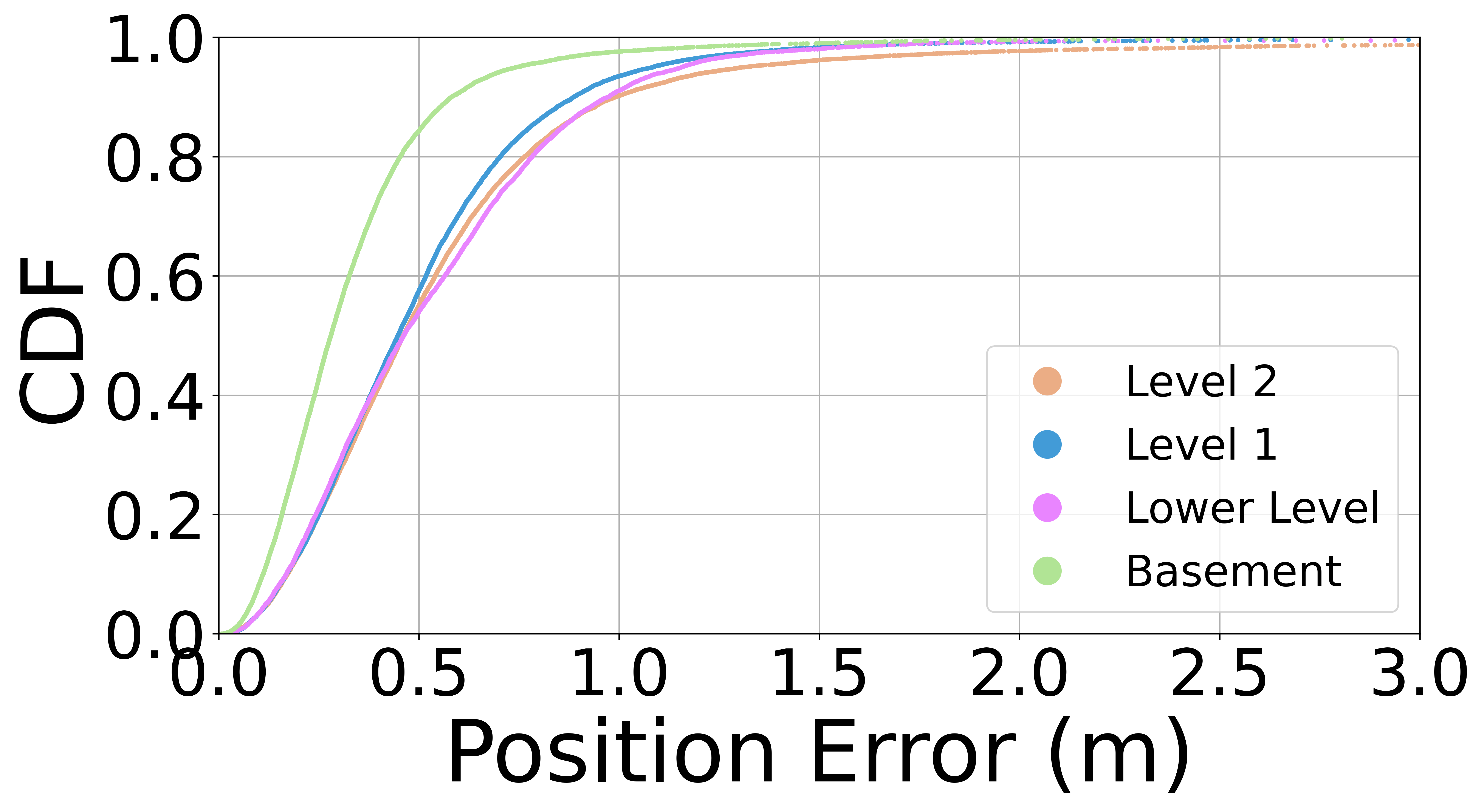}
        \includegraphics[width = 0.99\linewidth, height = 0.5\linewidth]{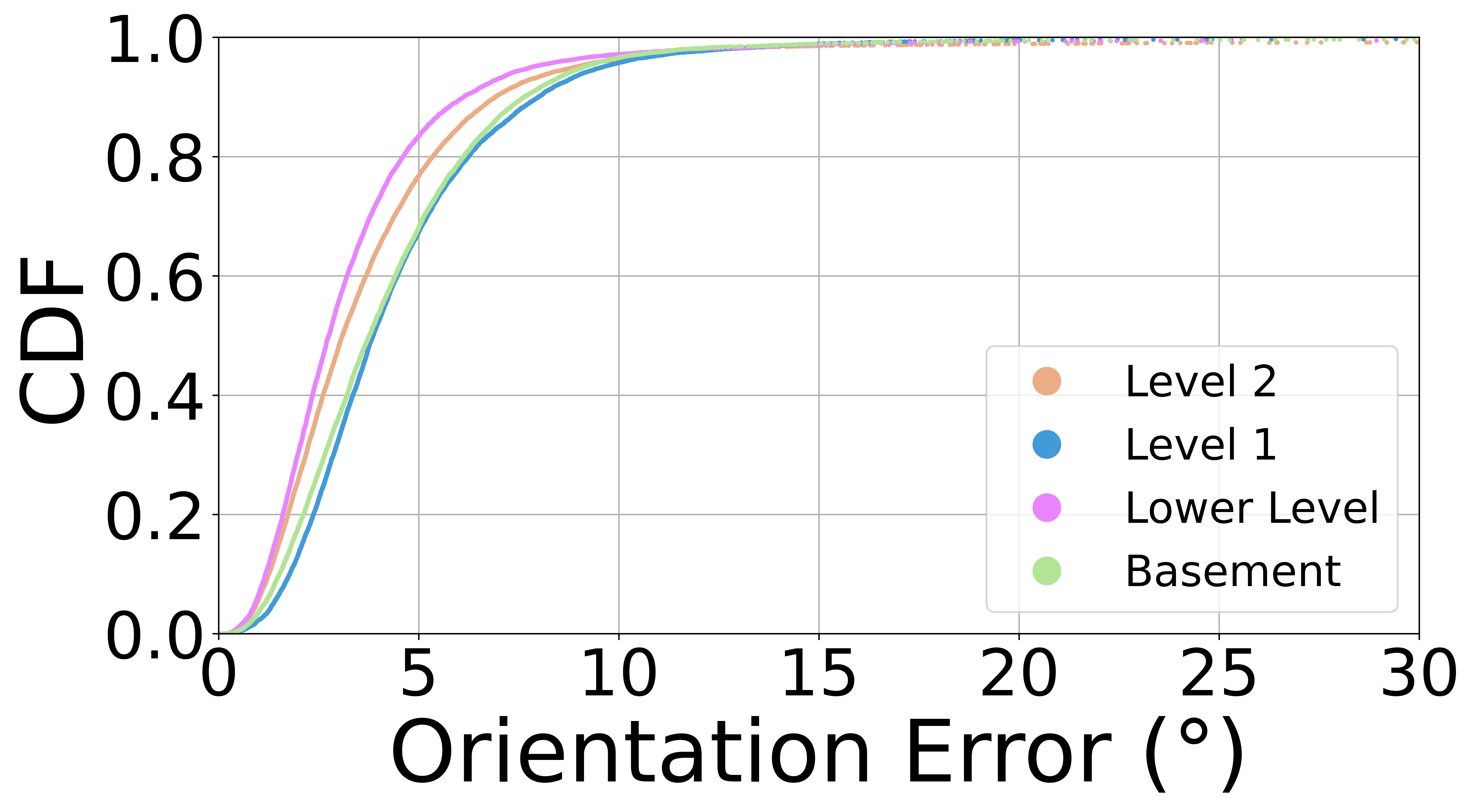}
        \vspace{-19pt}
        \caption{CDF of errors on MS-Transformer.}
    \label{fig:cdf}
\end{wrapfigure}
\textbf{Results.} Table \ref{tab:apr_benchmarking} shows the mean and median errors for both camera positions and orientations. Notably, even the pioneering work of PoseNet~\cite{kendall2015posenet} from 2015 can achieve sub-meter accuracy, meeting the requirements for indoor navigation applications. We observe that the best performance in average is achieved by MS-Transformer~\cite{shavit2021learning}, which benefits from its large model capacity enabled by the use of Transformers and a multi-scene mixing training strategy. Figure~\ref{fig:cdf} illustrates the cumulative distribution function (CDF) of MS-Transformer's predictions regarding mean errors in positions and orientations across four floors. For additional experimental results and visualization, please refer to \textbf{Appendix C}.



\subsection{Image Captioning}
\label{subsec:image_captioning}
We can train a captioning model using outputs from GPT-4 to meet various VIPs needs. Image captioning requires fluent descriptions to translate images into natural language~\cite{vinyals2016show, hossain2019comprehensive}. We mix image data from four floors and use concise descriptions as ground truth to distill~\cite{hinton2015distilling} our own models. We validate the results using BLEU-4~\cite{papineni2002bleu}, METEOR~\cite{denkowski2014meteor}, CIDEr~\cite{vedantam2015cider}, and SPICE~\cite{anderson2016spice} metrics. Additionally, we report the model size in the number of parameters and training time in GPU hours to indicate the feasibility and practicality of this visual intelligence support. The results in Table~\ref{tab:caption} demonstrate accurate predictions with smaller-size distilled models.

\begin{table}[t]
    \centering
    \caption{Experimental results on concise descriptions.}
    \label{tab:caption}
    \scalebox{0.75}{
        \begin{tabular}{cccccccc}
        \toprule
            Model & Backbone & BLEU-4$\uparrow$ & METEOR$\uparrow$ & CIDEr$\uparrow$ & SPICE$\uparrow$ & \#Params & Training Time$\downarrow$ \\
        \midrule
            ClipCap~\cite{mokady2021clipcap} & CLIP (ViT-B/32) + GPT-2 tuning & 35.18 & 27.34 & 113.83 & 20.19 & 156 M & 46h (A6000)\\
            OFA$_{\text{base}}$~\cite{wang2022ofa} & ResNet101 + Transformer & 36.31 & 30.02 & 126.77 & 26.85 & 180 M & 53h (A6000)\\
        \bottomrule
        \end{tabular}
    }
\end{table}

\section{Potential Applications}
\textbf{APR methods development.} Popular datasets commonly used for benchmarking APR methods include 7-Scenes~\cite{shotton2013scene} and Cambridge Landmarks~\cite{kendall2015posenet}. These datasets, however, are limited in scope and size, which can lead to the overfitting phenomenon already observed in previous research~\cite{kendall2015posenet, kendall2017geometric, walch2017image, melekhov2017image}. This limitation complicates the performance assessment of new proposed models, particularly in an era dominated by deep learning and PFMs. For instance, Bayesian PoseNet~\cite{kendall2016modelling} demonstrates inferior performance compared to the classic PoseNet using our dataset. This discrepancy arises because any dropout rate applied to PoseNet constrains its capacity rather than serving its intended purpose of regularization.

\textbf{VLN test under real-world environments.} We plan to release this comprehensive dataset that includes not only the original videos but also the floor plans annotated with PoIs. Each video can be segmented into various clips representing a unique navigation path. These clips are associated with an automated point-to-point (PoI-to-PoI) oracle that can be used for VLN model training. Moreover, since descriptions for each image will be available, exploring a VLN model based on PoI-to-PoI instruction that operates without the need for detailed step-by-step guidance in natural language appears promising, particularly in unknown real-world environments. Our dataset has the potential to significantly facilitate more robust and flexible navigation solutions, essential for navigating dynamic or unfamiliar spaces effectively.

\textbf{Floor plan navigation development.} Each image in our dataset is geo-registered to a corresponding floor plan, facilitating the development of learning-based methods that utilize only RGB images for localization. While LiDAR-based localization techniques have been explored extensively in recent research~\cite{boniardi2017robust, boniardi2019pose, li2020online, mendez2020sedar}, their practical application is often constrained by the hardware capabilities of commonly used mobile devices. In contrast, we anticipate that purely image-based floor plan navigation will become more viable with the availability of this large indoor navigation dataset. This could pave the way for more accessible and widely deployable navigation solutions that leverage visual data alone.

\textbf{Deployment and test of ``NaVIP'' everywhere.} We will provide comprehensive details regarding the setup and organization of our \textbf{\texttt{NaVIP}} solution within the Lehigh HST building. The data collection process requires only a mobile phone, and to simplify this process, we will also release the corresponding code, including data pre-processing and annotation. Given the limited time and resources invested in developing this effective pipeline, we have not yet tested our solutions in other buildings. We encourage any groups seeking an affordable and intelligent indoor navigation system to adopt and apply this pipeline in their settings.
\section{Limitations and Discussion}
\label{sec:limitations}
\textbf{Inherent biases of human data collection.} In \textbf{\texttt{NaVIP}}, we collected video data via sighted individuals, which inevitably introduced biases from collectors themselves. To mitigate this and reduce human labor in developing such data-driven navigation systems, one possible solution is the use of a simple robot equipped with a phone camera that navigates randomly to gather video data. Nonetheless, both sighted individuals and robotic agents fail to capture the user needs and data distribution pertinent to VIPs. Inspired by VizWiz~\cite{bigham2010vizwiz}, involving VIPs as participants in the data collection process is beneficial as our primary objective is to assist VIPs in navigating unfamiliar environments. Efforts to bridge this limitation will not only enhance the practicality of the solutions but also ensure their genuine benefit to the intended users.

\textbf{How to merge 3D reconstruction losslessly?} In our dataset, we utilize COLMAP to reconstruct the 3D models of indoor environments along each video path. Given that indoor environments can be ever-changing, there currently lacks a robust algorithm to effectively merge these 3D point clouds from different time. As a workaround, we utilize human supervision to pinpoint several anchor images directly onto the floor plan. These annotated points facilitate the geo-registration process built in COLMAP to transform the coordinates of all 3D point clouds. While this alignment method is not lossless--resulting in an inevitable system error in camera poses of approximately 0.5 meters--we will release both the sparse and dense models generated by COLMAP, before and after geo-registration, to facilitate further research in this area. This contribution is expected to further enhance the accuracy of positioning system in indoor navigation.

\textbf{How robust is the image-centric positioning solution?} We acknowledge the limitations in our experimental analysis regarding the robustness of this image-centric positioning system. Although our dataset was collected directly from real-world scenarios using common mobile devices, the performance of this system under extreme conditions, e.g., electronic failures leading to dark environments or post-construction changes within buildings, remains unexplored. To simulate a distribution shift due to changes in the building environment, our dataset incorporates a temporal gap between the training and testing datasets. Despite this, we observe no decline in performance over time.

\textbf{Is GPT-4 ready for assisting VIPs?} Although we meticulously designed prompts tailored to the needs of VIPs (refer to \textbf{Appendix B} for our exploration), we encountered challenges in meeting their personalized requirements. Key challenges include: 1) optimizing GPT-4 as an image descriptor for VIPs to enhance accuracy and eliminate misinformation, and 2) developing prompts that guide GPT-4 to generate information that aligns with user expectations. To further explore these issues, we have released the outputs of GPT-4 on our dataset of 300K images.
\section{Conclusion and Future Work}
In this paper, we have redirected our research from conventional sensor-based navigation systems to purely vision-based solutions for indoor environments. To facilitate this shift, we created a large image-centric dataset, named \textbf{\texttt{NaVIP}}, within the largest building at Lehigh University, the HST, specifically for research purposes. Our comprehensive pilot experiments on benchmarking APR methods by leveraging real-time end-to-end inference in deep neural networks have validated its feasibility and accuracy. This solution not only streamlines the system architecture but also enhances its applicability and usability, thereby extending its utility in assisting VIPs. The integration of image captioning models distilled from GPT-4 further highlights the potential to independent exploration for VIPs. As we look to the future, our research will focus on developing a mobile application that incorporates our trained model. We plan to test within the Lehigh community, supported by approval from the Institutional Review Board (IRB) to ensure comprehensive human feedback integration into the study. Moreover, we intend to explore advanced unsupervised learning techniques and reinforcement learning from human feedback (RLHF) to further enhance the functionality and user experience of our mobile application. The next phase of our research will concentrate on incorporating real-world user insights to refine and optimize the navigational aid, aiming to create a more adaptive and effective tool for end-users.
\begin{ack}
We thank HST building coordinator Emily Diaz-Kempf and laboratory manager Chris Panko Graff for their support in video recording in the HST public area. We also extend our gratitude to Eashan Adhikarla for providing the computer used to run COLMAP in parallel, and Deven Bhadane for his time and effort in designing the logo for this work. Special thanks to our visually impaired friend, Joel Isaac, for volunteering and sharing their needs for an indoor navigation application. This work was partially supported by the U.S. National Science Foundation through awards \#2409227, \#2340870, and \#2345057.
\end{ack}
\clearpage

{
\small
\bibliographystyle{unsrt}
\bibliography{reference}
}
\clearpage

\medskip


\clearpage

\appendix
\vspace{-30pt}
\section{Dataset Details}
\subsection{Organization}
To ensure consistency and facilitate ease of use for others, we organize our data using the COLMAP output structure~\cite{schoenberger2016sfm, schoenberger2016mvs, schoenberger2016vote} and adhere to the training standards derived from PoseNet~\cite{kendall2015posenet}. Our data maintains the same output format as that of COLMAP\footnote{\url{https://colmap.github.io/format.html}} for each video. Below is an example of the file structure for each video folder; we have a total of 400 COLMAP project folders structured in this manner:
\\
\\
\\
\\
\\
\\
\\
\\
\\
\\
\\
\\
\\
\\
\\
\\
\\
\\
\\
\\
\\
\\
\\
\\
\\
\\
\\
\\
\\
\begin{applebox}{Data Structure}
\dirtree{%
    .0 ./.
    .1 train/.
    .2 Basement/.
    .3 20231220\_141254\_proj/.
    .4 HAND\_20231220\_141254/.
    .5 HAND\_20231220\_141254\_frame\_000.3s.jpg.
    .5 ....
    .5 HAND\_20231220\_141254\_frame\_166.7s.jpg.
    .4 dense/.
    .5 0/.
    .6 images/.
    .7 HAND\_20231220\_141254\_frame\_000.3s.jpg.
    .7 ....
    .7 HAND\_20231220\_141254\_frame\_166.7s.jpg.
    .6 sparse/.
    .7 cameras.bin.
    .7 images.bin.
    .7 points3D.bin.
    .6 stereo/.
    .7 patch-match.cfg.
    .7 fusion.cfg.
    .7 ....
    .4 sparse/.
    .5 0/.
    .6 cameras.bin.
    .6 images.bin.
    .6 points3D.bin.
    .5 geo/.
    .6 cameras.bin.
    .6 images.bin.
    .6 points3D.bin.
    .4 movie/.
    .5 frame000000.png.
    .5 ....
    .5 frame000001.png.
    .4 HAND\_20231220\_141254.MOV.
    .4 camera2world\_6DoF.txt.
    .4 database.db.
    .4 geo\_coord.txt.
    .4 log.log.
    .4 path.png.
    .4 path\_stem.png.
    .3 ....
    .3 20240331\_083440\_proj/.
    .4 ....
    .3 image\_train\_all.txt.
    .3 geometric\_data.pkl.
    .2 Lower\_Level.
    .3 ....
    .2 ....
    .1 test/.
    .2 ....
}
\end{applebox}

For files or folders related to the standard outputs produced by COLMAP, please refer to the official tutorial\footnote{\url{https://colmap.github.io/}}. We will describe additional files. All video files, which end with .MOV or .MP4, are named beginning with the capture ways (either ``HAND'' or ``DJI''), followed by the timestamp of the recording. This naming convention is also applied to the folders containing extracted images and the images themselves, with the addition of a frame timestamp at the end. It is noticed that we preserved the padded image frames for some videos into another folder, named beginning with "HAND\_pad" or "DJI\_pad". The following folder structure is used to initiate any COLMAP processing:

\begin{applebox}{Data Sturcture Before Running COLMAP}
\dirtree{%
    .1 20231220\_141254\_proj/.
    .2 HAND\_20231220\_141254/.
    .3 HAND\_20231220\_141254\_frame\_000.3s.jpg.
    .3 ....
    .3 HAND\_20231220\_141254\_frame\_166.7s.jpg.
}
\end{applebox}

\subsubsection{Geo-Registration}

\begin{figure}[ht]
    \centering
    \begin{subfigure}[b]{0.22\linewidth}
        \includegraphics[width=\linewidth]{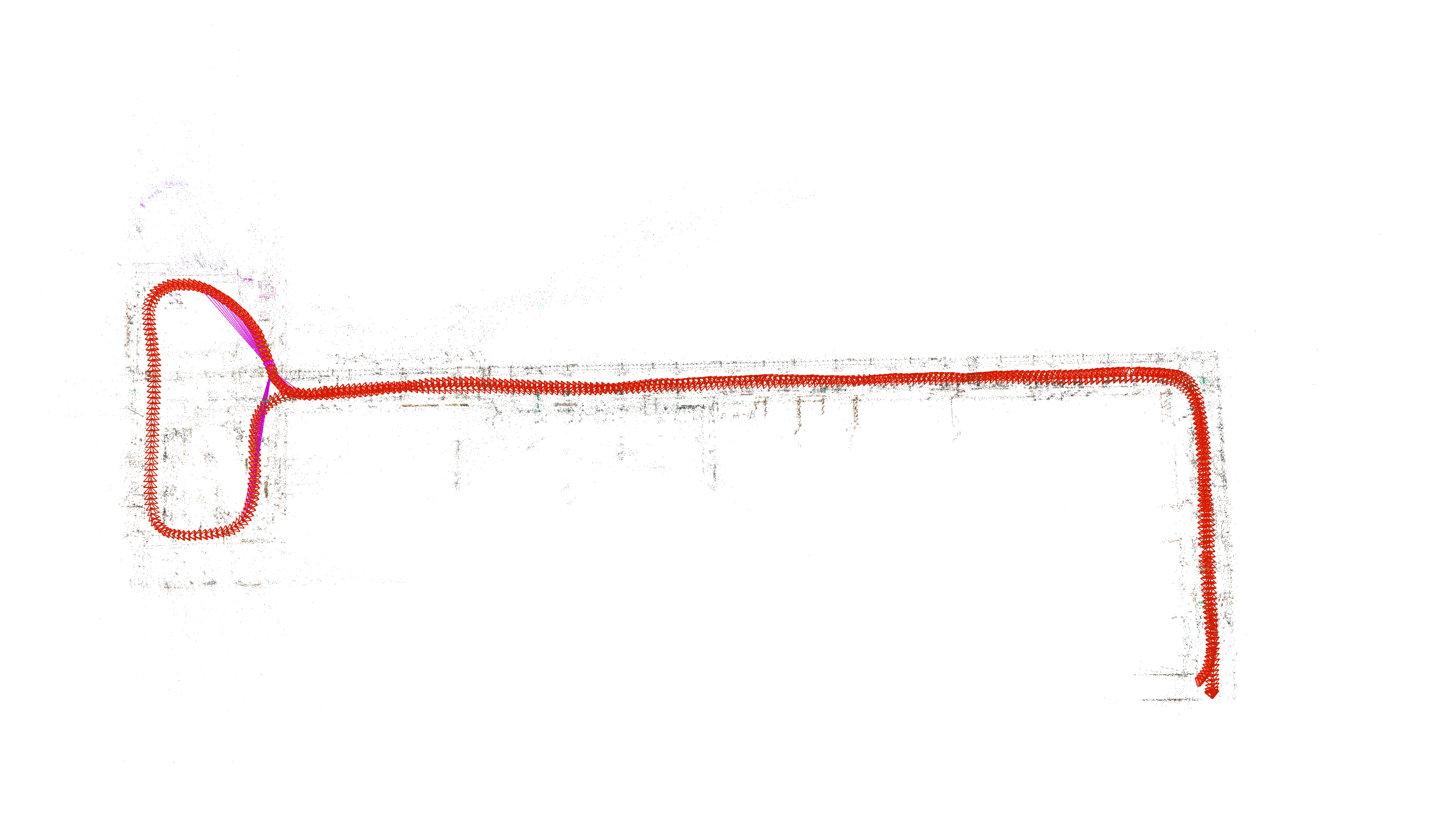}
        \caption{frame000000}
    \end{subfigure}
    \begin{subfigure}[b]{0.22\linewidth}
        \includegraphics[width=\linewidth]{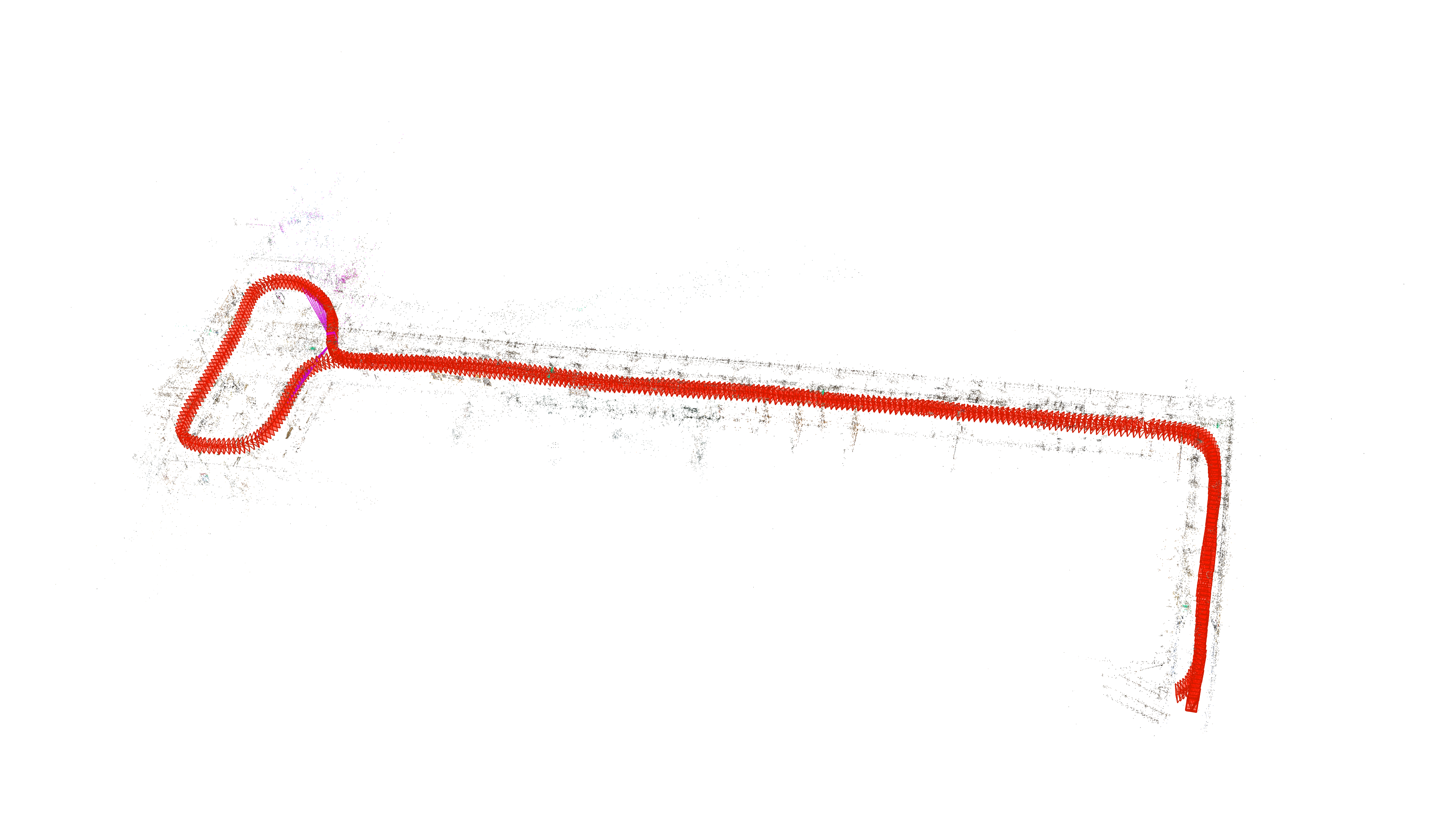}
        \caption{frame000001}
    \end{subfigure}
    \begin{subfigure}[b]{0.22\linewidth}
        \includegraphics[width=\linewidth]{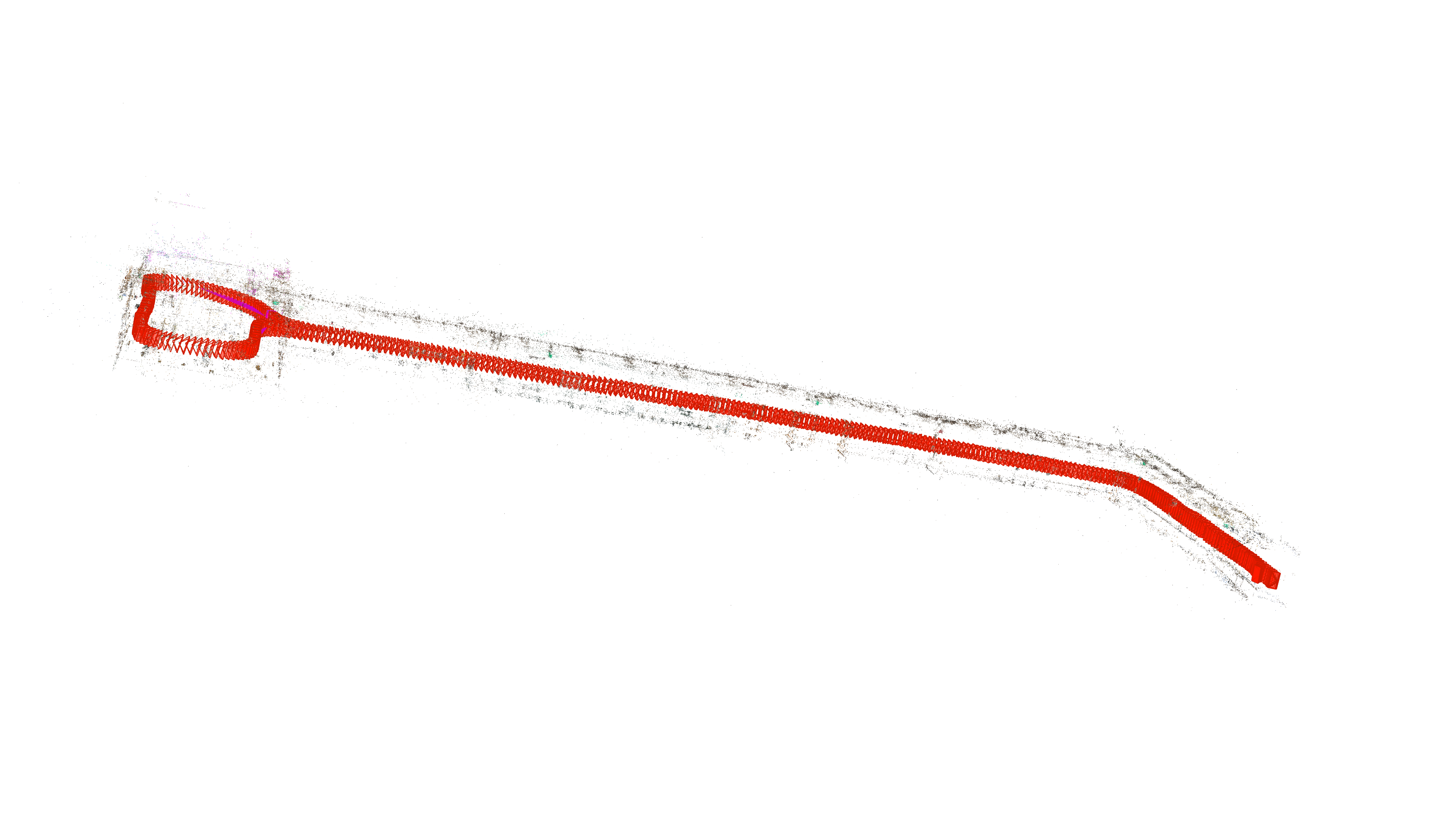}
        \caption{frame000002}
    \end{subfigure}
    \begin{subfigure}[b]{0.22\linewidth}
        \includegraphics[width=\linewidth]{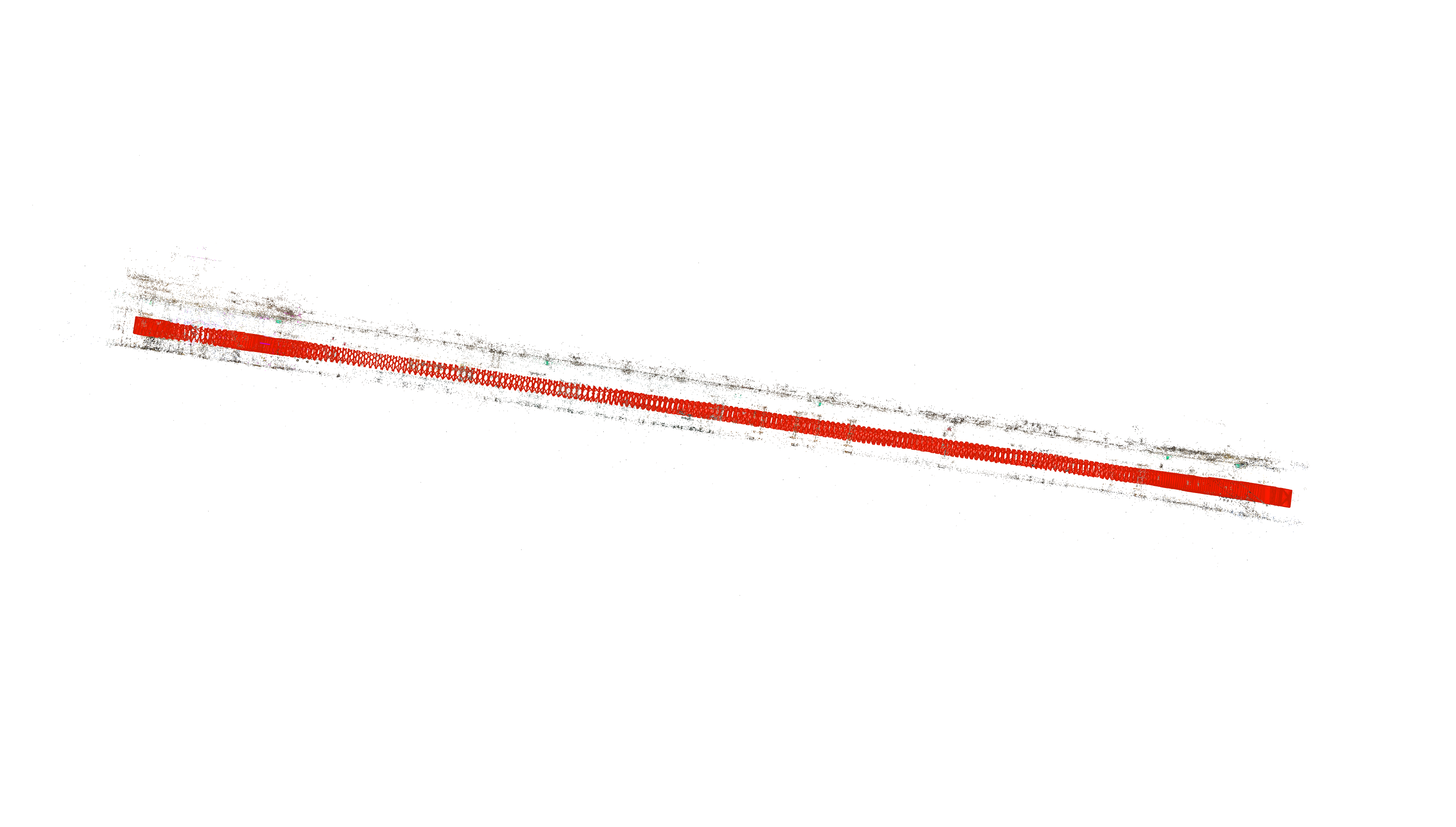}
        \caption{frame000003}
    \end{subfigure}
    \begin{subfigure}[b]{0.22\linewidth}
        \includegraphics[width=\linewidth]{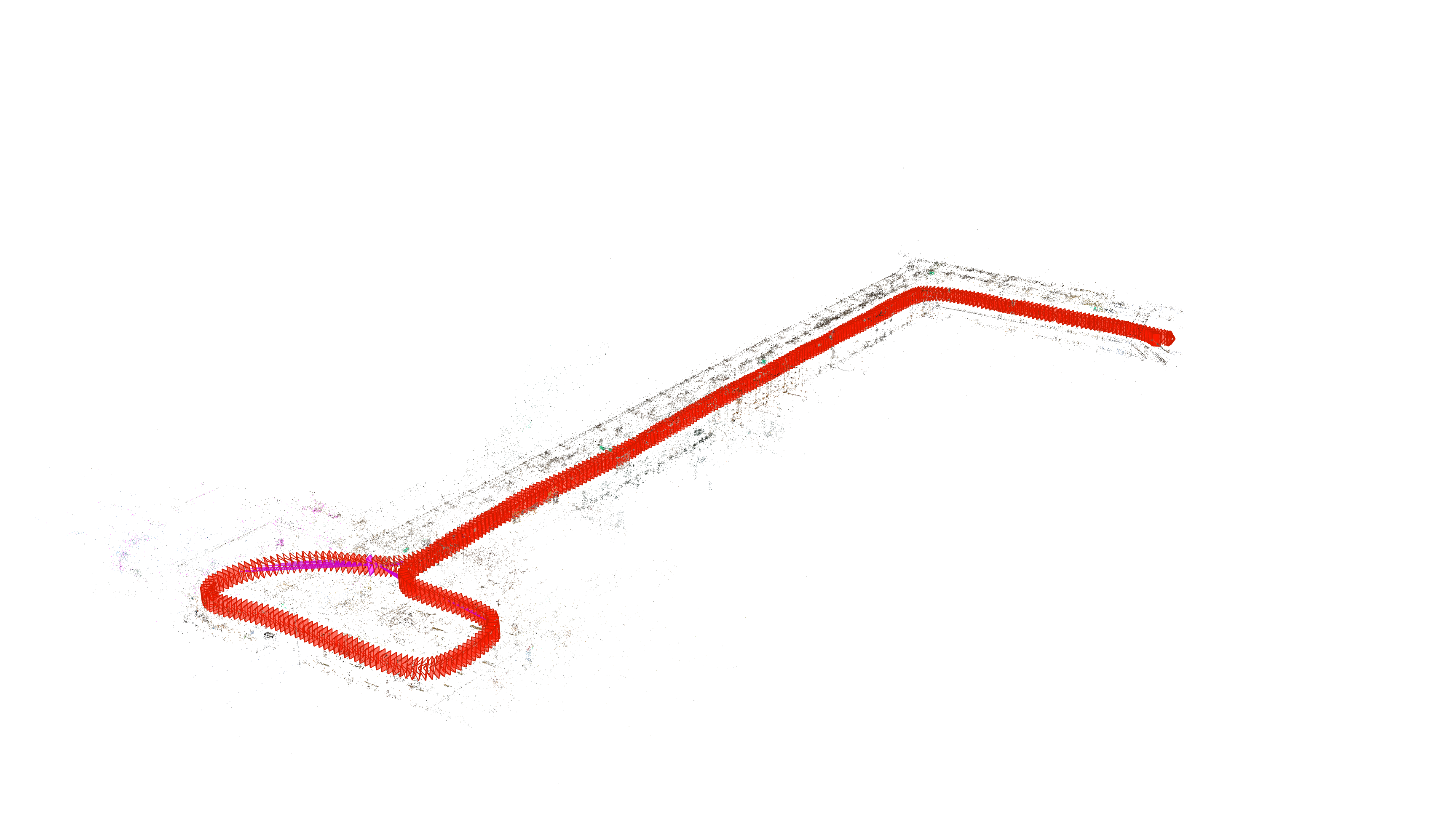}
        \caption{frame000004}
    \end{subfigure}
    \begin{subfigure}[b]{0.22\linewidth}
        \includegraphics[width=\linewidth]{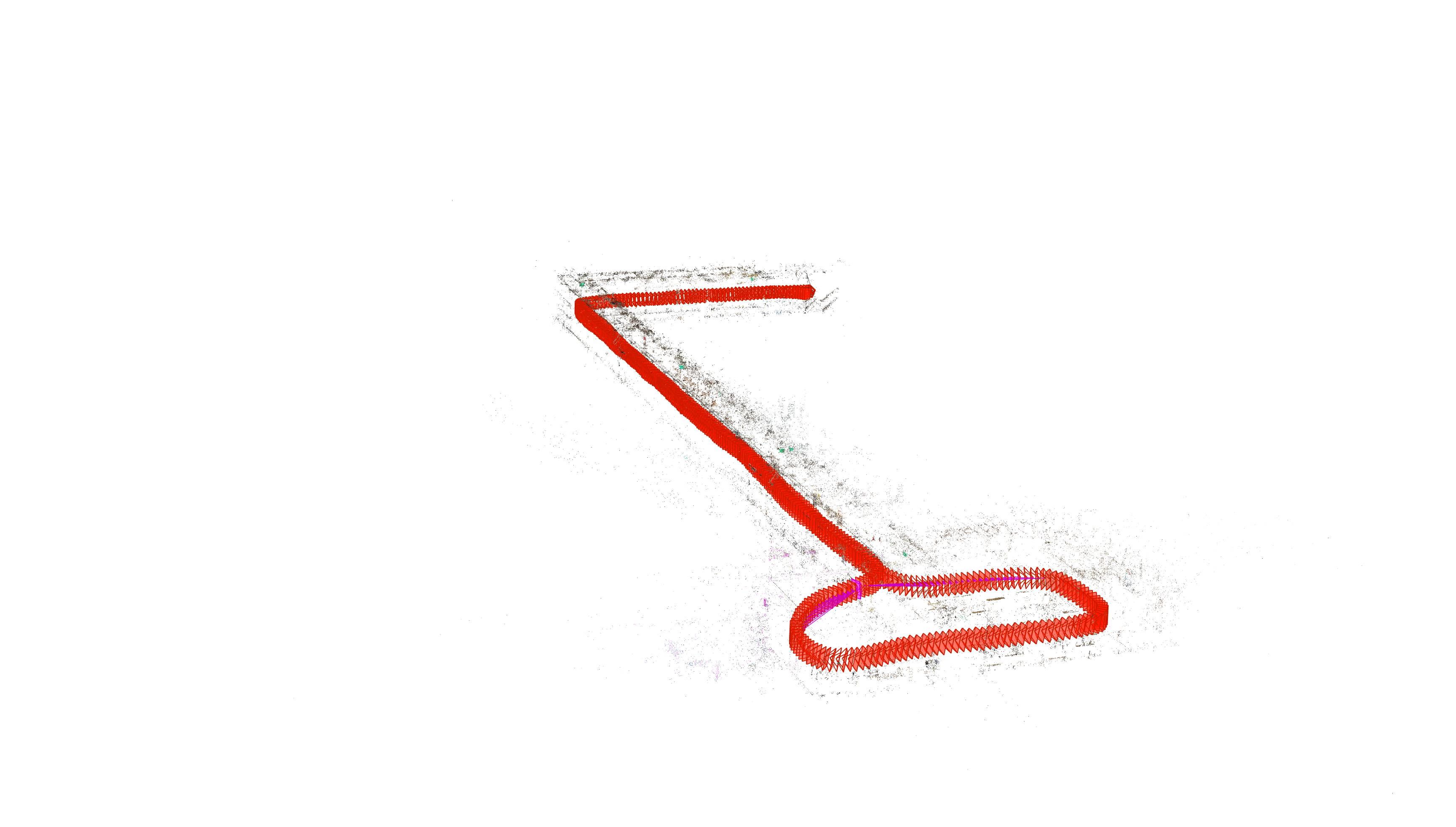}
        \caption{frame000005}
    \end{subfigure}
    \begin{subfigure}[b]{0.22\linewidth}
        \includegraphics[width=\linewidth]{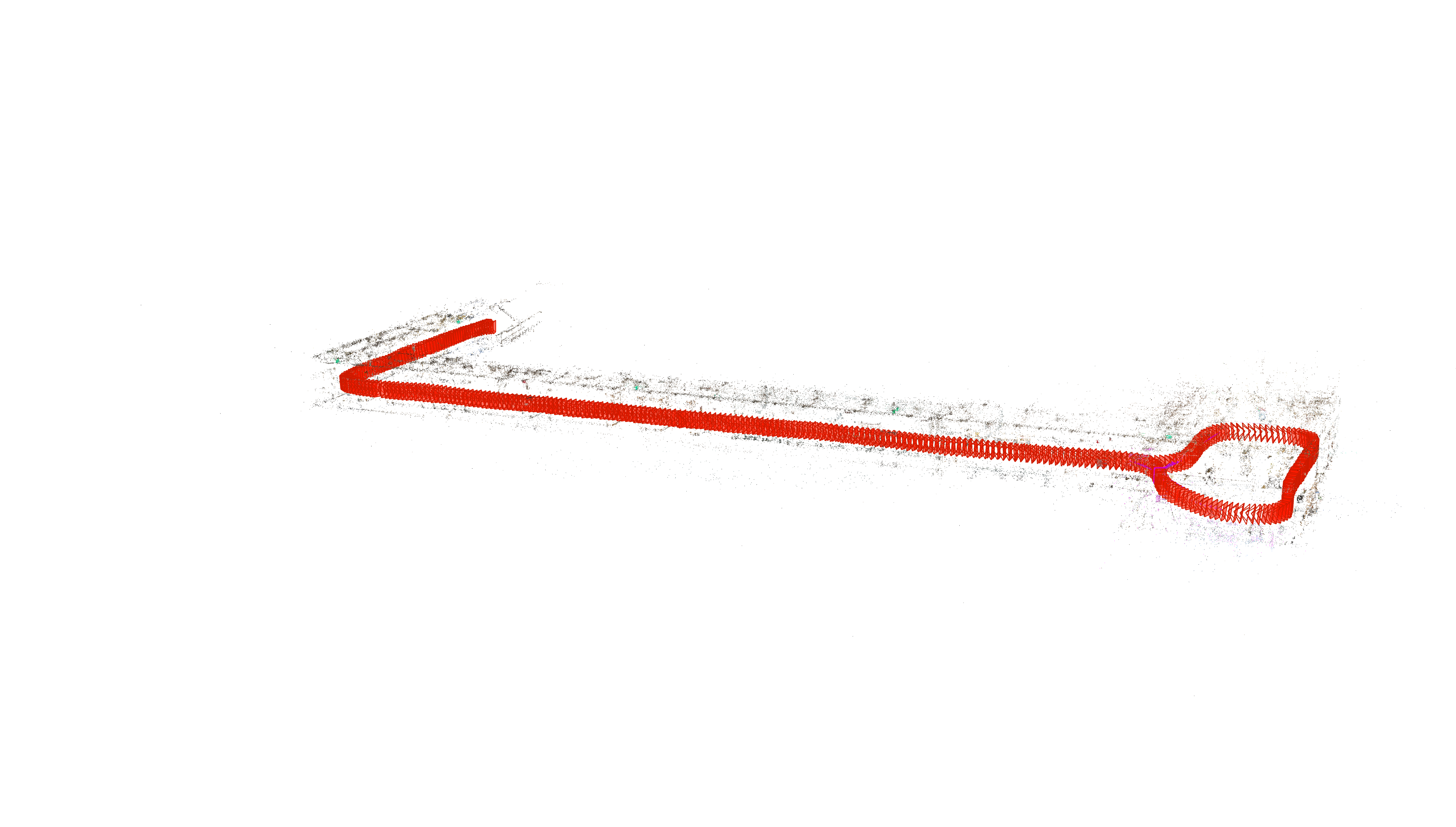}
        \caption{frame000006}
    \end{subfigure}
    \begin{subfigure}[b]{0.22\linewidth}
        \includegraphics[width=\linewidth]{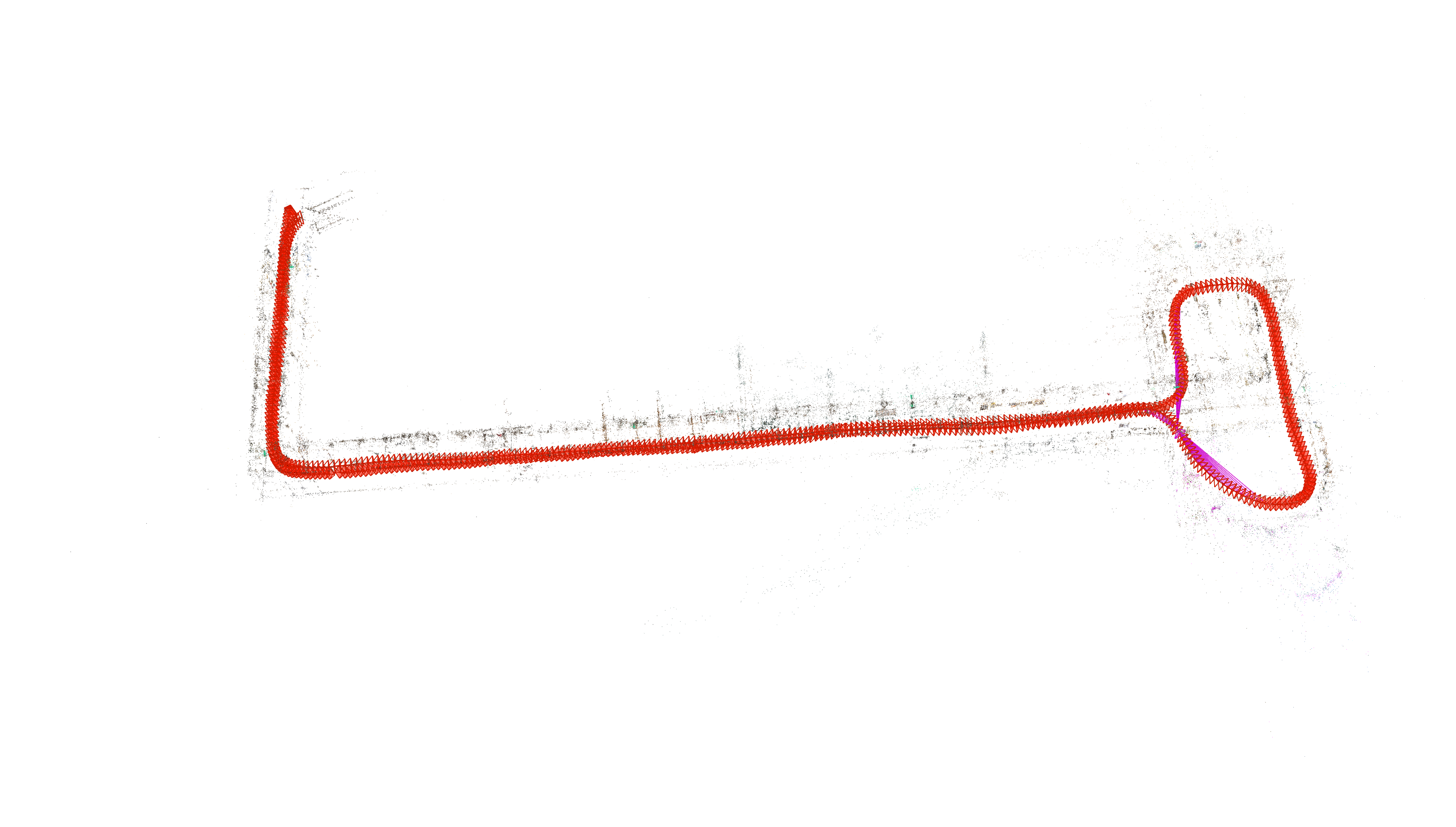}
        \caption{frame000007}
    \end{subfigure}
    \caption{Movies grabbed from the 3D reconstruction of the project \texttt{20231220\_141254\_proj/}.}
    \label{fig:movies}
\end{figure}

After generating the standard outputs using COLMAP, we first stored the training report of COLMAP to the file \texttt{log.log}, and then captured ``movies'' of the 3D reconstructions to provide readers with a direct view of the reconstruction status. For each reconstruction, we saved 5 to 10 images from various perspectives. Please refer to Figure~\ref{fig:movies} as an example.

Users can reload models via the COLMAP GUI to view the 3D reconstructions from any perspective.

Additionally, because the original world coordinates are not aligned for different paths, we need to complete the geo-registration process. This involves using 5 to 10 images, manually pinpointed to the floor plan, to accurately align the models. These ground truths are saved in the file \texttt{geo\_coord.txt} as

\begin{applebox}{Ground Truth Data for Image Localization Relative to the Floor Plan}
1. HAND\_20231220\_141254\_frame\_000.3s.jpg 1052 2113 0

2. HAND\_20231220\_141254\_frame\_019.3s.jpg 464 2082 0

3. HAND\_20231220\_141254\_frame\_035.4s.jpg 449 1503 0

4. HAND\_20231220\_141254\_frame\_074.8s.jpg 253 188 0

5. HAND\_20231220\_141254\_frame\_089.9s.jpg 700 292 0

6. HAND\_20231220\_141254\_frame\_100.1s.jpg 467 434 0

7. HAND\_20231220\_141254\_frame\_113.1s.jpg 450 915 0
\end{applebox}

\begin{figure}[h]
    \centering
    \begin{subfigure}[b]{0.42\linewidth}
        \includegraphics[width=\linewidth]{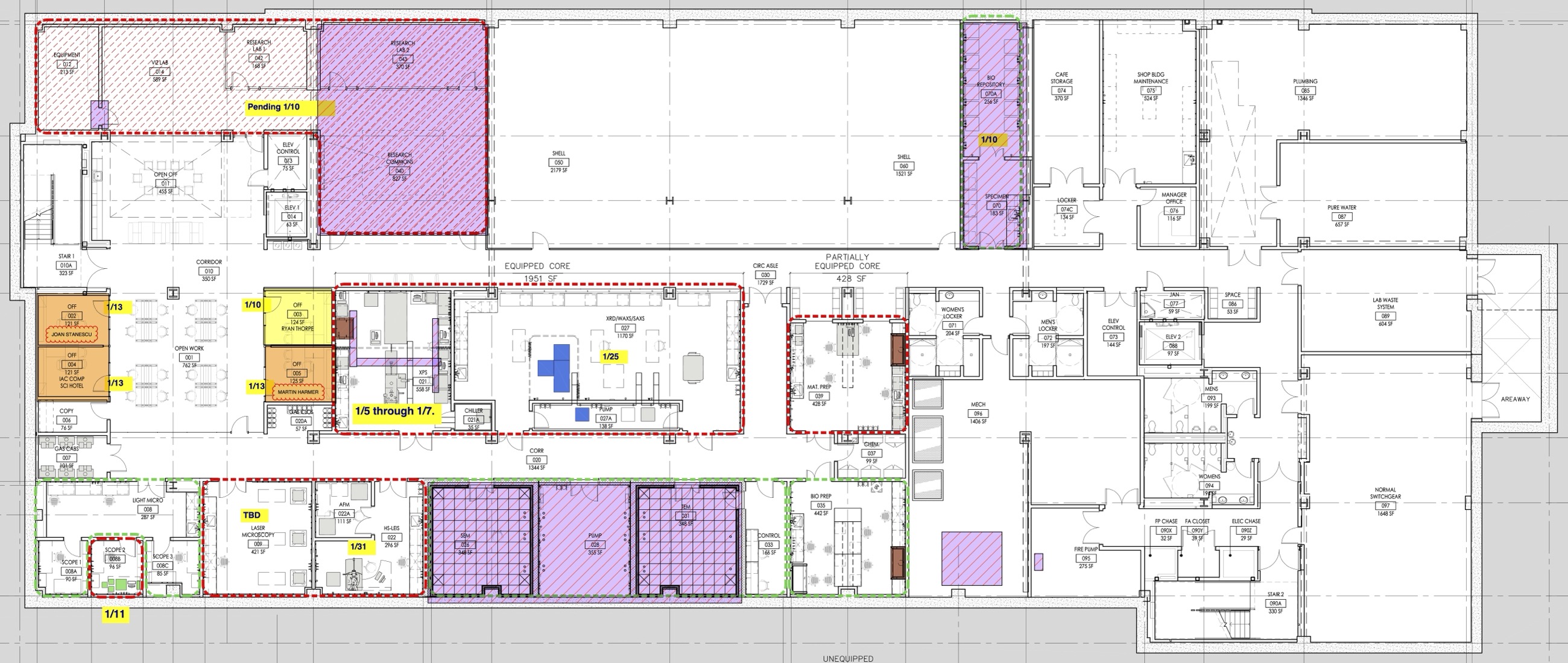}
        \caption{Basement}
    \end{subfigure}\quad
    \begin{subfigure}[b]{0.42\linewidth}
        \includegraphics[width=\linewidth]{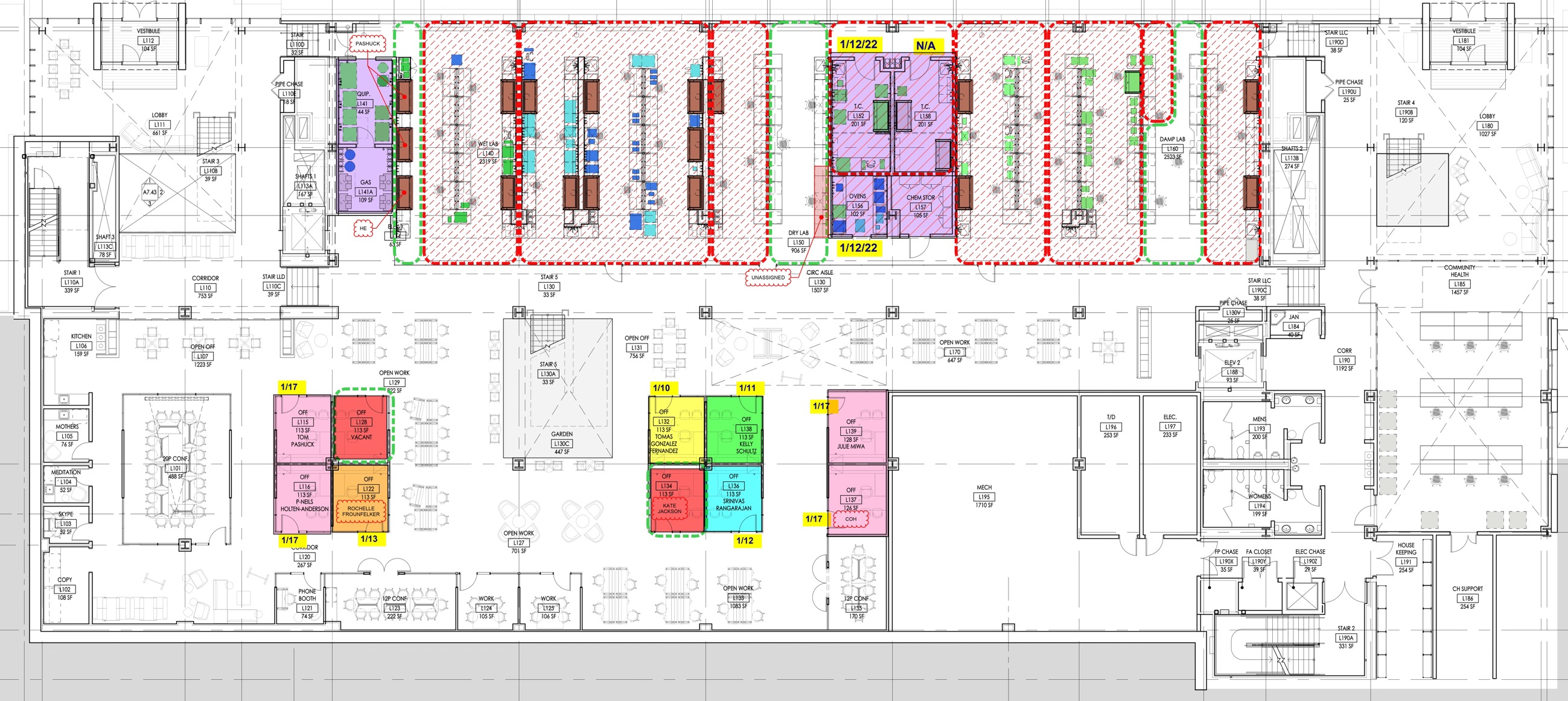}
        \caption{Lower Level}
    \end{subfigure} \\
    \begin{subfigure}[b]{0.42\linewidth}
        \includegraphics[width=\linewidth]{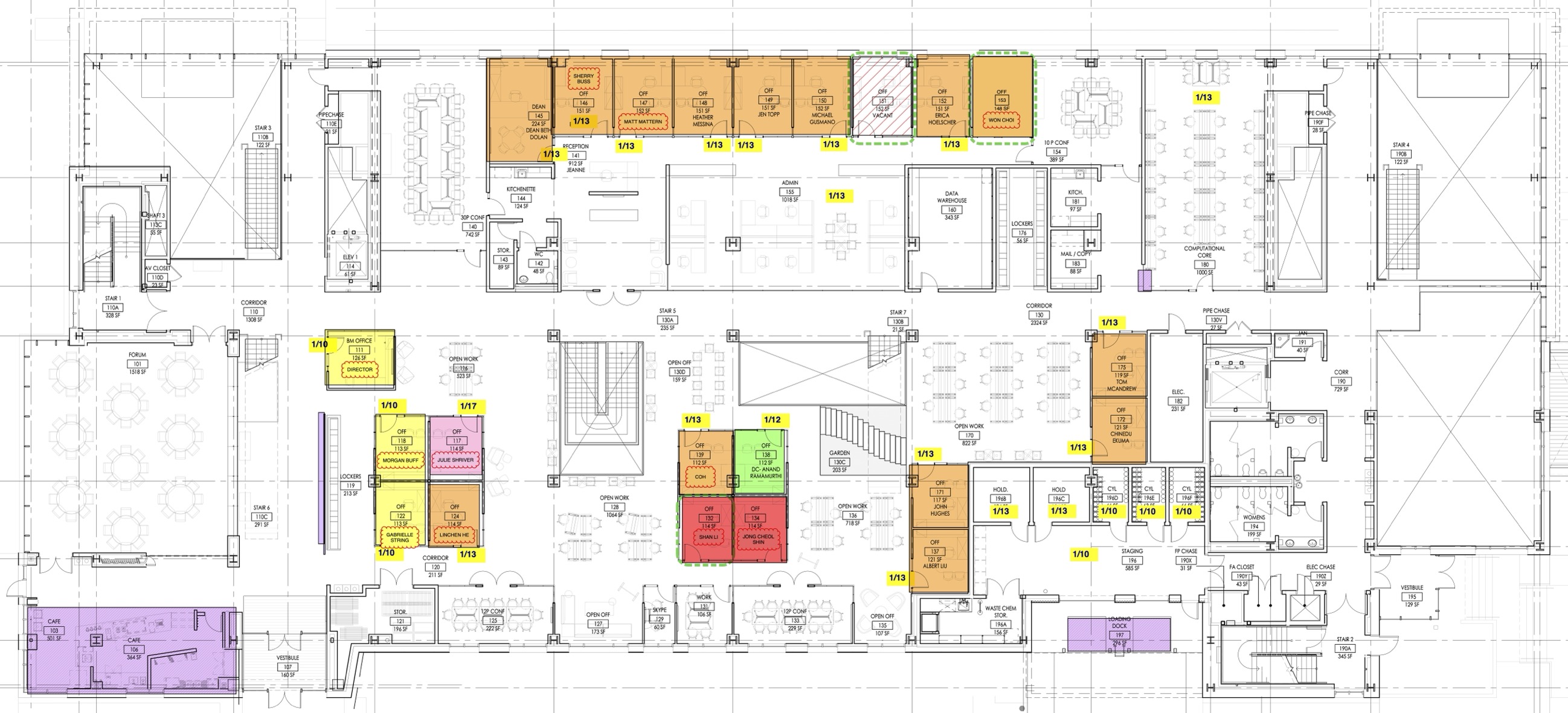}
        \caption{Level 1}
    \end{subfigure} \quad
    \begin{subfigure}[b]{0.42\linewidth}
        \includegraphics[width=\linewidth]{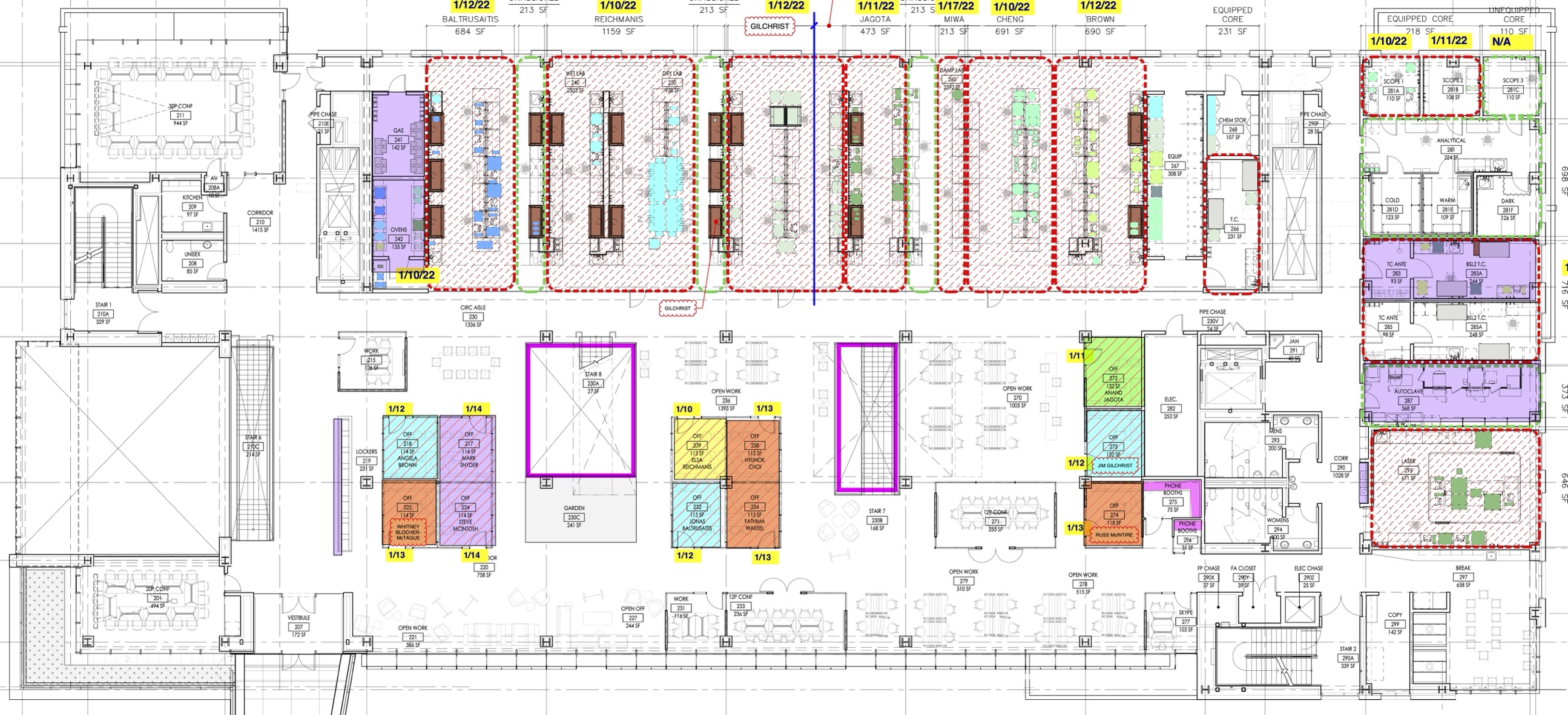}
        \caption{Level 2}
    \end{subfigure}
    \caption{Images of Floor Plan across four floors of HST.}
    \label{fig:floor_plan}
\end{figure}

We name these image frames ``anchor frames'' or ``anchor images'' throughout this paper. The origins of their coordinates are aligned to the top-left pixel of the floor plan images, as shown in Figure~\ref{fig:floor_plan}. The second and third columns above indicate the number of pixels from the origin in terms of height and width, respectively.

For detailed instructions on how to geo-register the entire path using annotated images, please refer to the official tutorial
\footnote{\url{https://colmap.github.io/faq.html\#geo-registration}}.

We stored the geo-registered model in the folder \texttt{20231220\_141254\_proj/sparse/geo}. Since the camera poses represent the coordinates of the world relative to the camera center and include redundant feature point information, we consolidated and streamlined this data from \texttt{../sparse/geo/images.bin} into \texttt{../camera2world\_6DoF.txt}.

For each floor, we have released data in standard formats, \texttt{image\_train\_all.txt} and \texttt{geometric\_data.pkl}, to serve as deep learning training inputs. The first four lines of \texttt{image\_train\_all.txt} are displayed as

\begin{applebox}{Data Format of \texttt{image\_train\_all.txt}}
Lehigh Health Science and Technology (HST) Building (https://www2.lehigh.edu/news/new-health-science-and-technology-building-a-hub-for-interdisciplinary-research).

IMG\_PATH, IMG\_ID, QW, QX, QY, QZ, TX, TY, TZ

20231220\_141254\_proj/HAND\_20231220\_141254/HAND\_20231220\_141254\_frame\_153.4s.jpg,  541, 0.46282440, -0.48897273, 0.52665017, -0.51897865, 602.36415529, 2137.65949852, -0.94367326

20231220\_141254\_proj/HAND\_20231220\_141254/HAND\_20231220\_141254\_frame\_153.1s.jpg,  540, 0.46784778, -0.48610208, 0.51996406, -0.52388988, 592.78433312, 2137.81172284, -1.24991543
\end{applebox}

This training format, inherited from PoseNet~\cite{kendall2015posenet}, aligns with the prevailing styles used in deep APR training~\cite{kendall2016modelling, kendall2017geometric, walch2017image, melekhov2017image, cai2019hybrid, chen2021direct, brahmbhatt2018geometry, shavit2021learning, shavit2022camera, chen2022dfnet, chen2023refinement}.

The file \texttt{geometric\_data.pkl} is stored as a python set in the format of

\begin{applebox}{Data Format of \texttt{geometric\_data.pkl}}
\dirtree{%
    .1 Basement/.
    .2 "train":[].
    .3 0:\{"image\_path":,"w\_t\_c":,"c\_q\_w":,"c\_R\_w":,"w\_P":,"c\_p":,"K":\}.
    .3 1:\{"image\_path":,"w\_t\_c":,"c\_q\_w":,"c\_R\_w":,"w\_P":,"c\_p":,"K":\}.
    .3 ....
    .3 39842:\{"image\_path":,"w\_t\_c":,"c\_q\_w":,"c\_R\_w":,"w\_P":,"c\_p":,"K":\}.
    .2 "test":[].
    .3 0:\{"image\_path":,"w\_t\_c":,"c\_q\_w":,"c\_R\_w":,"w\_P":,"c\_p":,"K":\}.
    .3 1:\{"image\_path":,"w\_t\_c":,"c\_q\_w":,"c\_R\_w":,"w\_P":,"c\_p":,"K":\}.
    .3 ....
    .3 16180:\{"image\_path":,"w\_t\_c":,"c\_q\_w":,"c\_R\_w":,"w\_P":,"c\_p":,"K":\}.
}
\end{applebox}

We encourage readers to review the code file at \url{https://github.com/junfish/VIP\_Navi/blob/master/dataset\_utils/extract\_geometric\_data.py} for a detailed understanding and generate \texttt{geometric\_data.pkl} on your own. This data file can be utilized for the training of camera geometric data, e.g., Geometric PoseNet~\cite{kendall2017geometric, pepe2023cga, pepe2024cgaposenet+}.

\begin{figure}[h]
    \centering
    \begin{subfigure}[b]{0.45\textwidth}
        \centering
        \includegraphics[width=\textwidth]{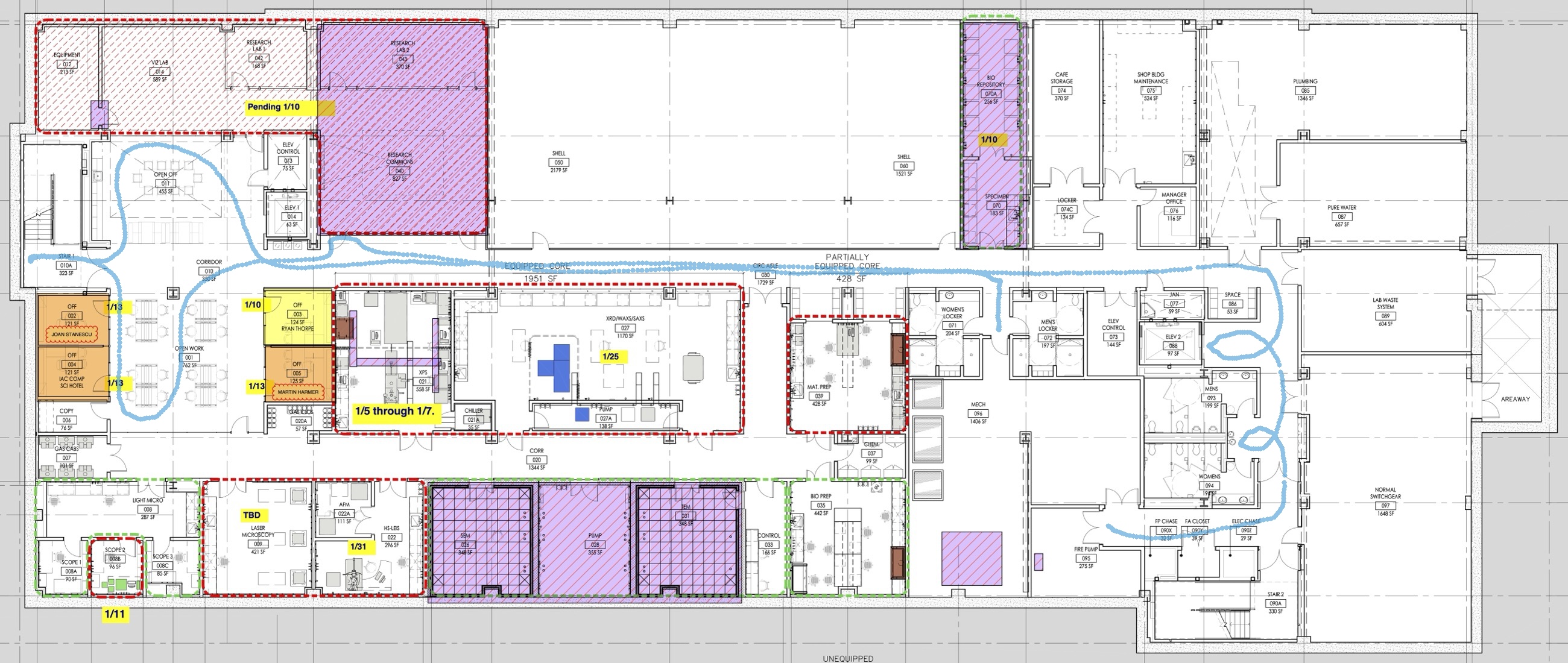}
        \caption{Basement}
        \label{basement-path}
    \end{subfigure}
    \hfill
    \begin{subfigure}[b]{0.45\textwidth}
        \centering
        \includegraphics[width=\textwidth]{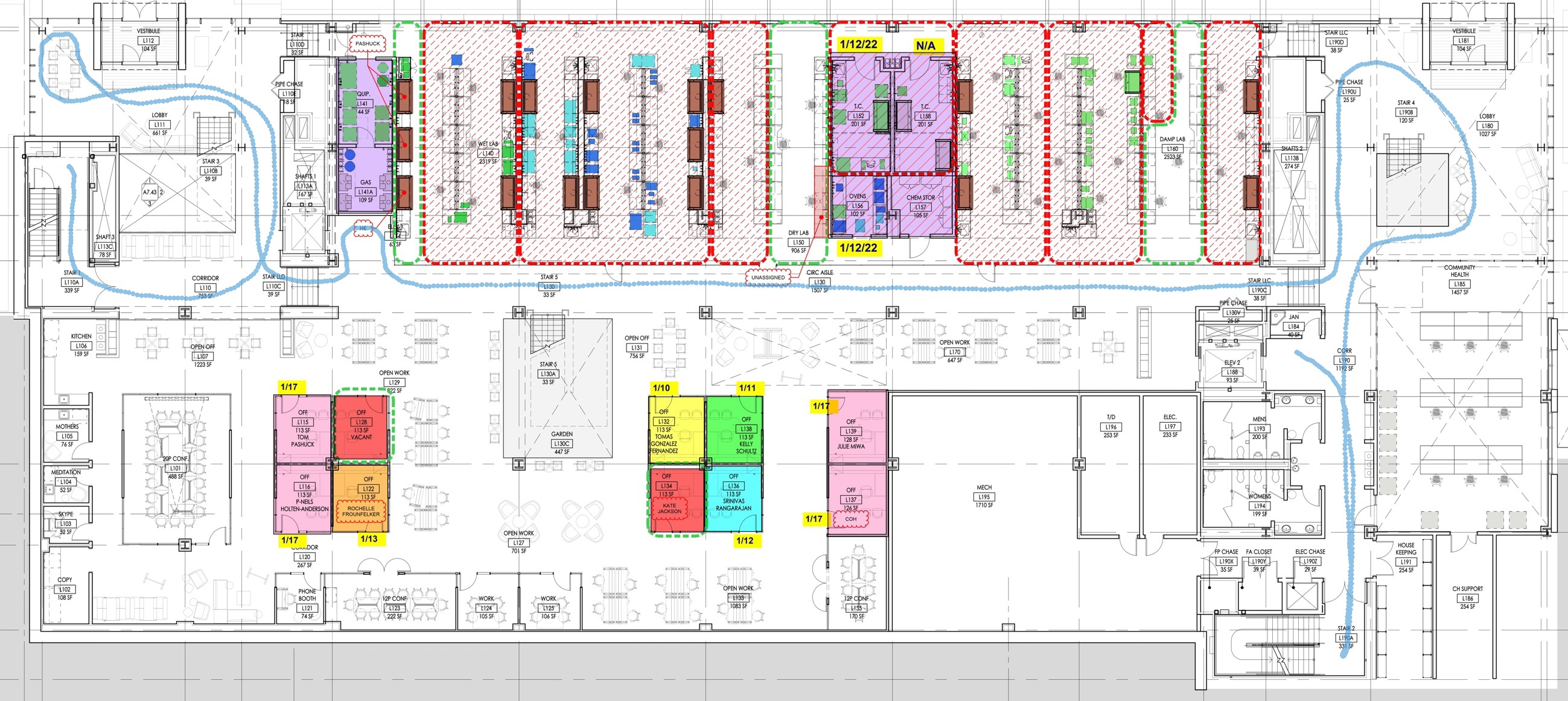}
        \caption{Lower Level}
        \label{lower-path}
    \end{subfigure}

    \begin{subfigure}[b]{0.45\textwidth}
        \centering
        \includegraphics[width=\textwidth]{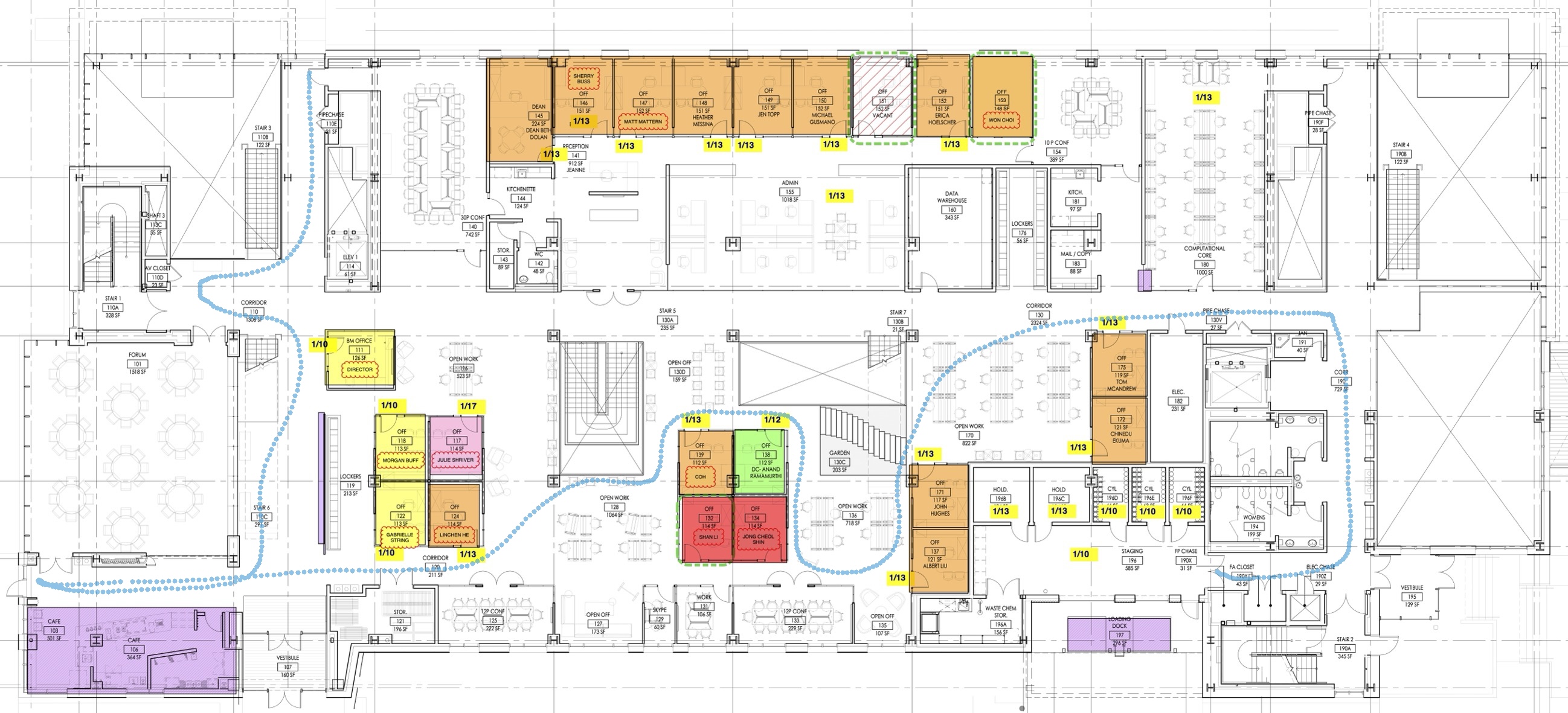}
        \caption{Level 1}
        \label{level-1-path}
    \end{subfigure}
    \hfill
    \begin{subfigure}[b]{0.45\textwidth}
        \centering
        \includegraphics[width=\textwidth]{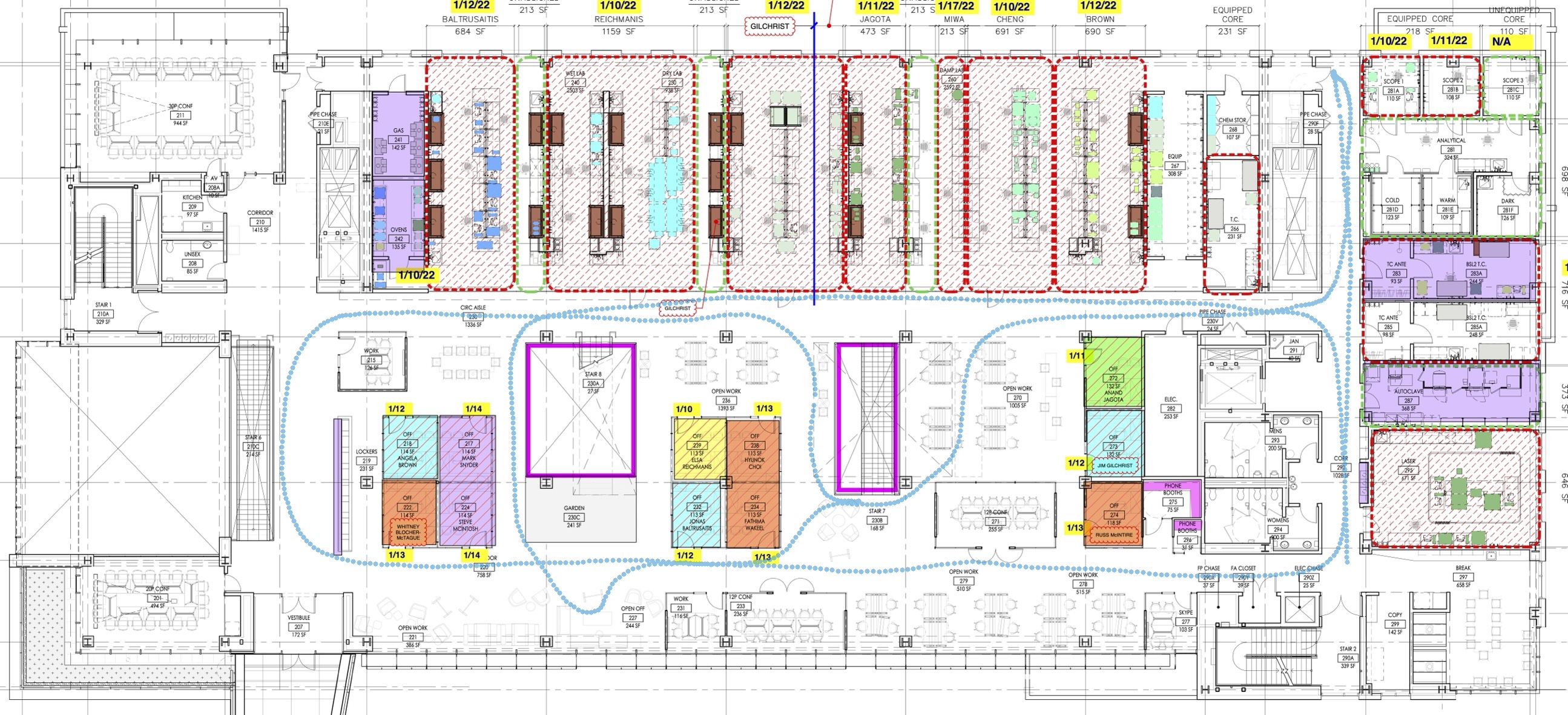}
        \caption{Level 2}
        \label{level-2-path}
    \end{subfigure}

    \begin{subfigure}[b]{0.34\textwidth}
        \centering
        \includegraphics[width=\textwidth]{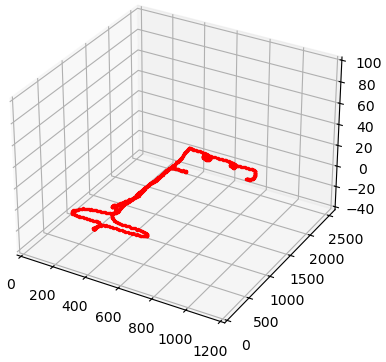}
        \caption{Basement}
        \label{basement-path-stem}
    \end{subfigure} \quad\quad\quad\quad\quad\quad\quad\quad
    \begin{subfigure}[b]{0.34\textwidth}
        \centering
        \includegraphics[width=\textwidth]{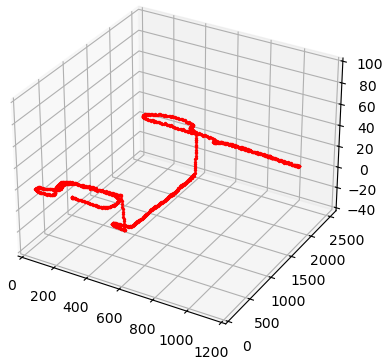}
        \caption{Lower Level}
        \label{lower-path-stem}
    \end{subfigure}

    \begin{subfigure}[b]{0.34\textwidth}
        \centering
        \includegraphics[width=\textwidth]{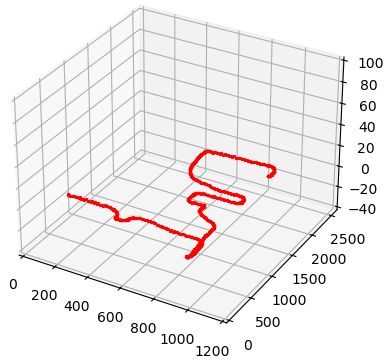}
        \caption{Level 1}
        \label{level-1-path-stem}
    \end{subfigure} \quad\quad\quad\quad\quad\quad\quad\quad
    \begin{subfigure}[b]{0.34\textwidth}
        \centering
        \includegraphics[width=\textwidth]{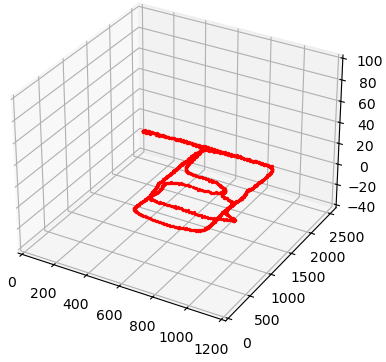}
        \caption{Level 2}
        \label{level-2-path-stem}
    \end{subfigure}

    \caption{Examples of path visualizations on the floor plan and in 3D world coordinates across four floors.}
    \label{fig:path-visualization}
\end{figure}

As illustrated in Figure~\ref{fig:path-visualization}(a-d) and \ref{fig:path-visualization}(e-h), the files \texttt{path.png} and \texttt{path\_stem.png} demonstrate the geo-registration on the floor plan and in the 3D world coordinates, respectively, to validate their accuracy. All the codes for the aforementioned steps are publicly available at \url{https://github.com/junfish/VIP\_Navi/tree/master/dataset\_utils}. 

\subsubsection{GPT-4 Captions for VIPs}
We have eight \texttt{.csv} files below to store the pseudo ground truth produced by GPT-4.

\begin{applebox}{Captions for VIPs}

1. train\_Basement\_output.csv

2. train\_Lower\_Level\_output.csv

3. train\_Level\_1\_output.csv

4. train\_Level\_2\_output.csv

5. test\_Basement\_output.csv

6. test\_Lower\_Level\_output.csv

7. test\_Level\_1\_output.csv

8. test\_Level\_2\_output.csv

\end{applebox}

Each file above contains three columns, labeled \texttt{image\_id}, \texttt{image\_file}, and \texttt{caption}. 
\subsection{Examples}
In this section, we first present examples of images from our dataset along with the reference objects used for the accurate human annotations of anchor frames. Subsequently, we randomly select images with captions for VIPs.

\subsubsection{Image Frames}
\begin{figure}[h]
    \centering
    \includegraphics[width=0.95\linewidth, height=1.25\linewidth]{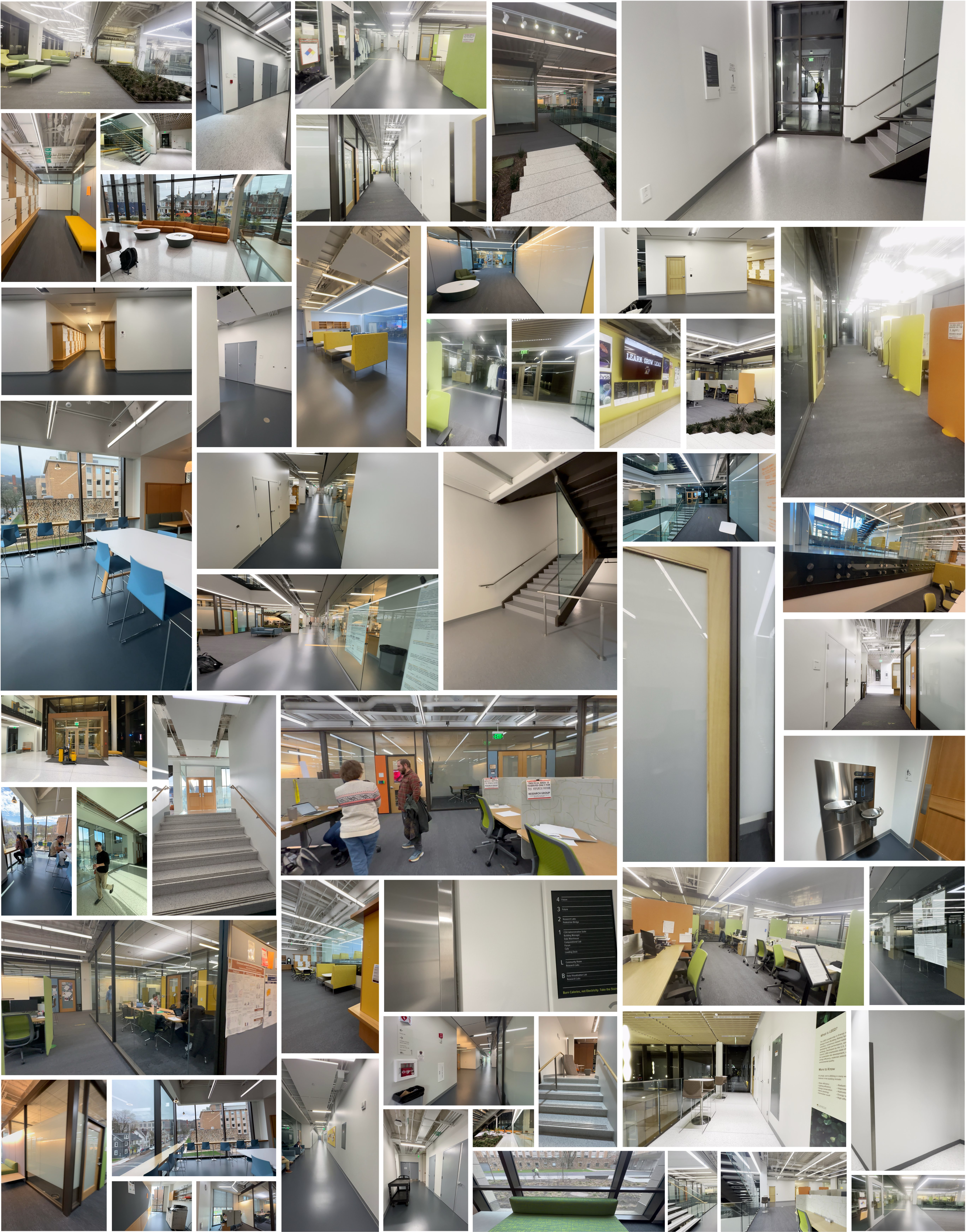}
    \caption{Randomly selected pictures extracted from different 4K video recordings. These videos are captured by various mobile devices using either human hands or a DJI stabilizer, including both portrait and landscape orientations.}
    \label{fig:frame_example}
\end{figure}

Figure~\ref{fig:frame_example} presents examples of images extracted from various videos recorded to develop this image-centric indoor navigation solution. We observe that some indoor environments are occasionally textureless and dynamic.

\subsubsection{Reference Objects}
\begin{figure}[]
    \centering
    \begin{subfigure}[b]{0.45\textwidth}
        \centering
        \includegraphics[width=\textwidth]{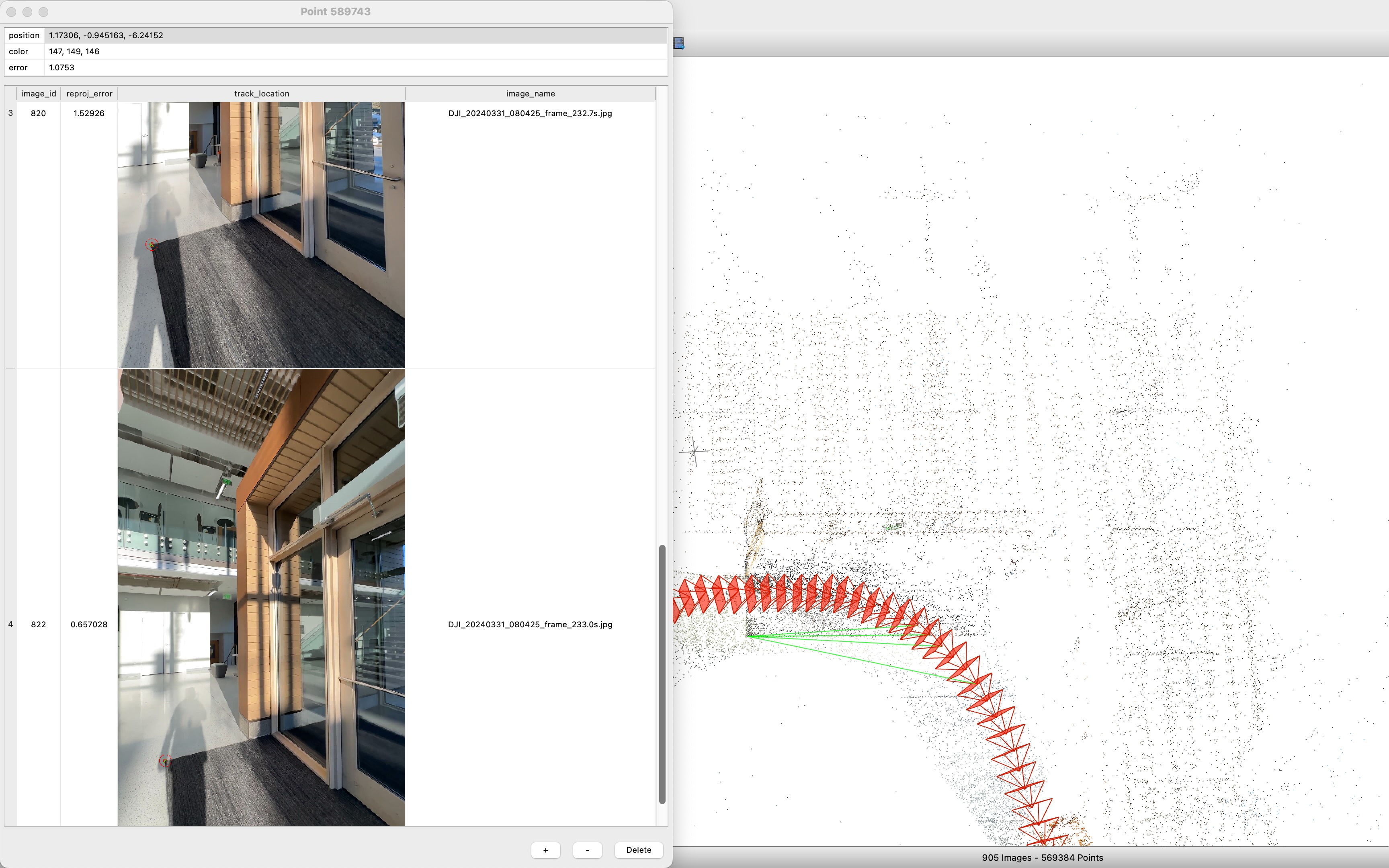}
        \caption{Carpet Corner}
    \end{subfigure}
    \hfill
    \begin{subfigure}[b]{0.45\textwidth}
        \centering
        \includegraphics[width=\textwidth]{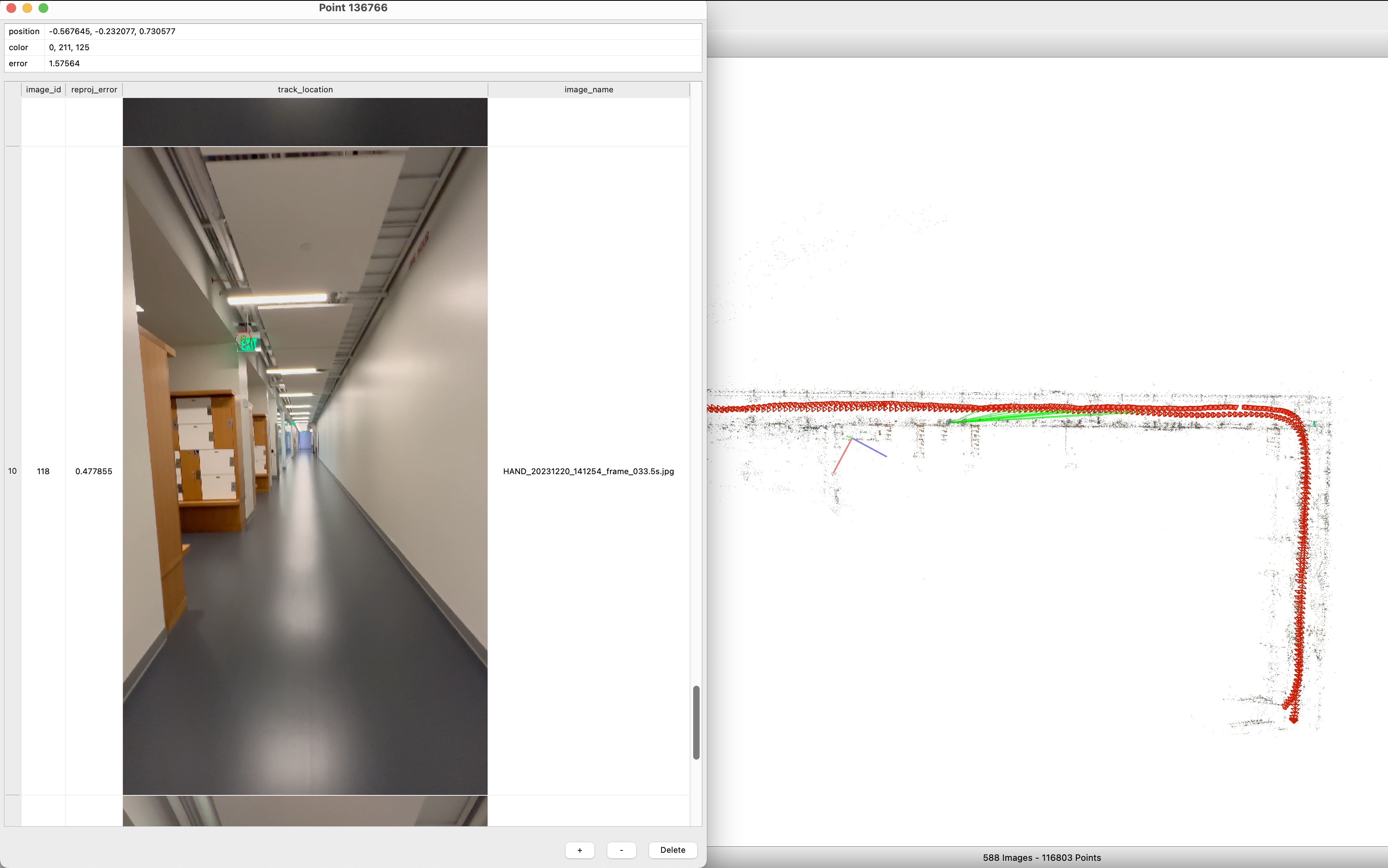}
        \caption{Exit Sign}
    \end{subfigure}

    \begin{subfigure}[b]{0.45\textwidth}
        \centering
        \includegraphics[width=\textwidth]{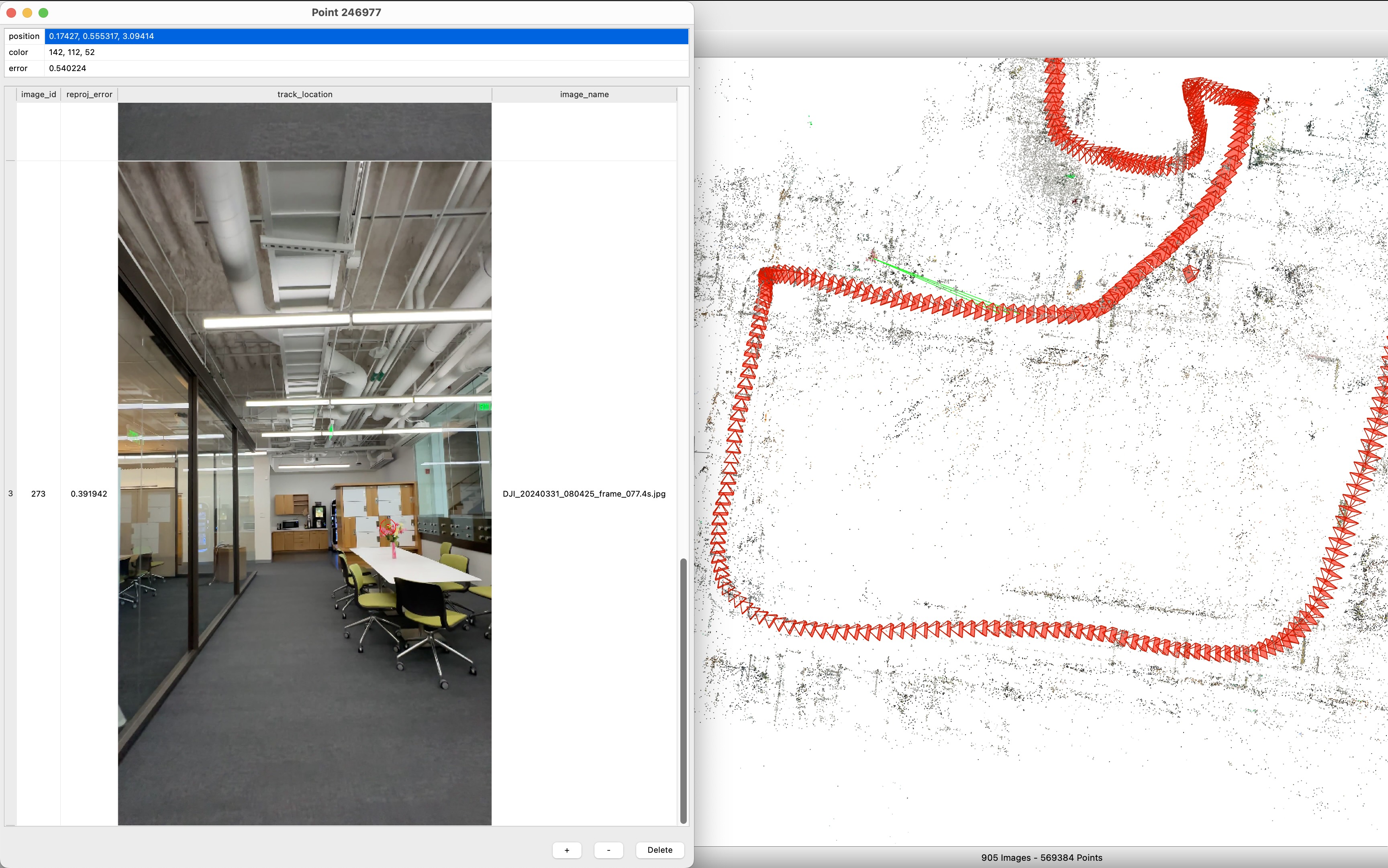}
        \caption{Flower}
    \end{subfigure}
    \hfill
    \begin{subfigure}[b]{0.45\textwidth}
        \centering
        \includegraphics[width=\textwidth]{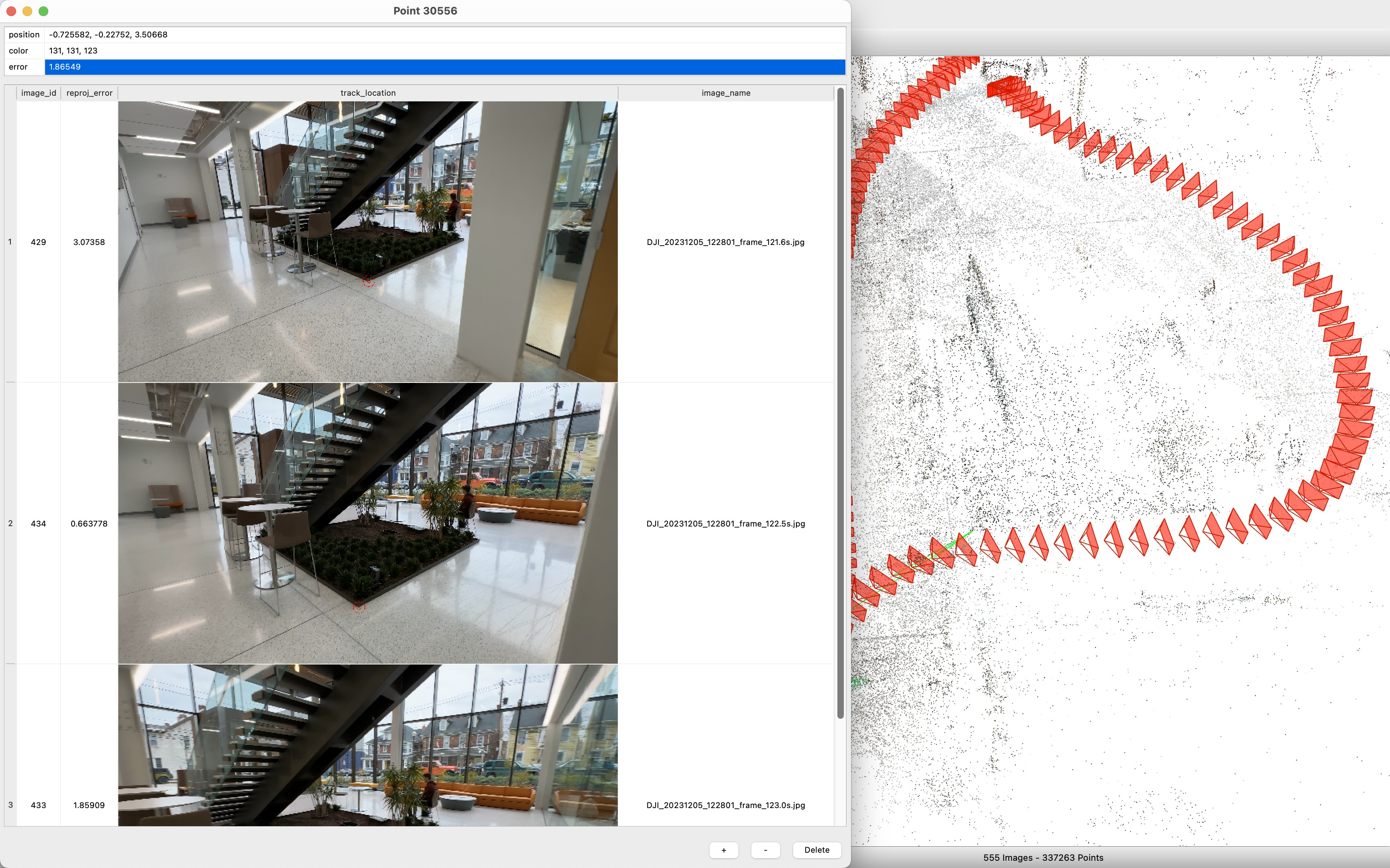}
        \caption{Garden Corner}
    \end{subfigure}

    \begin{subfigure}[b]{0.45\textwidth}
        \centering
        \includegraphics[width=\textwidth]{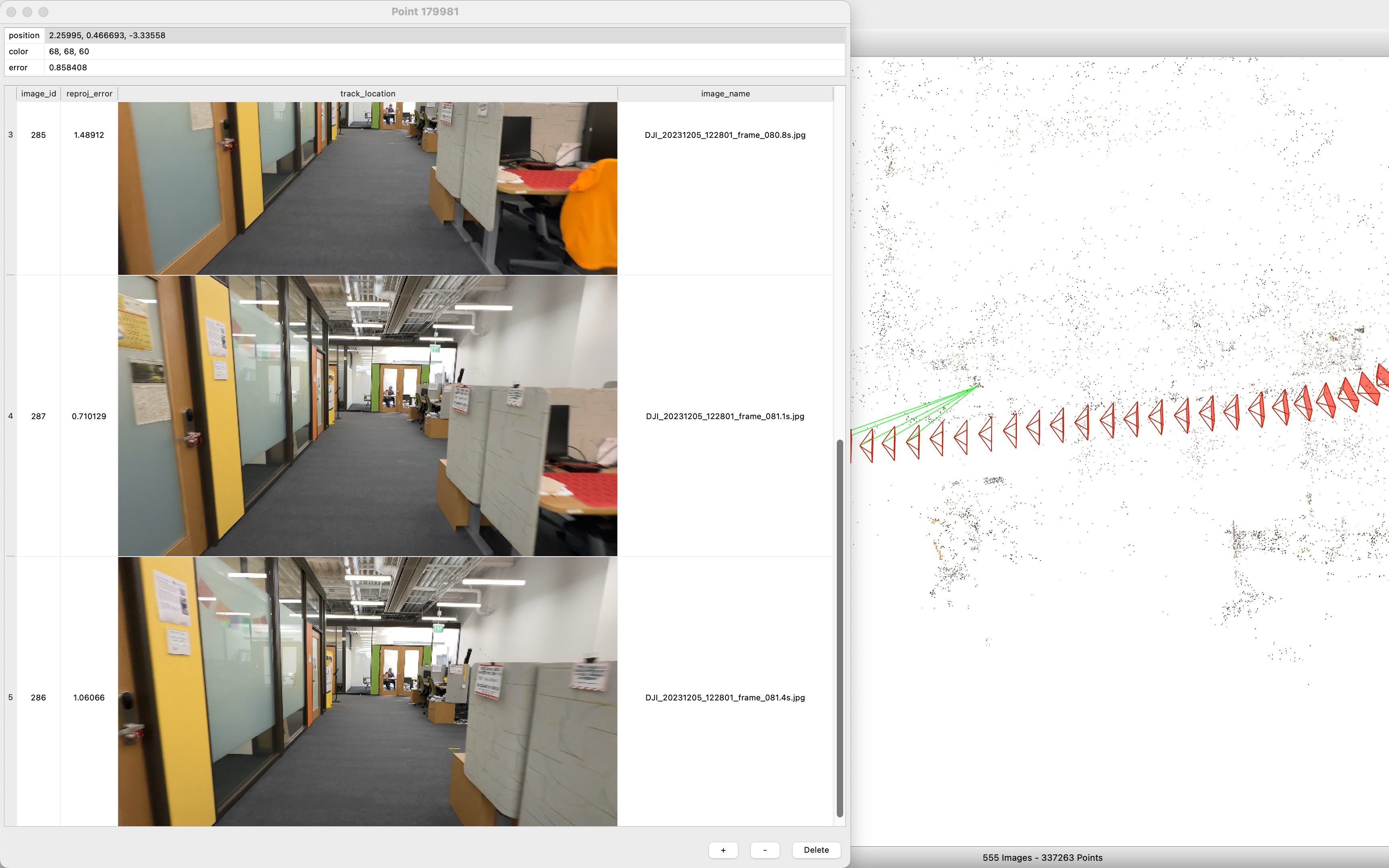}
        \caption{Lock Handle}
    \end{subfigure}
    \hfill
    \begin{subfigure}[b]{0.45\textwidth}
        \centering
        \includegraphics[width=\textwidth]{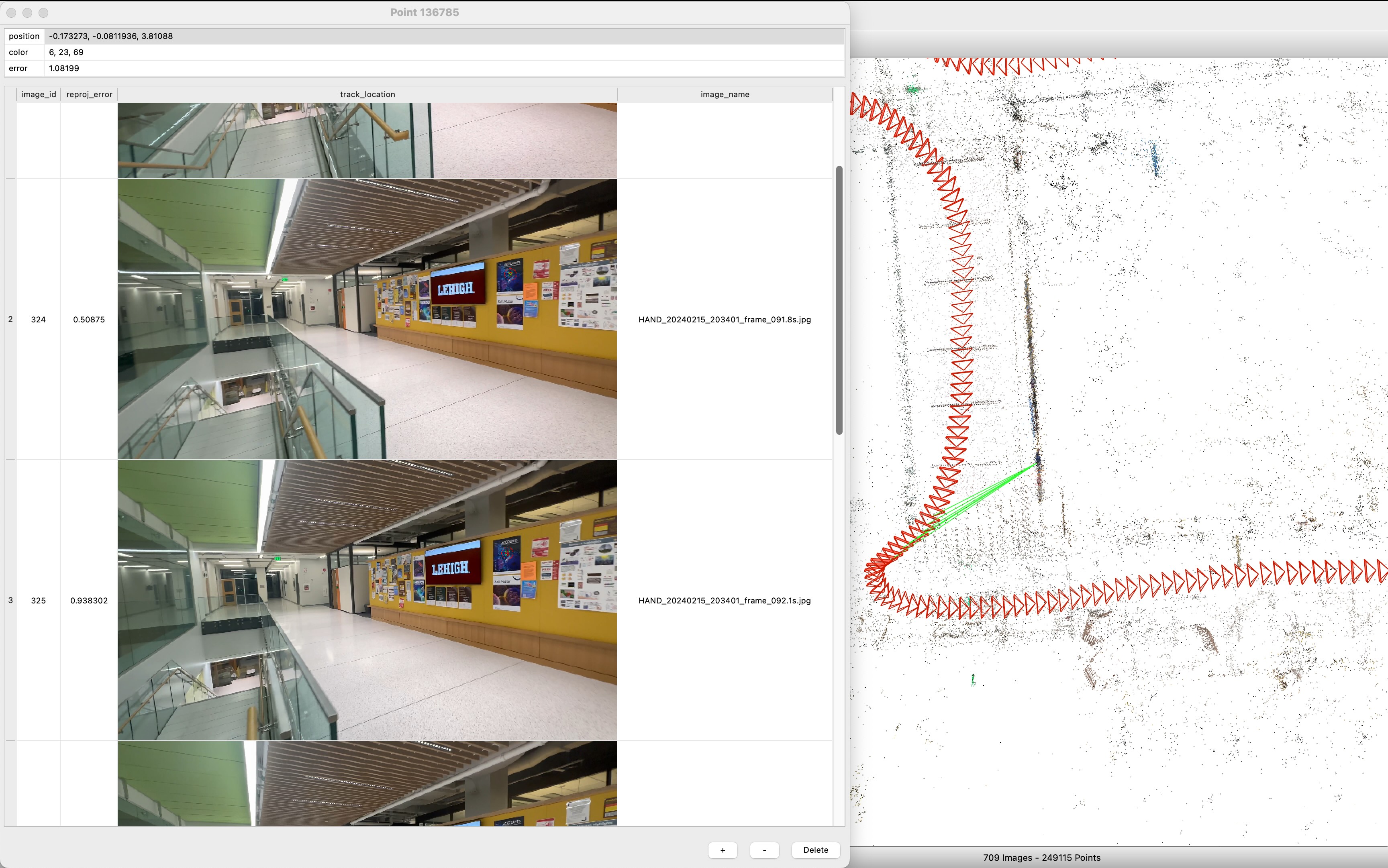}
        \caption{Poster}
    \end{subfigure}

    \begin{subfigure}[b]{0.45\textwidth}
        \centering
        \includegraphics[width=\textwidth]{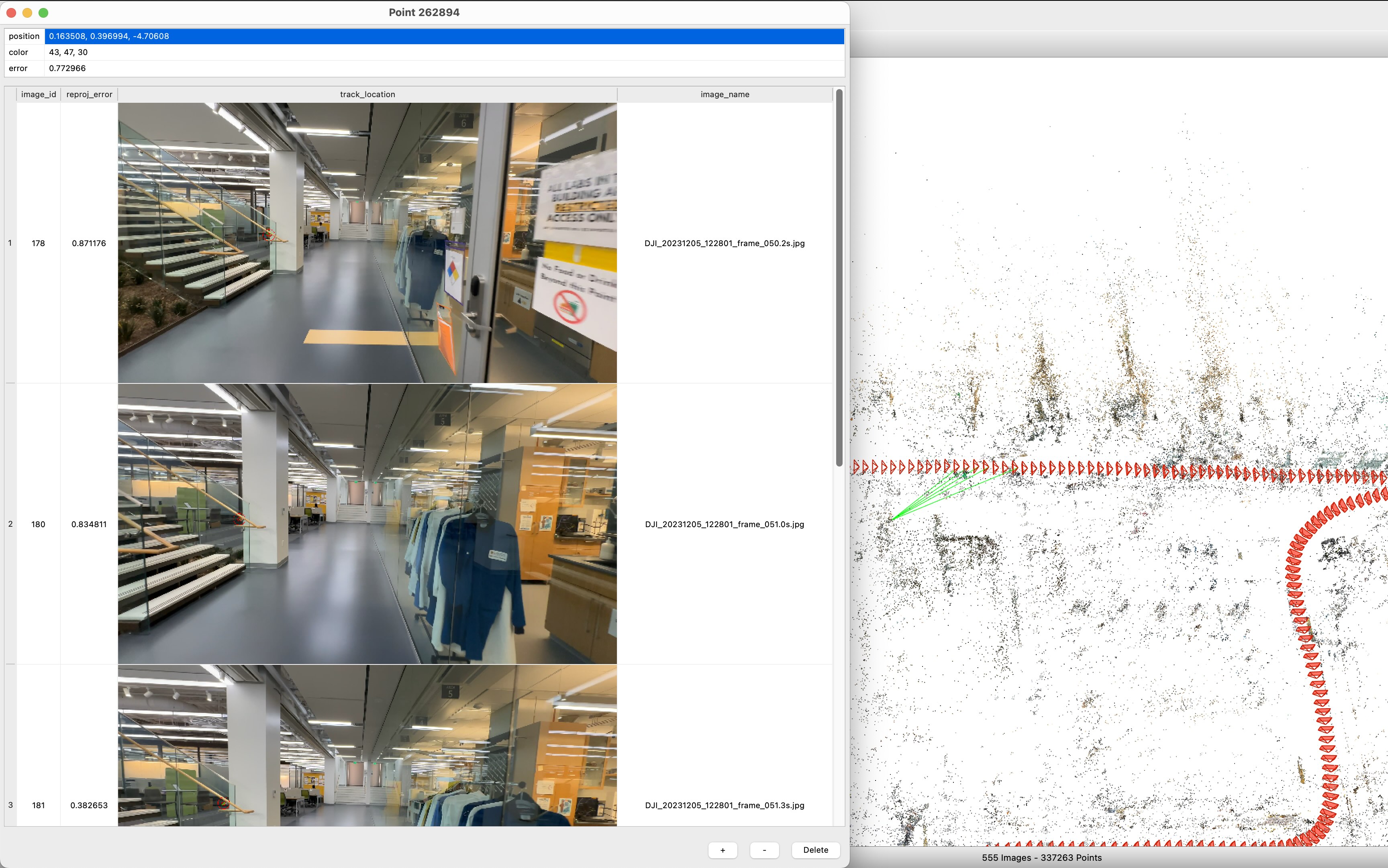}
        \caption{Banister}
    \end{subfigure}
    \hfill
    \begin{subfigure}[b]{0.45\textwidth}
        \centering
        \includegraphics[width=\textwidth]{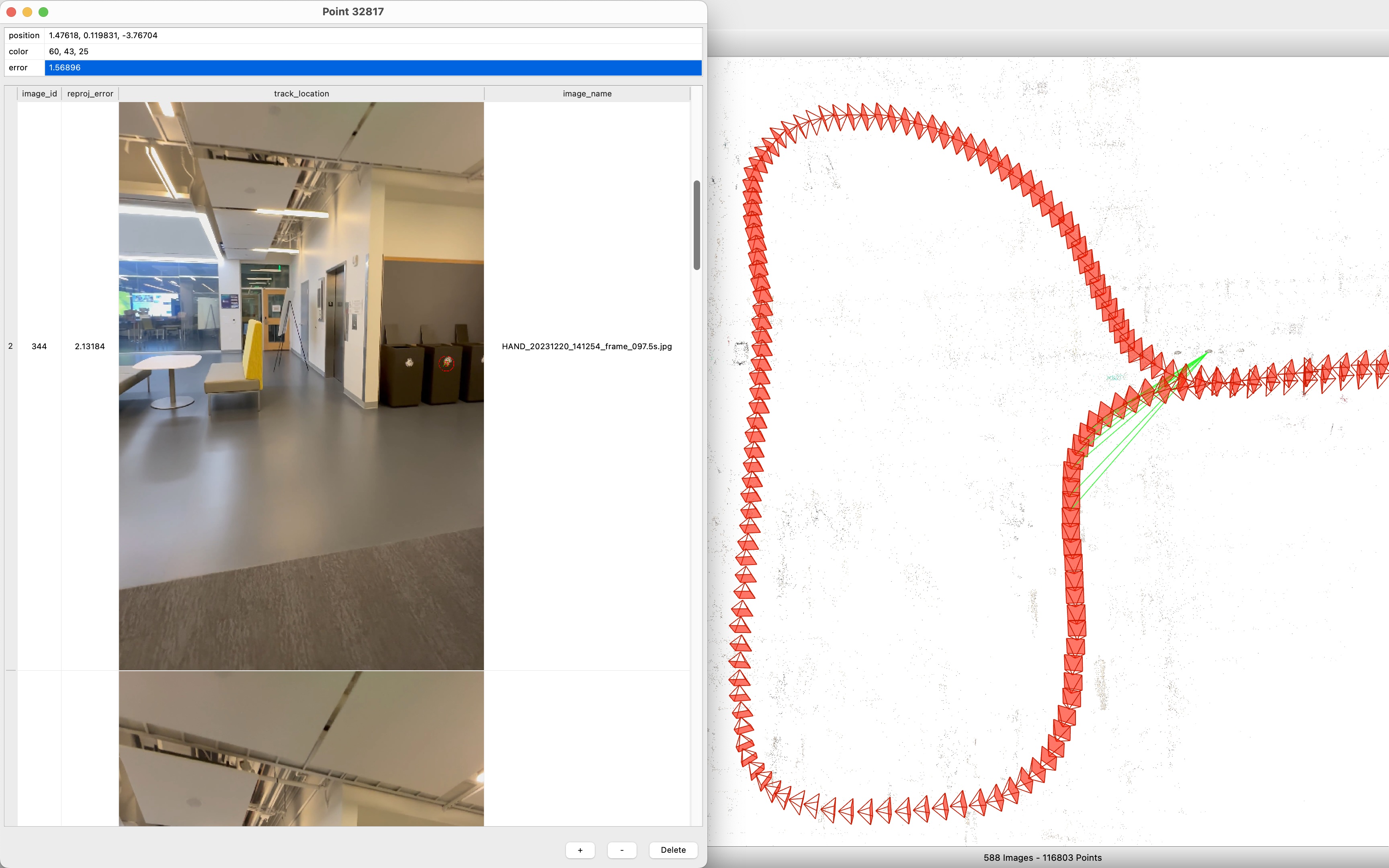}
        \caption{Trash Bin}
    \end{subfigure}

    \caption{Examples of reference objects used to accurately annotate the anchor frames along the path. Multiple reference objects can be selected collaboratively to cross-validate the ground truth positions on the floor plan.}
    \label{fig:reference_object}
\end{figure}

To obtain accurate camera poses for the anchor frames used in geo-registrations, we identify and use prominent objects in the environment that are integrated into the sparse model of COLMAP 3D reconstructions. These objects include exit signs, stairs, gardens, walls with posters, pillars, trash bins, door frames, lockers, water dispensers, etc. Figure~\ref{fig:reference_object} displays selected examples of the mapping of these feature points from the images to the 3D sparse point cloud.

\clearpage
\subsubsection{Captions by GPT-4}
We randomly show three examples from our dataset below.
\begin{applebox}{captions by GPT-4}
\begin{minipage}[t]{0.27\linewidth}
    \vspace*{1pt}
         \includegraphics[height=5.5\linewidth]{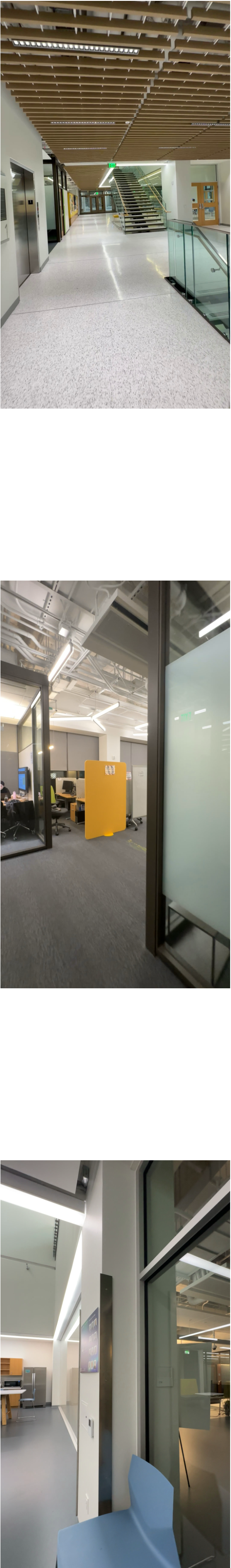}
\end{minipage}
\begin{minipage}[t]{0.7\linewidth}
\small
\textcolor{black}{\textbf{Concise Version}}:
\tiny
    \begin{itemize}[leftmargin=*]
        \item A modern hallway with stairs and elevators, featuring wooden ceiling slats, clean floors, and glass railings.
    \end{itemize}
\vspace{0.1cm}

\small   
\textcolor{black}{\textbf{Detailed Version}}:
\vspace{0.2cm}

\small
\colorbox{mygreen}{\textbf{For Early-Blind}} 
\tiny
    \begin{itemize}[leftmargin=*]
        \item You are in a spacious, modern hallway, about 6 feet wide. The floor is smooth with a slight speckled texture, providing a comfortable walking surface. Overhead, there are evenly spaced wooden slats along the ceiling, offering a unique spatial pattern. At 9 o'clock, about 6 feet to your left, is an elevator integrated into the hallway wall; it has a flat surface and poses no tripping risks. Straight ahead, the hallway lengthens around 20 feet. To the right, at 3 o'clock and roughly 10 feet from your current position, there is a broad staircase featuring glass railings. This staircase leads to the upper floor. The path in the hallway is clear of obstructions, making for a safe, unobstructed journey. Further down, past the stairs, the corridor proceeds and is lined with glass doors and other branching hallways.
    \end{itemize}
    
\small
\colorbox{myorange}{\textbf{For Low-Vision \& Late-Blind}} 
\tiny
    \begin{itemize}[leftmargin=*]
        \item You are in a modern hallway, approximately 6 feet wide. The floor is polished and smooth, with a slight speckled texture that is easy to walk on. Overhead, there are uniform wooden slats on the ceiling running parallel to your path, providing an aesthetic touch. On your left at 9 o'clock, there's an elevator with metallic doors set into a plain wall about 6 feet away. It's flush with the hallway wall, without any tripping hazards. In front of you, the corridor stretches about 20 feet before widening slightly. To your right, at 3 o'clock and about 10 feet away, a wide, open staircase with transparent glass railings ascends to the next floor. There are no visible obstacles on the floor, ensuring an easy and safe walk. Farther ahead, beyond the stairs, the hallway continues, flanked by glass doors and additional hallways.
    \end{itemize}
\vspace{0.5cm}
\small
\textcolor{black}{\textbf{Concise Version}}:
\tiny
    \begin{itemize}[leftmargin=*]
        \item Office with a partition, workstations, and two people, visible furniture, and lighting.
    \end{itemize}
\vspace{0.1cm}

\small   
\textcolor{black}{\textbf{Detailed Version}}:
\vspace{0.2cm}

\small
\colorbox{mygreen}{\textbf{For Early-Blind}} 
\tiny
    \begin{itemize}[leftmargin=*]
        \item You are in a busy office environment. Directly to your left at 9 o'clock, two people are engaged with their computer monitors in a glass cubicle. Ahead, you'll encounter a substantial partition around 10 feet from you. This partition functions as a spatial divider unique to this setting. Walking past the partition, there are additional work areas equipped with desks, chairs, and computer systems. Overhead, the office has ceiling fixtures that likely create a well-lit environment. To your right, at 3 o'clock, there's a wall made of frosted glass, part of the office’s contemporary design. Exercise caution as you proceed: there are chair legs by the pathway which may pose a slight tripping hazard. To safely reach the next workstation, steer slightly to your right to maintain a clear pathway free of obstacles.
    \end{itemize}
    
\small
\colorbox{myorange}{\textbf{For Low-Vision \& Late-Blind}} 
\tiny
    \begin{itemize}[leftmargin=*]
        \item You are in a modern office space. To your immediate left at 9 o'clock, there are two people sitting at desks with computers in a glass-walled section. Moving forward, there is a large yellow partition about 10 feet away, which stands approximately 6 feet high and 4 feet wide, acting as a divider in the middle of the room. Directly ahead, just beyond the partition, are more workstations with office chairs, desks, and computer monitors. Fluorescent lights hanging from the white ceiling brighten the room. To your right, at 3 o'clock, you can see a frosted glass wall, which is part of a modern office design. Be cautious of the chair legs close to the pathway and navigate straight or slightly to the right to avoid obstacles and reach a workstation.
    \end{itemize}

    \vspace{0.5cm}
\small
\textcolor{black}{\textbf{Concise Version}}:
\tiny
    \begin{itemize}[leftmargin=*]
        \item Office space with a kitchen, glass-enclosed room, and various furniture arranged neatly.
    \end{itemize}
\vspace{0.1cm}

\small   
\textcolor{black}{\textbf{Detailed Version}}:
\vspace{0.2cm}

\small
\colorbox{mygreen}{\textbf{For Early-Blind}} 
\tiny
    \begin{itemize}[leftmargin=*]
        \item The scene is a tidy, modern office. In front of you, a glass partition is at 3 o'clock. Immediately to your left, around two feet, is a kitchen area with a countertop, cabinets, and appliances. You can navigate safely along a path roughly three feet wide leading into the kitchen. About ten feet beyond the kitchen, there’s a metallic refrigerator. To your right, approximately four feet away, a blue chair is near the glass partition. Inside the glass-enclosed room at 3 o'clock, there's a white chair under a table. The floor is clear of obstacles, though watch for the sharp corners of the kitchen countertop to your left.
    \end{itemize}
    
\small
\colorbox{myorange}{\textbf{For Low-Vision \& Late-Blind}} 
\tiny
    \begin{itemize}[leftmargin=*]
        \item You are in a modern office space. Directly ahead, a glass wall creates a partition at 3 o'clock. To your immediate left, around two feet away, the kitchen area has a counter, appliances, and cabinets. Beyond the kitchen, approximately ten feet away, is a stainless-steel refrigerator. There's ample walking space, roughly three feet wide, leading to the kitchen. At the entrance to your right, around four feet away, a blue chair is securely placed near the glass wall. As you look further right at 3 o'clock, there is a white chair positioned under a table inside the glass-enclosed room. There are no visible hazards; the floor appears clear, but be mindful of the sharp corners of the countertop to your left.
    \end{itemize}

\end{minipage}
\end{applebox}

\subsection{Statistics}
\begin{figure}[h]
    \centering
    \begin{subfigure}[b]{0.64\textwidth}
        \centering
        \includegraphics[width=\textwidth]{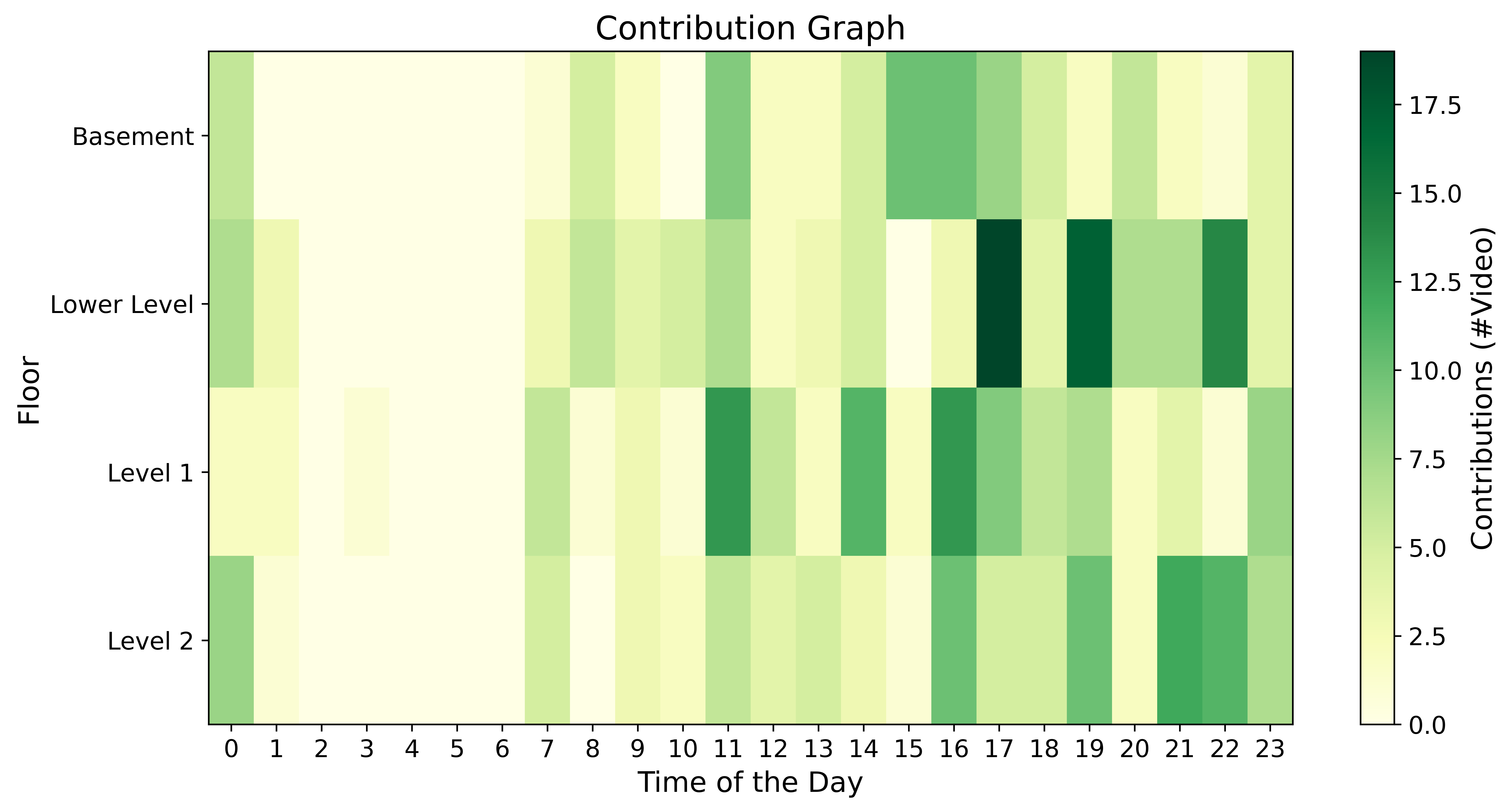}
        \caption{Distribution of video contributions throughout daytime hours.}
    \end{subfigure}
    \hfill
    \begin{subfigure}[b]{0.3\textwidth}
        \centering
        \includegraphics[width=\textwidth]{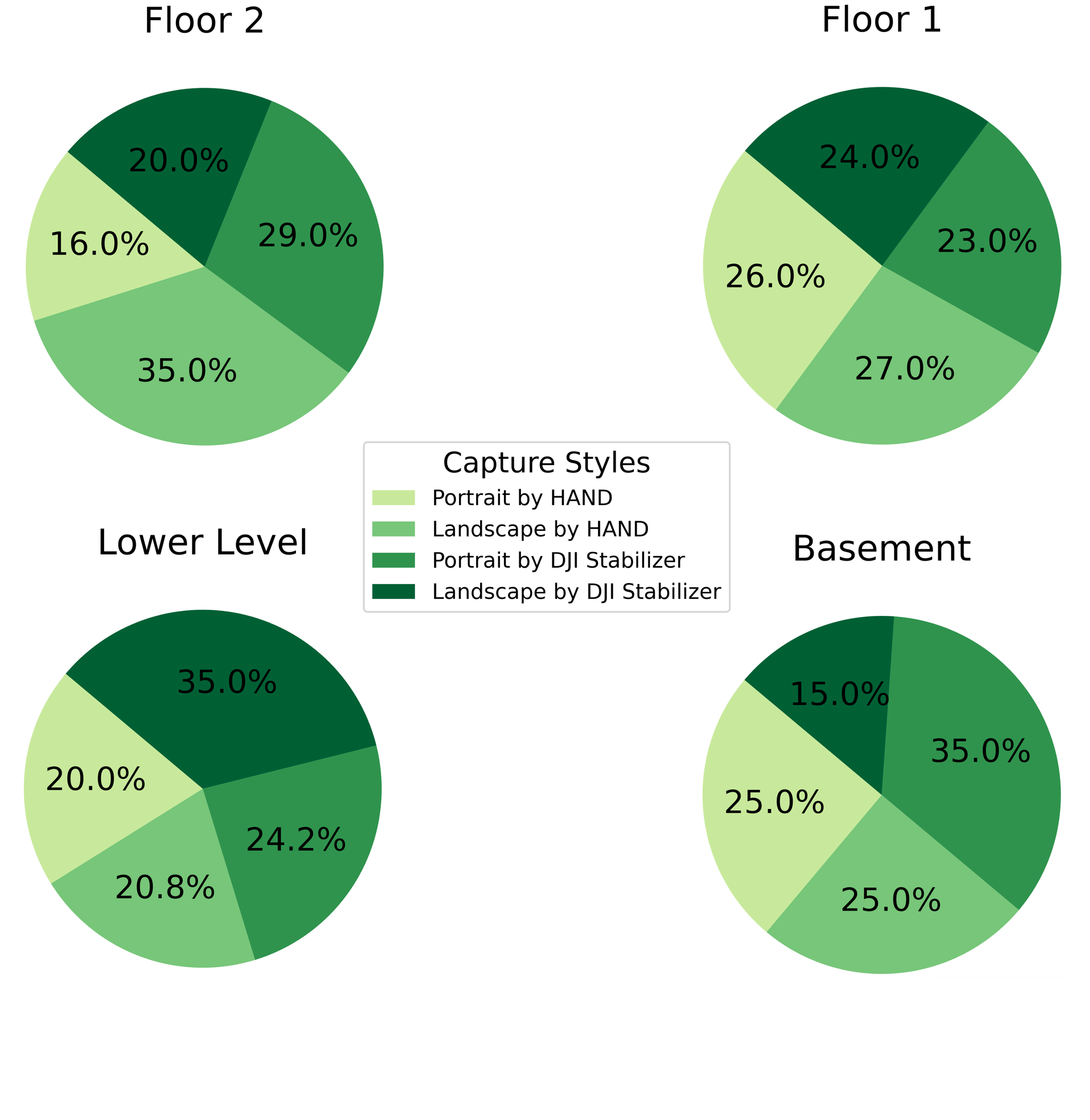}
        \caption{Proportions of video postures and styles.}
    \end{subfigure}

    \begin{subfigure}[b]{0.95\textwidth}
        \centering
        \includegraphics[width=\textwidth]{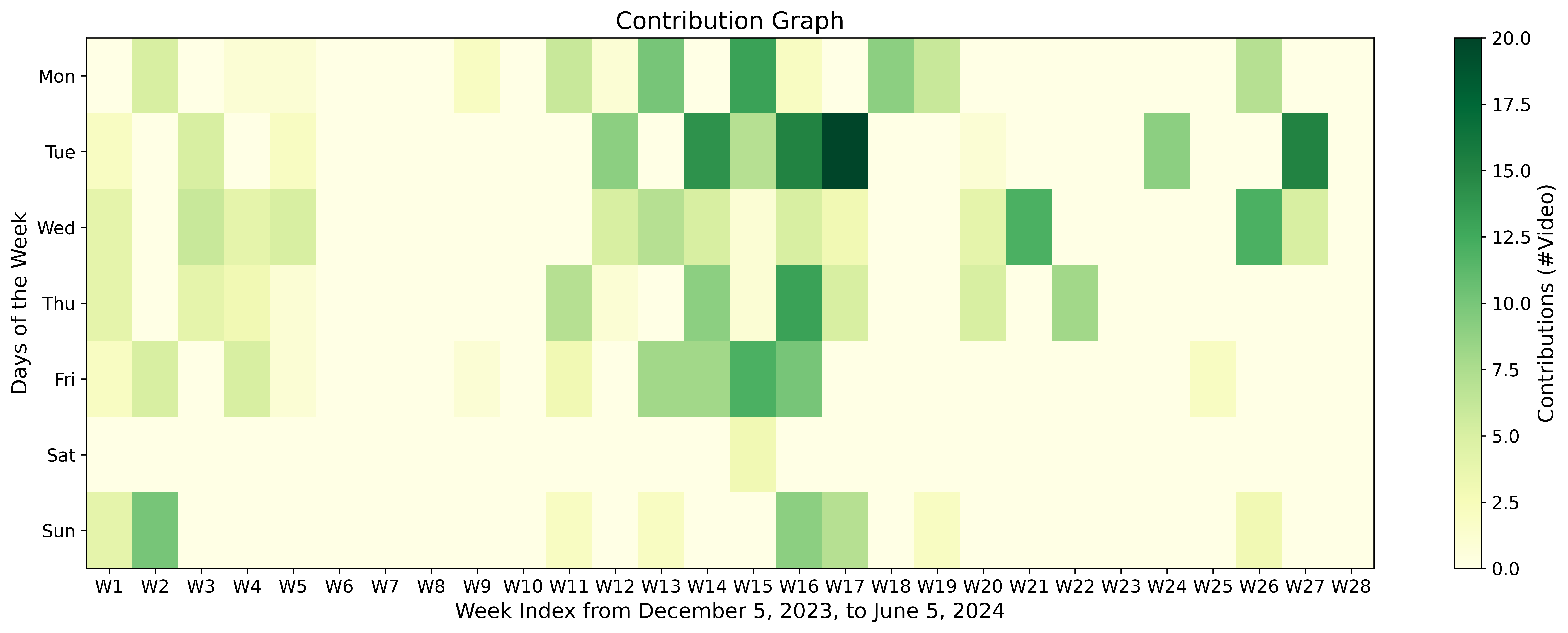}
        \caption{video contributions over weeks from December 2023 to June 2024.}
    \end{subfigure}
    \caption{Statistics of videos recorded for NaVIP. (a) Contribution Graph depicting the distribution of video recordings across different times of the day for each floor. The heatmap illustrates variations in the number of video contributions from Basement to Level 2, with darker greens indicating higher activity during specific hours. (b) Distribution of video capture styles across different floors, illustrated through pie charts. Each chart represents the proportion of video capture modes: Portrait by Hand, Landscape by Hand, Portrait by DJI Stabilizer, and Landscape by DJI Stabilizer. The charts highlight the variance in capture styles from Floor 2 to the Basement. (c) Contribution Graph showing the distribution of video recordings by days of the week over a period from December 5, 2023, to June 5, 2024. The heatmap illustrates the number of video contributions per day, highlighting activity patterns with darker shades representing higher frequencies of video recordings.}
    \label{fig:statistics}
\end{figure}

As illustrated in Figure~\ref{fig:statistics}, we have conducted basic statistical analyses of our video recordings to indicate potential biases inherent in data collected by humans. We note that our dataset lacks videos from the hours between 2:00 AM and 6:00 AM, as well as on Saturdays and during the Christmas holidays.

\section{VIP User Needs and GPT-4 Interaction}
\subsection{VIP User Needs and GPT-4 Text Prompts Design}
We aim to experiment with the use of GPT-4 for processing images captured by visually impaired individuals using their smartphones and generating controlled, effective textual descriptions to facilitate barrier-free living. 

Initially, we crafts a preliminary prompt for GPT, drawing inspiration from Jain et al.'s work \cite{jain2023want}. Jain introduces the concept of "Exploration Assistance," an evolution of Navigation Assistance Systems (NASs) that empowers VIPs to explore unfamiliar environments. They study and analyze VIP user needs, such as what information VIPs require to explore unfamiliar environments and what factors influence these needs among individuals. Inspired by Jain et al.'s work\cite{jain2023want}, we categorize visually impaired users into early-blind, late-blind, and low-vision groups. We test GPT's ability to produce descriptions at three detail levels: spatial information, independent exploration, and collaborative exploration.

We find that GPT's performance limitations prevent it from generating distinct descriptions for different user needs, leading to redundant information in multiple versions of image descriptions and diminishing user experience. To address this, we reassessed user categories and requirements and designed an improved prompt. Experiments show that this improved prompt version enhances the accuracy and readability of GPT-generated descriptions. Additionally, we observe that GPT struggles with accurately describing direction information. To mitigate this, we introduce a visual prompt that corrects these errors, providing users with more precise navigation information.

\subsubsection{Designs for GPT-4 Text Prompts: First Version}
\label{subsubsec:design_prompt_first}
Please help me to annotate images with natural language descriptions for visually impaired people (VIPs). It is important to be as descriptive and clear as possible. For each image, please give me two versions (concise \& detailed version) of descriptions for three VIP groups (VIPs who are low-vision, early-blind \& late-blind), respectively.

My requirements of two versions are:
\begin{itemize}
    \item \textbf{Concise version} contains only one sentence (no more than $12$ words). Use specific nouns and adjectives to give environment summarization, such as ``\textit{a quiet lab where some people are reading books}.''
    \item \textbf{Detailed version} contains the descriptions of three different levels, as outlined below. Each level represents a progressively higher need within the visually impaired population. Please ensure that the information for each level is contained within its own separate paragraph.
    \begin{itemize}
        \item \textbf{\textit{Level 1}. Spatial information needs (perceptual insight cravings)}. VIPs need two types of spatial information: shape information and layout information, to get a high-level overview of a space. Your descriptions should help VIPs gather shape and layout information in a manner that facilitates active engagement with the environment.
        \item \textbf{\textit{Level 2}. Independent exploration needs (self-directed learning desires)}. VIPs face difficulties in making navigation decisions based on spatial information collected via their non-visual senses and your descriptions. Additionally, acquiring appropriate orientation and mobility (O\&M) training and maintaining their O\&M skills can be a struggle for them, which negatively impacting their confidence to explore independently. You should try your best to afford VIPs precise and reliable spatial information and should ensure that VIPs make accurate navigation decisions based on this information. You can serve as an O\&M education assistance tool to give VIPs the confidence to explore unfamiliar environments independently.
        \item \textbf{\textit{Level 3}. Collaborative exploration needs (social interaction needs)}. Social pressures pose a major challenge to VIPs receiving help and to non-VIPs providing help when exploring environments collaboratively. VIPs want verbal assistance in a comprehensible format when exploring collaboratively but find it challenging to communicate this preference to others. You should normalize VIPs’ exploration behaviors by introducing social norms for exploration assistance in the current situation. If possible, you could scaffold and facilitate collaborative exploration by telling VIPs how to distribute requests for collaboration and what kind of translations may be needed for VIPs.
    \end{itemize}
\end{itemize}
Here are some tips for you to refine my requirements above:

\begin{applebox}{Tips}
    \begin{enumerate}[leftmargin=0.4cm]
    
    \item \textbf{Object Identification and Description}: Clearly identify each object in the image. Use specific nouns and adjectives to describe the objects, such as ``\textit{round red apple}'' or ``\textit{high laboratory shelves}.''
    
    \item \textbf{Spatial Relationships}: Describe the spatial relationships between objects. Use terms like ``\textit{to the left of},'' ``\textit{behind},'' ``\textit{in front of},'' ``\textit{next to},'' and ``\textit{above}'' to explain how objects are situated in relation to each other.
    
    \item \textbf{Approximate Measurements}: Provide approximate measurements to help the listener understand distances. Use units like steps, arm's length, or common objects as references, for example, ``\textit{about three steps away}'' or ``\textit{an arm's length apart}.''
    
    \item \textbf{Textures and Surfaces}: Mention the texture or surface of objects if relevant. For instance, ``\textit{the desk has a smooth, laminate surface}'' or ``\textit{the corridor is lined with polished tiles}.''
    
    \item \textbf{Colors and Contrasts}: Even though the individual may not perceive colors, describing them can help build a richer picture or convey information to those who have partial vision or can perceive some colors. Contrasts can also be useful, like ``\textit{the bright fluorescent lights against the neutral-colored ceiling}.''
    
    \item \textbf{Layout Description}: Offer a general layout of the scene. For example, ``\textit{The room is rectangular, with a door on the shorter wall and a window on the longer wall}.''
    
    \item \textbf{Directional Orientation}: Use cardinal directions if the person is familiar with them, like ``\textit{the conference room is to the east},'' or relative directions like ``\textit{the sun sets in front of you when you face the window}.''

    \item \textbf{Consistency}: Keep descriptions consistent throughout the annotations. If you start by describing objects from left to right, maintain that order. You can also start by describing the immediate area, followed by objects in the middle distance, and then far away items.


    \item \textbf{Interactive Elements}: Point out any interactive elements or objects that can be used, such as ``\textit{a button to press at the crosswalk}'' or ``\textit{to your right, there's a window that can be opened for fresh air}.''

    \item \textbf{Avoid Subjectivity}: Keep descriptions objective and avoid personal interpretations or non-evident assumptions about the scene. When uncertain about quantities or functionalities, employ phrases such as ``\textit{Some chairs feature a green accent}'' or ``\textit{these numbered, circular floor markers with arrows possibly serve for wayfinding or organizing seating}.''

    \item \textbf{Clarity and Brevity}: Be clear and concise. While detail is important, unnecessary information can clutter the mental image. For instance, instead of ``\textit{a rectangular, mahogany coffee table with intricate carvings},'' use ``\textit{a long, wooden coffee table in front of the couch}.''
    
    \item \textbf{Safety and Navigation Cues}: Highlight potential obstacles or safety hazards (like low-hanging objects, uneven surfaces) and include navigational cues if possible. Use sentences like ``\textit{Please be careful of the small step down as you enter the room. There's a low coffee table in the center, about two feet high}.'' or ``\textit{If you keep walking straight for about six steps, you'll reach the kitchen door. You can feel a wooden chair right before the entrance.}''\\
    
    \end{enumerate}
\end{applebox} 

By considering these tips, you can create descriptions in natural language that are informative and useful for building a mental map for VIPs.

Requirements of three groups of VIPs are:
\begin{itemize}
    \item \textbf{Early-blind VIPs} are a group of people who are blind by birth or developed vision impairments early in life. They have learned to trust their non-visual senses to collect spatial information over time, and have no concept or sense of colors. Please describe using texture or tactile feel, such as ``\textit{the bench is crafted from unvarnished timber, giving it a textured, coarse feel},'' rather than using color or contrast descriptions like ``\textit{an aisle carpeted in dark gray}.''
    \item \textbf{Late-blind VIPs} are a group of people who are blind after they have a basic understanding of our world; they know things similar to normal people. They prefer receive various information in the descriptions with as much visual detail as possible.
    \item \textbf{Low vision VIPs} are a group of people who have some degree of visual function and usually rely on combining their remaining vision and other senses to interact with their environment. Unlike those who are completely blind, individuals with low vision may use their visual capabilities to assist in the navigation and identification of objects, but they may have a loss of ability of Central vision, Peripheral vision, Depth perception, Contrast sensitivity, and Glare resistance.
\end{itemize}

Below are specialized tips for early-blind VIPs. Prioritize these over the initial 12 tips, as they align more closely with the unique experiences of early-blind VIPs. In cases of conflicting advice, please give these tips precedence:

\begin{applebox}{Tips}
    \begin{enumerate}[leftmargin=0.4cm]
        \item \textbf{Emphasize Non-Visual Senses}: Since early-blind individuals may have limited or no visual memories, focus on descriptions that leverage other senses like touch, sound, smell, and taste based on the visuals in image. For example, ``\textit{the carpet underfoot looks thick and soft, with a wavy texture; the walls seem to be smooth and cool to the touch}.''
        
        \item \textbf{Spatial Layout and Orientation}: Provide detailed information about the layout and orientation of spaces and objects, as this group relies heavily on spatial awareness for navigation and understanding their environment. Use language like ``\textit{walk straight for about ten steps to reach the sofa}.''

        \item \textbf{Texture and Shape Over Color}: Describe the texture and shape of objects, as these will be more meaningful than color descriptions, such as ``\textit{the shape of the object in front of you is similar to a large loaf of bread}.''

        \item \textbf{Consistent Terminology}: Use consistent terminology and reference points to avoid confusion, as early-blind individuals may have their unique ways of conceptualizing their environment. If you use steps as a measurement, as in ``\textit{the table is four steps away from you},'' continue using steps for other distances instead of suddenly switching to feet or meters. 

        \item \textbf{Contextual Descriptions}: Offer context for objects and environments that may not be immediately apparent, like the purpose of a certain object or the typical activities in a space. For example, ``\textit{you are now in an office, a place for work and study; the object with a flat top and drawers you can feel is a desk, used for writing and placing computers}.''

    \end{enumerate}
\end{applebox}

Below are specialized tips for low-vision VIPs. Prioritize these over the initial 12 tips, as they align more closely with their unique experiences. In cases of conflicting advice, please give these tips precedence:

\begin{applebox}{Tips}
    \begin{enumerate}[leftmargin=0.4cm]
        \item People with low vision often rely on color contrast to discern details. Describe the colors and the contrast between elements in the scene.
        
        \item Mention the scene's lighting as it can significantly affect how someone with low vision perceives the image.

        \item Provide context for the size of objects by comparing them to common items. For example, ``\textit{the dog is about the size of a bicycle}.''
    \end{enumerate}
\end{applebox} 

\subsubsection{Designs for GPT-4 Text Prompts: Improved Version} 
\label{subsubsec:design_prompt_improved}
I need three paragraphs of scene descriptions. 
\begin{itemize}
\item \textbf{The first description} should be concise and within 15 words.
\item \textbf{The second description} should be detailed and 150 words and generated for people with vision impairments (low vision or late blindness). Follow these tips:
\begin{applebox}{Tips}
    \begin{enumerate}[leftmargin=0.4cm]
        \item \textbf{Use clear and concise language:} Choose words carefully to provide clear, concise descriptions, using descriptive adjectives and adverbs for relevant information. 
        
        \item \textbf{Provide directional and distance information:} Include information about the layout of the space, who and where any people are, and what they are doing. Give a reliable description of how to navigate the space, describing any safety hazards in detail. Include information on the direction and distance of points of interest. Use clock face references (e.g., ``\textit{to your right at 3 o’clock}.'') to give a sense of orientation. Use common reference objects to describe distances, sizes, and other measurements.

        \item \textbf{Describe surroundings in stages:} Detail the environment in sections, starting with the immediate area, then describing nearby objects or obstacles, and finally providing information about the destination or the route ahead.
    \end{enumerate}
\end{applebox} 
\item \textbf{The third description} should also be detailed and 150 words, but generated for people who have been blind since birth. Follow the same rules as the second description, with one additional consideration:
\begin{applebox}{Tips}
    \begin{enumerate}[leftmargin=0.4cm]
        \item \textbf{Avoid describing color information:} Do not include color details that are difficult to imagine for a user who has been blind since birth.
    \end{enumerate}
\end{applebox} 
\end{itemize}

\subsubsection{Examples of GPT-4 Responses}
First, we show the image description results GPT generated through the first text prompt version. The first version of text prompts is shown in \S~\ref{subsubsec:design_prompt_first}.

Inspired by the work of Jain et al.\cite{jain2023want}, we design the first version of text prompts to achieve two main goals: 1) Generate three different concise versions of image descriptions for the three user categories. 2) Generate nine detailed image descriptions for the three different levels of needs for each user category. Our customized guidelines for detailed descriptions include but are not limited to, detailed object shapes and environmental safety factors. This approach ensures that each description level is finely tuned to meet the varied and specific needs of users based on their visual impairment conditions. We hope this customization addresses the unique challenges each group faces in navigating their environments effectively.

Secondly, we show the image description results generated by GPT through the improved version of text
prompts. The improved version of text
prompts is shown in \S~\ref{subsubsec:design_prompt_improved}.

By manually screening and scoring the results of the three levels of descriptions, our team unanimously found that, in the concise description, the descriptions for the three groups of users are very similar, containing basic scene layout information. In the detailed description, the descriptions at Level 2 (Independent exploration needs) are more accurate and contain the fewest errors, thus aligning more closely with the users' needs. Additionally, we observe that the descriptions for late-blind and low-vision users are remarkably similar. Considering the limitations of GPT and user needs, we make the following improvements in the improved version of the prompt: 1) Generate a common concise description for all VIP users; 2) Merge the detailed description needs of late-blind and low-vision users into one category; 3) Focus on the requirements of Level 2 and rewrite a simplified version of the GPT prompt to meet this requirement. The simpler and clearer the prompt language, the more accurate and readable the generated description sentences are.

\begin{applebox}{First Version of Text Prompts \& Response}
\begin{minipage}[t]{0.3\linewidth}
    \vspace*{0pt}
        \includegraphics[height=4.5\linewidth]{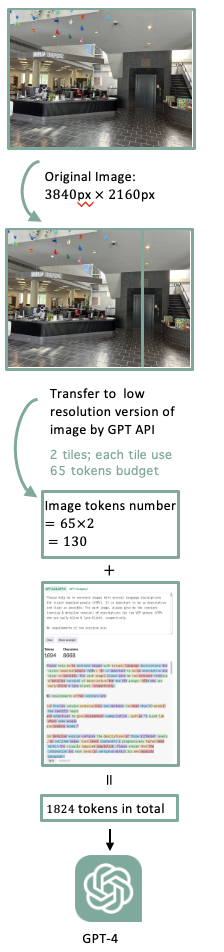}
\end{minipage}
\begin{minipage}[t]{0.67\linewidth}
\small

\colorbox{mypurple}{\textbf{For Low-Vision}} 
\begin{itemize}[leftmargin=*]
    \small
    \item \textcolor{mydarkpurple}{\textbf{Concise Version}}:
    \tiny
    \begin{itemize}[leftmargin=*]
        \item A modern help desk in a bright, spacious library.
    \end{itemize}
    \small
    \item \textcolor{mydarkpurple}{\textbf{Detailed Version}}: 
    \tiny
    \begin{itemize}[leftmargin=*]
        \item Level 1: The image shows a spacious library with a semi-circular help desk prominently in the foreground. The help desk is made of dark, smooth materials and is located centrally within the open area. To the right of the help desk, there is a set of elevators framed by dark tiles, and a staircase with a metal railing leading to an upper floor. The ceiling above the desk features hanging decorations in various colors, resembling paper cranes, adding a decorative touch to the space.
        \item Level 2: The help desk is positioned near the entrance, acting as a central hub for visitors. Behind the help desk are several computer workstations where people are seated, working. The flooring throughout the area consists of dark polished tiles, which could be slightly slippery. The elevators to the right are accessible by taking a few steps from the help desk, and the staircase next to them provides an alternative route to the upper floors. The area is well-lit with large windows that let in natural light, creating a bright environment.
        \item Level 3: For assistance, you can approach the staff at the help desk, who are seated behind the counter. They are available to provide directions or support with navigating the library. The elevators to the right are easy to locate and can be used to access different floors, with the staircase offering another option. The colorful paper cranes hanging above the help desk serve as a visual landmark, which can help in giving directions or coordinating with others in the library. The help desk is a good starting point for exploring the library, and the staff can guide you to specific sections or resources.
    \end{itemize}
\end{itemize}

\colorbox{mygreen}{\textbf{For Early-Blind}} 
\begin{itemize}[leftmargin=*]
    \small
    \item \textcolor{mydarkgreen}{\textbf{Concise Version}}:
    \tiny
    \begin{itemize}[leftmargin=*]
        \item A help desk in a spacious library with hanging decorations.
    \end{itemize}
    \small
    \item \textcolor{mydarkgreen}{\textbf{Detailed Version}}: 
    \tiny
    \begin{itemize}[leftmargin=*]
        \item Level 1: The image shows a help desk in a large, open library space. The help desk is semi-circular and made of smooth, hard materials. It is centrally located with workstations behind it and to the sides. Directly to the right of the help desk, there is a set of elevators, encased in a structure of smooth, hard tiles. To the far right, there is a staircase with a sturdy metal railing leading upwards.
        \item Level 2: The help desk is centrally placed for easy access. Behind the desk are several workstations, each with a computer and seating arrangements. The flooring throughout the area consists of polished tiles that feel smooth underfoot and may be slippery. The elevators to the right are accessible by walking a few steps from the desk. The staircase, located further right, has a metal handrail for support. There are tactile potted plants on the desk, offering a rougher texture compared to the smooth surfaces around.
        \item Level 3: For assistance, approach the help desk staff, who are positioned behind the counter. They can provide directions and support. When entering the library, you can ask for help with locating specific areas or using the workstations. The elevators are to the right and can be used to access different floors, with the staircase providing an alternative route. The hanging decorations above the desk, which have a smooth, paper-like texture, can serve as an auditory or tactile landmark when providing directions or coordinating with others in the library.
    \end{itemize}
\end{itemize}

\colorbox{myorange}{\textbf{For Late-Blind}}
\begin{itemize}[leftmargin=*]
    \small
    \item \textcolor{mydarkorange}{\textbf{Concise Version}}:
    \tiny
    \begin{itemize}[leftmargin=*]
        \item \small A help desk in a modern library with colorful paper cranes above.
    \end{itemize}
    \small
    \item \textcolor{mydarkorange}{\textbf{Detailed Version}}:
    \tiny
    \begin{itemize}[leftmargin=*]
        \item Level 1: The image depicts a help desk situated in the foreground of a modern library. The help desk is semi-circular, made of dark materials, and positioned centrally in the scene. To the right of the help desk, there's a set of elevators, framed in a dark tiled wall, with a staircase leading upwards on the far right. The ceiling above the help desk features a display of colorful paper cranes hanging from strings, adding a vibrant touch to the otherwise neutral-toned environment.
        \item Level 2: The help desk is positioned near the entrance, acting as a central point for visitors. Behind the help desk, there are several computer workstations occupied by people, and large windows allowing natural light to flood the space. The elevators to the right are accessible via a short walk from the help desk, and the staircase next to the elevators leads to the upper floors. The flooring is made of polished dark tiles, which may be slightly slippery. There are potted plants on the desk, contributing to a welcoming atmosphere.
        \item Level 3: If you need assistance, the help desk staff are seated behind the counter and are easily approachable. When entering, you can ask for directions to specific sections of the library or for help using the computers. The elevators to the right are useful for reaching different floors, and the staircase next to them is another option if you prefer walking. The colorful paper cranes hanging above can serve as a visual landmark when providing directions or coordinating with someone else in the library.
    \end{itemize}
\end{itemize}

\end{minipage}
\end{applebox}
\begin{applebox}{Improved Version of Text Prompts \& Response}
\begin{minipage}[t]{0.27\linewidth}
    \vspace*{1pt}
         \includegraphics[height=2.9\linewidth]
        {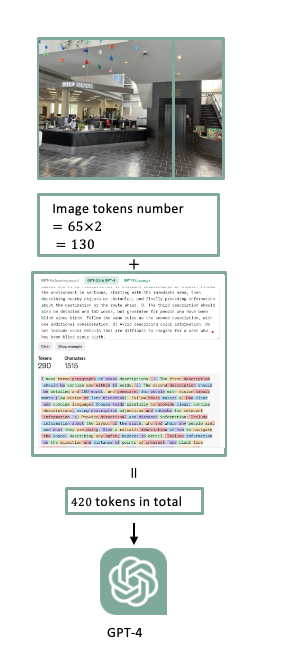}
\end{minipage}
\begin{minipage}[t]{0.7\linewidth}
\small
\textcolor{black}{\textbf{Concise Version}}:
\small
    \begin{itemize}[leftmargin=*]
        \item Help desk area with elevators, stairs, and hanging artistic paper cranes.
    \end{itemize}
\vspace{0.3cm}

\small   
\textcolor{black}{\textbf{Detailed Version}}:
\vspace{0.2cm}

\small
\colorbox{mygreen}{\textbf{For Early-Blind}} 
\small
\begin{itemize}[leftmargin=*]
        \item You are standing in a large, open lobby. Directly ahead, approximately 10 feet in front of you, is a curved help desk with a person available to assist. Moving to the right, at 3 o'clock, are sets of elevators within a tiled encasement. Next to the elevators, stairs begin about 20 feet away, going up to the next floor. Behind the help desk, at your 9 o'clock position, is a spacious area filled with tables, where several people are using computers. This section spans about 30 feet from where you are. The ceiling features artistic hanging objects above the help desk. The floor consists of solid tiles, ensuring an even surface without visible obstacles between you and the help desk.
    \end{itemize}
    
\small
\colorbox{myorange}{\textbf{For Low-Vision \& Late-Blind}} 
\small
\begin{itemize}[leftmargin=*]
        \item You are in a spacious lobby area. Immediately in front of you, around 10 feet away, is a curved help desk, manned by a person. To your right, at 3 o'clock, are elevators encased in a dark tiled structure. A set of stairs is also to the right of the elevators, leading upwards. The stairs begin around 20 feet away from you. Directly behind the help desk, to the left at 9 o'clock, is a large open area with multiple tables and people seated working on computers. This area extends about 30 feet from your current position. Above the help desk, look up around 45 degrees to see colorful paper cranes hanging from the ceiling. The floor is covered with dark tiles, providing a stable and even surface. There are no apparent obstacles between you and the help desk.
    \end{itemize}
\end{minipage}
\end{applebox}

\subsection{Visual Prompts for Correcting Directional Errors}
We find that GPT's ability to count and recognize direction is weak. For example, it sometimes cannot accurately describe the number of doors and tables. The left and right orientations cannot be accurately described based on the direction of the user's line of sight (camera direction). Taking into account spatial orientation information is crucial for BLV users. In addition to text prompts, we introduce vision prompt information, which means embedding left and right orientation information in each picture and instructing GPT to use the orientation information embedded in the picture to describe the object's relative position. Experiments show that the proposed orientation-related vision prompt method can effectively solve the problem of orientation errors in descriptions.


\subsubsection{Examples of GPT-4 Responses}
First, we present one example of image description generated by GPT using only the improved text prompt. We observed that GPT struggles to accurately describe the location of objects on the left and right sides. For instance, GPT's description states that the lockers are on the right side of the image and the glass wall is on the left side, when the opposite is true.

Next, we show the description results generated by combining the visual and text prompts. This corrected the inaccurate direction descriptions.

\begin{applebox}{Improved Version of Text Prompts Only \& Response}
\begin{minipage}[t]{0.27\linewidth}
    \vspace*{1pt}
         \includegraphics[height=2.9\linewidth]
        {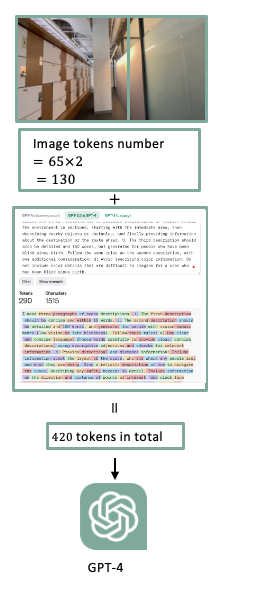}
\end{minipage}
\begin{minipage}[t]{0.7\linewidth}
\small
\textcolor{black}{\textbf{Concise Version}}:
\small
    \begin{itemize}[leftmargin=*]
        \item Modern office hallway with wooden lockers and glass walls.
    \end{itemize}
\vspace{0.3cm}

\small   
\textcolor{black}{\textbf{Detailed Version}}:
\vspace{0.2cm}

\small
\colorbox{mygreen}{\textbf{For Early-Blind}} 
\small
\begin{itemize}[leftmargin=*]
        \item This scene is set in a modern office environment, characterized by a straight, extended hallway about 20 feet in length. Along the right side of the hallway, there is a row of lockers, each equipped with metallic handles positioned at waist height to facilitate easy opening. These lockers are uniformly spaced and extend the full length of the hallway. On the opposite side, a smooth, hard barrier runs parallel to the lockers, indicating a separation from another area. The floor beneath is flat and made from a hard material, contributing to the echo of footsteps, which can help in gauging the hallway’s length and your proximity to the walls. Overhead, consistent intervals of sound suggest the presence of lighting fixtures. The hallway is about five feet wide, offering clear passage. At the corridor's end, there is a turn to the left, though this is beyond the immediate vicinity. The space is notably quiet, hinting at either a time of low activity or effective soundproofing.
    \end{itemize}
    
\small
\colorbox{myorange}{\textbf{For Low-Vision \& Late-Blind}} 
\small
\begin{itemize}[leftmargin=*]
        \item You are in a contemporary office hallway that extends straight ahead, about 20 feet long. To your right, a row of tall, wooden lockers lines the wall, featuring metallic handles at waist height for easy access. Directly opposite these lockers, to your left, is a long, smooth glass wall, beyond which lies another office space, although specifics are obscured by the wall’s frosted lower half. The ceiling overhead is equipped with evenly spaced strip lights, providing bright illumination that reflects off the polished concrete floor beneath your feet. The hallway is wide, approximately five feet across, allowing ample room for navigation without obstacles. At the end of this hallway, the path turns left, though this is just out of immediate reach. The atmosphere is quiet, with minimal noise, suggesting a less busy time of day or a well-insulated space.
    \end{itemize}
\end{minipage}
\end{applebox}

\begin{applebox}{Combined Version of Visual and Improved Text Prompts \& Response}
\begin{minipage}[t]{0.27\linewidth}
    \vspace*{1pt}
         \includegraphics[height=2.9\linewidth]
        {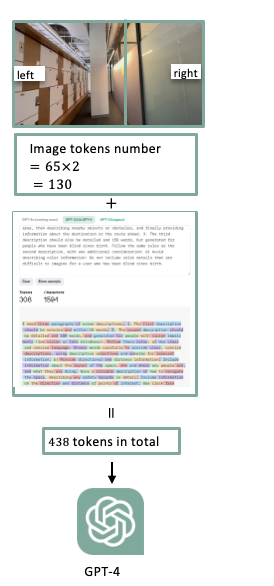}
\end{minipage}
\begin{minipage}[t]{0.7\linewidth}
\small
\textcolor{black}{\textbf{Concise Version}}:
\small
    \begin{itemize}[leftmargin=*]
        \item Hallway with lockers on left and office partitions on the right.
    \end{itemize}
\vspace{0.3cm}

\small   
\textcolor{black}{\textbf{Detailed Version}}:
\vspace{0.2cm}

\small
\colorbox{mygreen}{\textbf{For Early-Blind}} 
\small
\begin{itemize}[leftmargin=*]
        \item You are in a hallway. To your left at 9 o'clock there is a series of lockers mounted on a wooden base. These lockers are square and follow a grid pattern along the wall. On your right at 3 o'clock, there is an office partition wall with panes that give a semi-transparent appearance. The floor you are standing on is carpeted, and the area around is well illuminated. In front and slightly to the center at about 12:30, there is a structural pillar emerging from the floor and extending towards the ceiling. The hallway extends further ahead past the pillar. The ceiling has visible pipes and ductwork arranged neatly. Ensure to navigate around the pillar as you proceed forward. The route seems clear and straight ahead, eventually leading towards a brighter area that suggests an external light source from a window.
    \end{itemize}
    
\small
\colorbox{myorange}{\textbf{For Low-Vision \& Late-Blind}} 
\small
\begin{itemize}[leftmargin=*]
        \item You are in a hallway with square lockers to your left at 9 o'clock. These lockers are mounted on a wooden base and appear to be organized in a grid pattern. To your right at 3 o'clock, there is an office partition wall, with frosted glass panels near the top. The floor is carpeted, and the hallway appears to be well lit. Directly ahead, there is a supporting pillar in the center of the hallway. Beyond it, the hallway continues straight ahead. The ceiling above has exposed pipes and ducts arranged in an orderly manner. There are no immediate hazards, but be mindful of the pillar directly ahead. The space seems to lead into an area further on with visible light from a distant window.
    \end{itemize}
\end{minipage}
\end{applebox}

\section{Experimental Settings and Additional Results}
\subsection{Learning settings}
For the learning settings, we employed the Adam optimizer. The configurations of optimizer and other hyper parameters in our experiments are shown in Table~\ref{tab:learning_settings}.
\begin{table}[]
    \centering
    \caption{Learning Settings across all model trainings.}
    \label{tab:learning_settings}
    \scalebox{1.2}{
        \begin{tabular}{cc}
        \toprule
            Parameter & Value \\
        \midrule
            Intial sx & 0.0 \\
            Intial sq & -5.0 \\
            Intial Learning Rate & 0.0001 \\
            Optimizer & Adam \\
            Weight Decay & 0.0005 \\
            Gamma & 0.1 \\
            Scheduler & StepLR \\
            Step Size & 50 \\
            Epoch & 200 \\
            Batch Size & 16 \\
            Pre-trained Weights & ImageNet1K \\
        \bottomrule
        \end{tabular}
    }
\end{table}

\subsection{Visualizations}
\begin{figure}[ht]
    \centering
    \begin{subfigure}[b]{0.45\linewidth}
        \includegraphics[width=\linewidth]{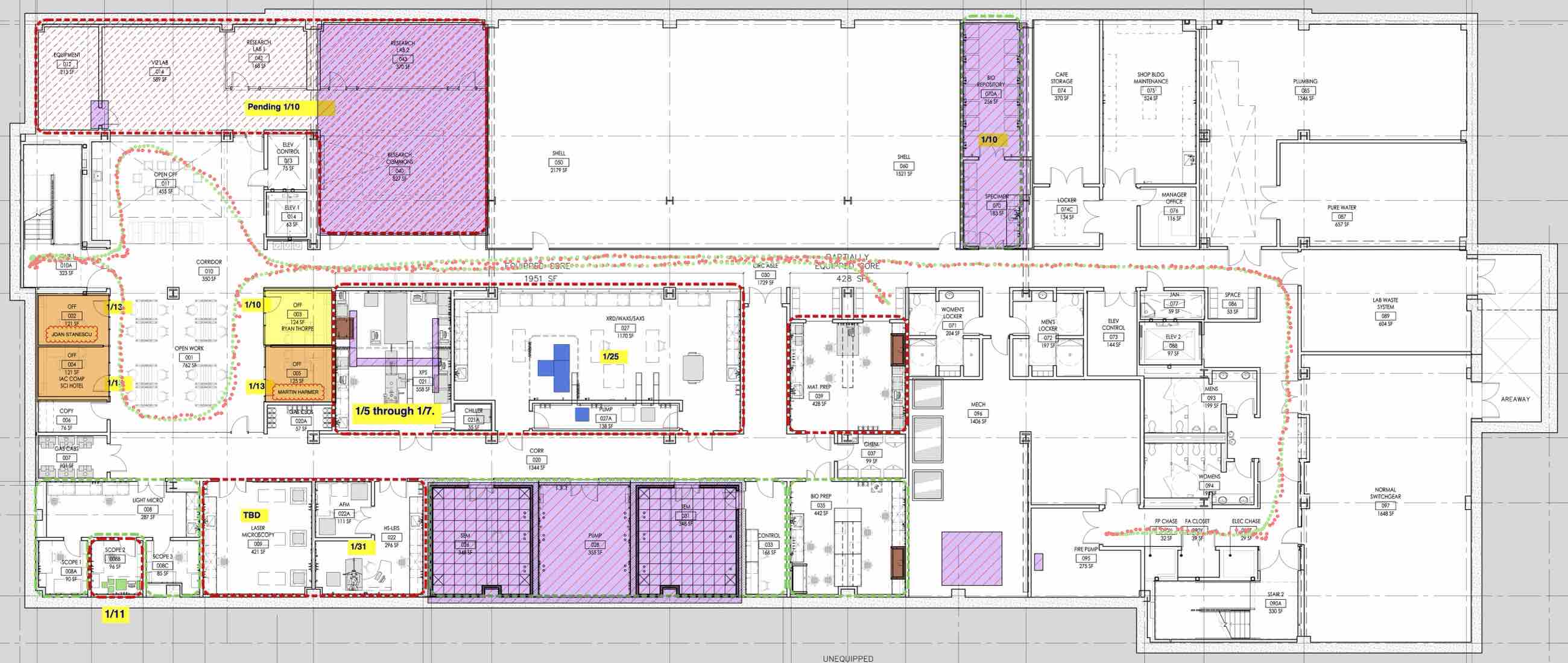}
        \caption{Basement: 2024/04/08-11:34:37}
    \end{subfigure}\quad
    \begin{subfigure}[b]{0.45\linewidth}
        \includegraphics[width=\linewidth]{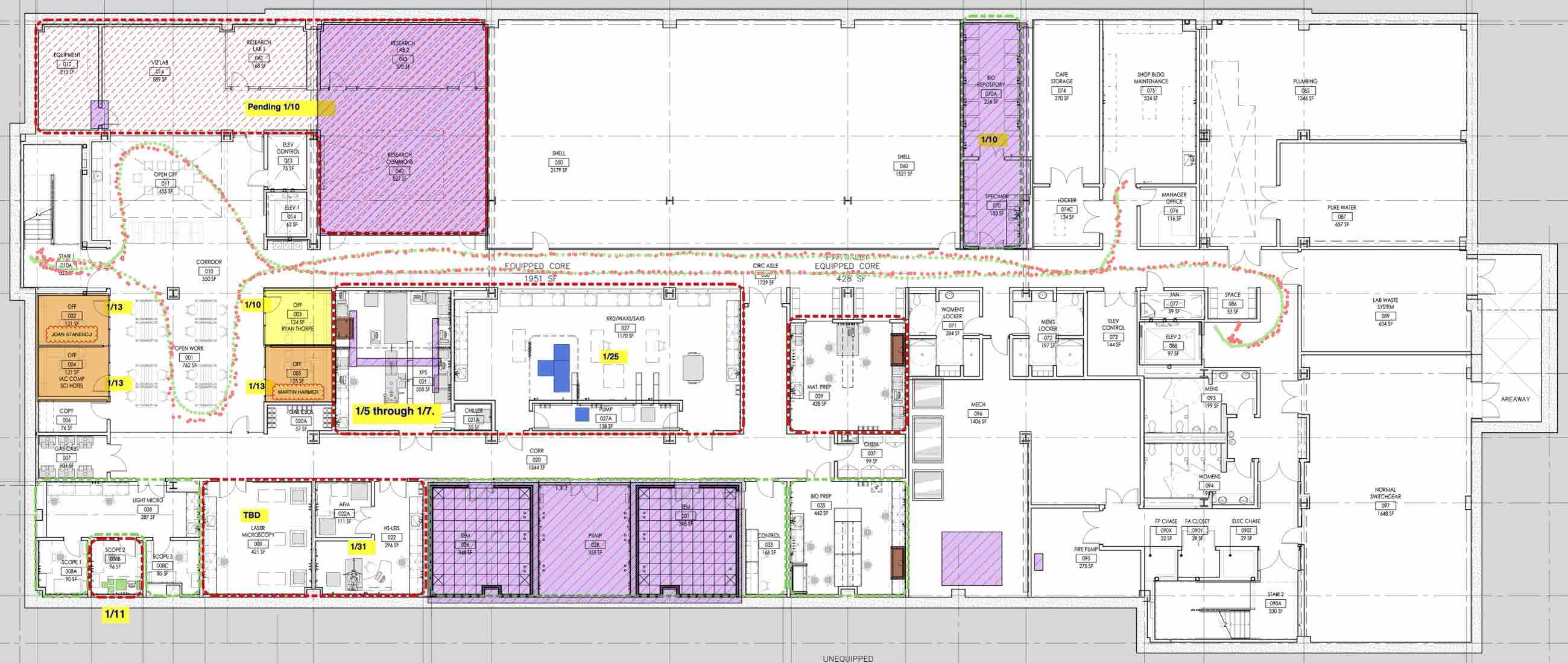}
        \caption{Basement: 2024/04/08-11:40:19}
    \end{subfigure} \\
    \begin{subfigure}[b]{0.45\linewidth}
        \includegraphics[width=\linewidth]{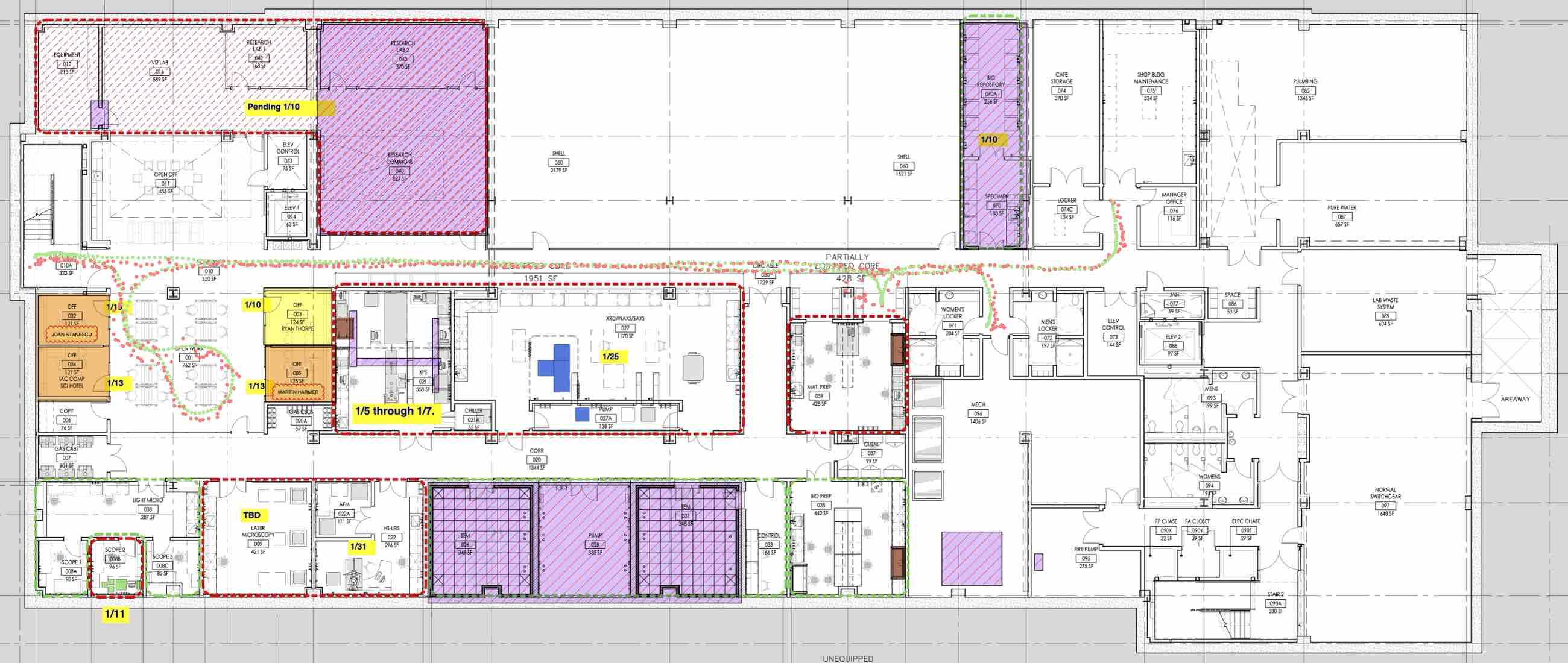}
        \caption{Basement: 2024/04/16-23:59:26}
    \end{subfigure} \quad
    \begin{subfigure}[b]{0.45\linewidth}
        \includegraphics[width=\linewidth]{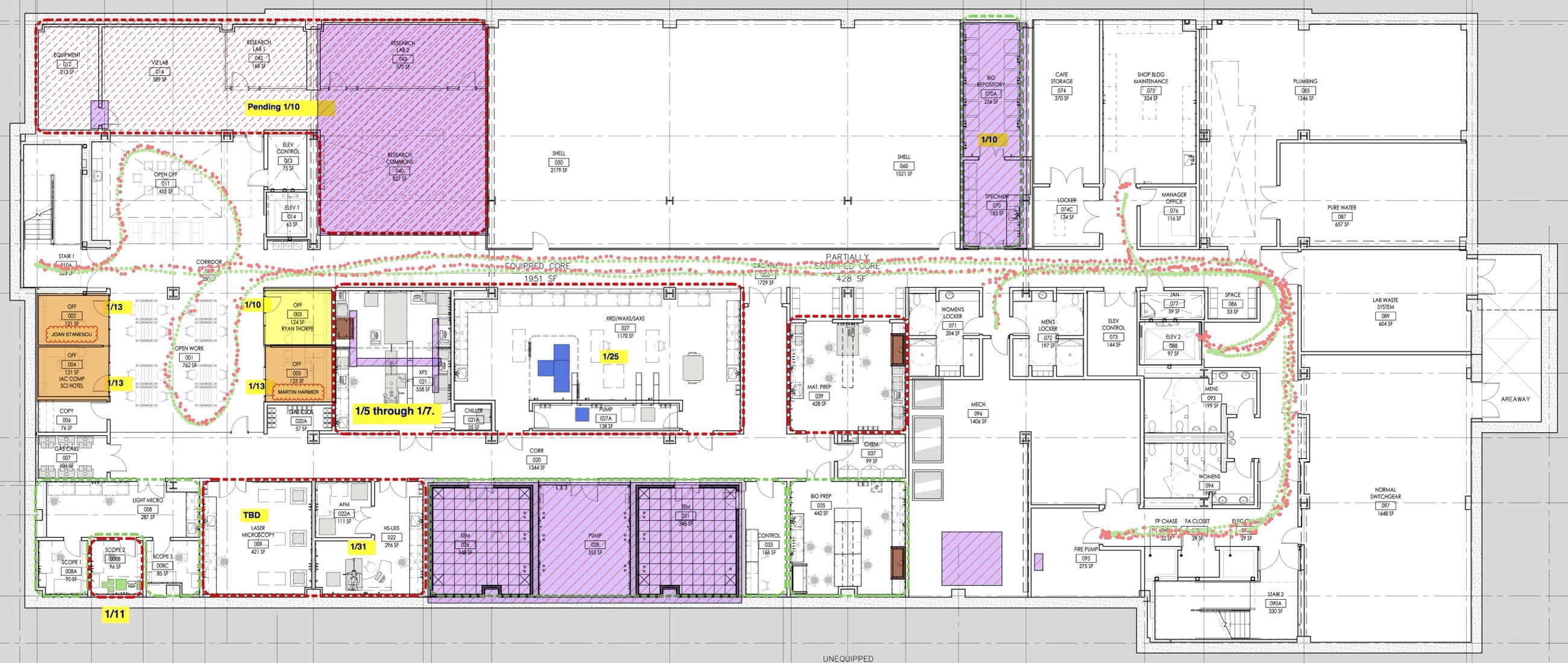}
        \caption{Basement: 2024/05/14-08:51:38}
    \end{subfigure}   
    
    \begin{subfigure}[b]{0.45\linewidth}
        \includegraphics[width=\linewidth]{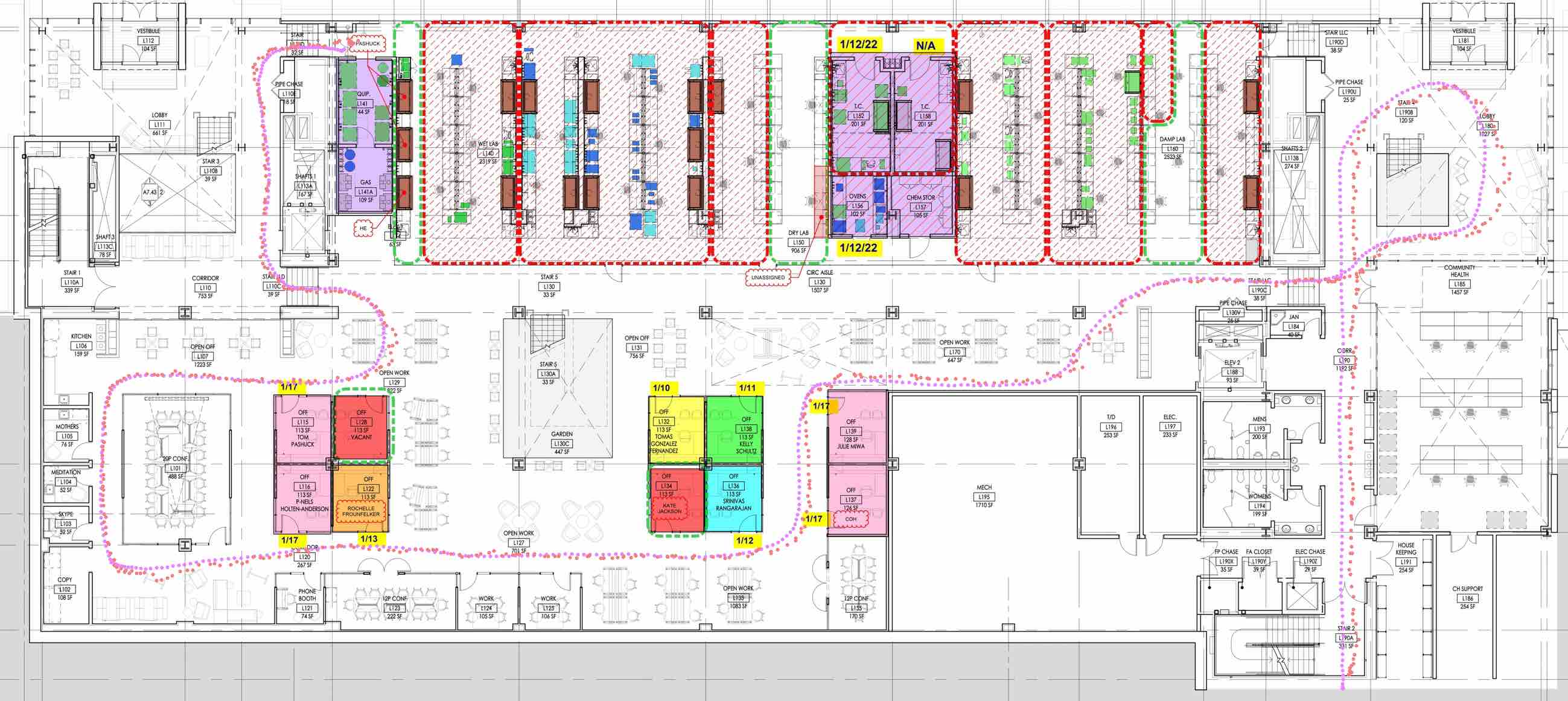}
        \caption{Lower Level: 2024/04/08-11:44:36}
    \end{subfigure}\quad
    \begin{subfigure}[b]{0.45\linewidth}
        \includegraphics[width=\linewidth]{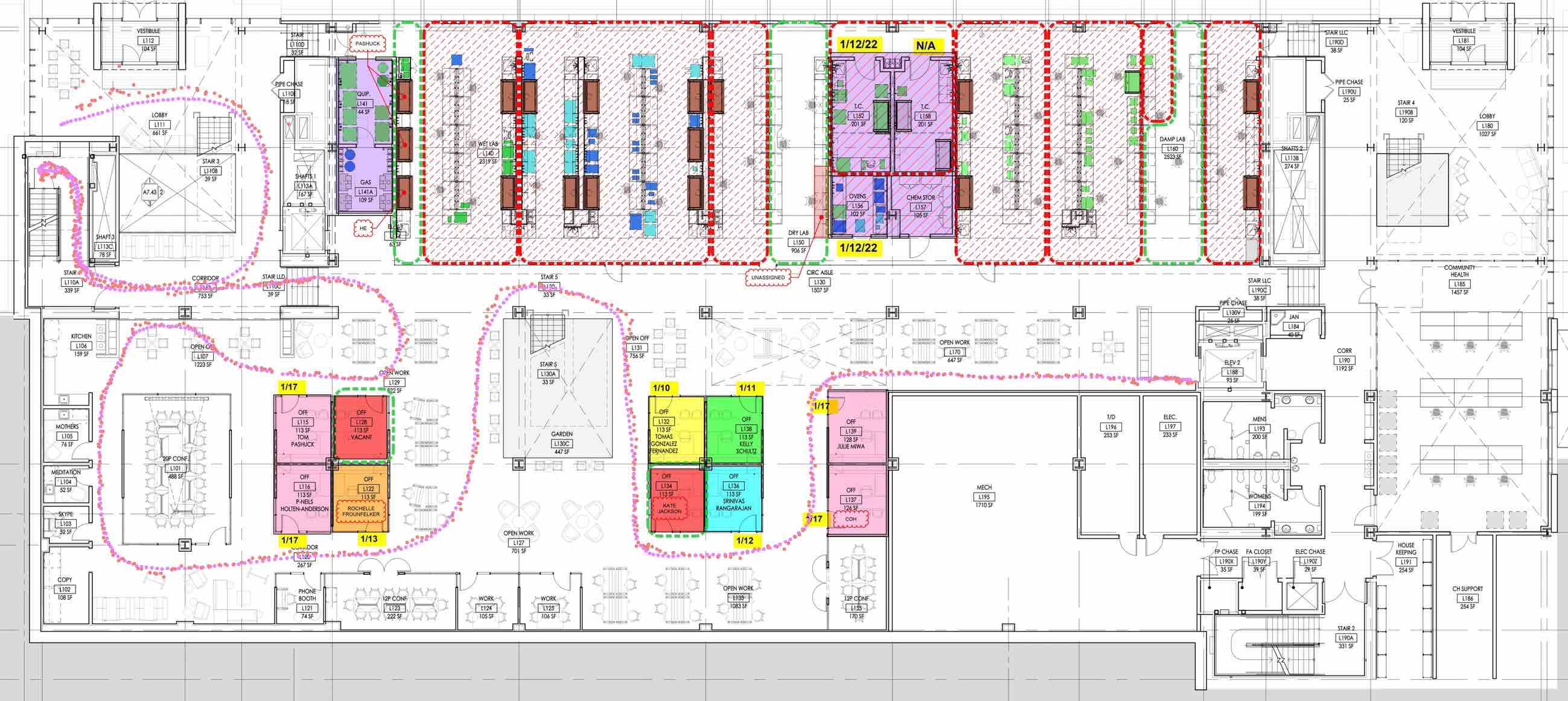}
        \caption{Lower Level: 2024/04/15-11:49:29}
    \end{subfigure} \\
    \begin{subfigure}[b]{0.45\linewidth}
        \includegraphics[width=\linewidth]{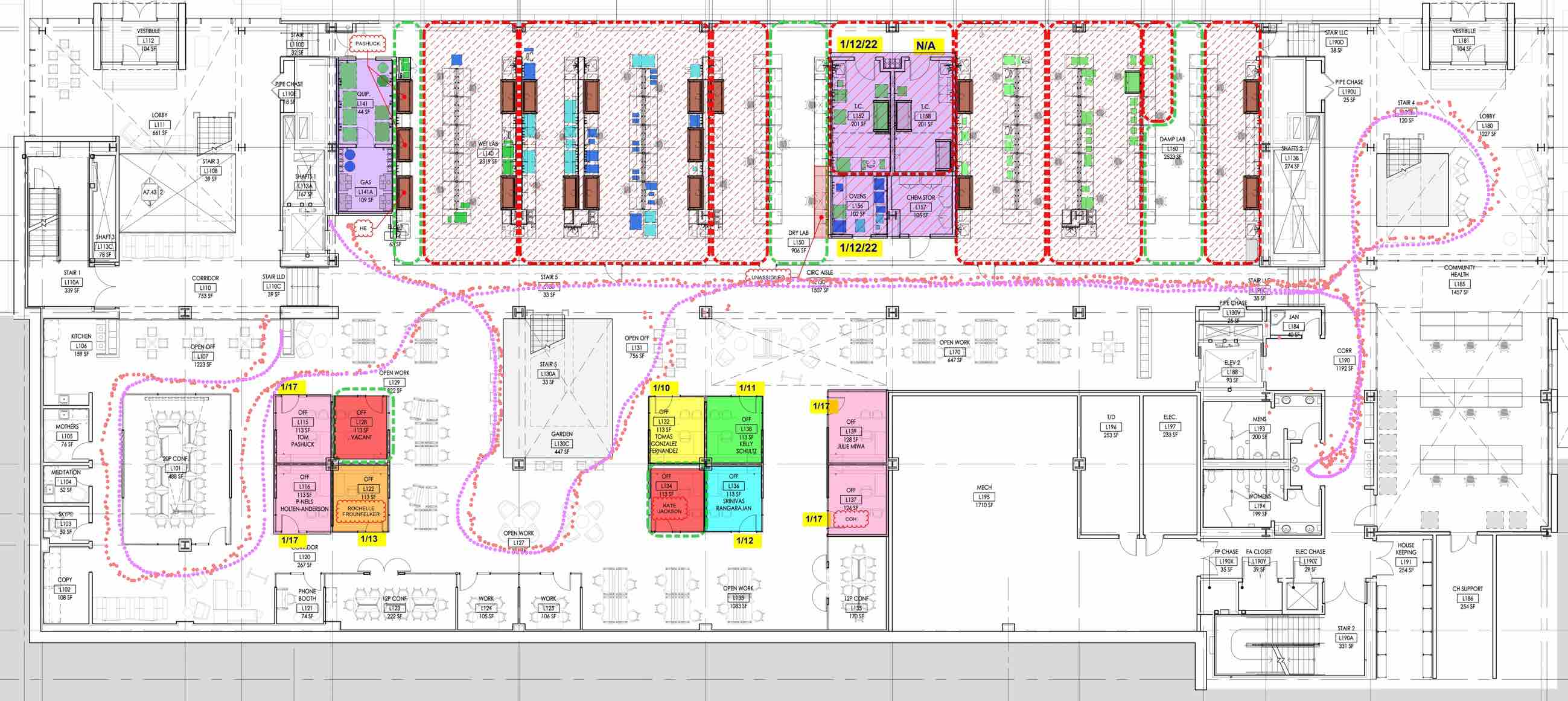}
        \caption{Lower Level: 2024/04/17-00:12:46}
    \end{subfigure} \quad
    \begin{subfigure}[b]{0.45\linewidth}
        \includegraphics[width=\linewidth]{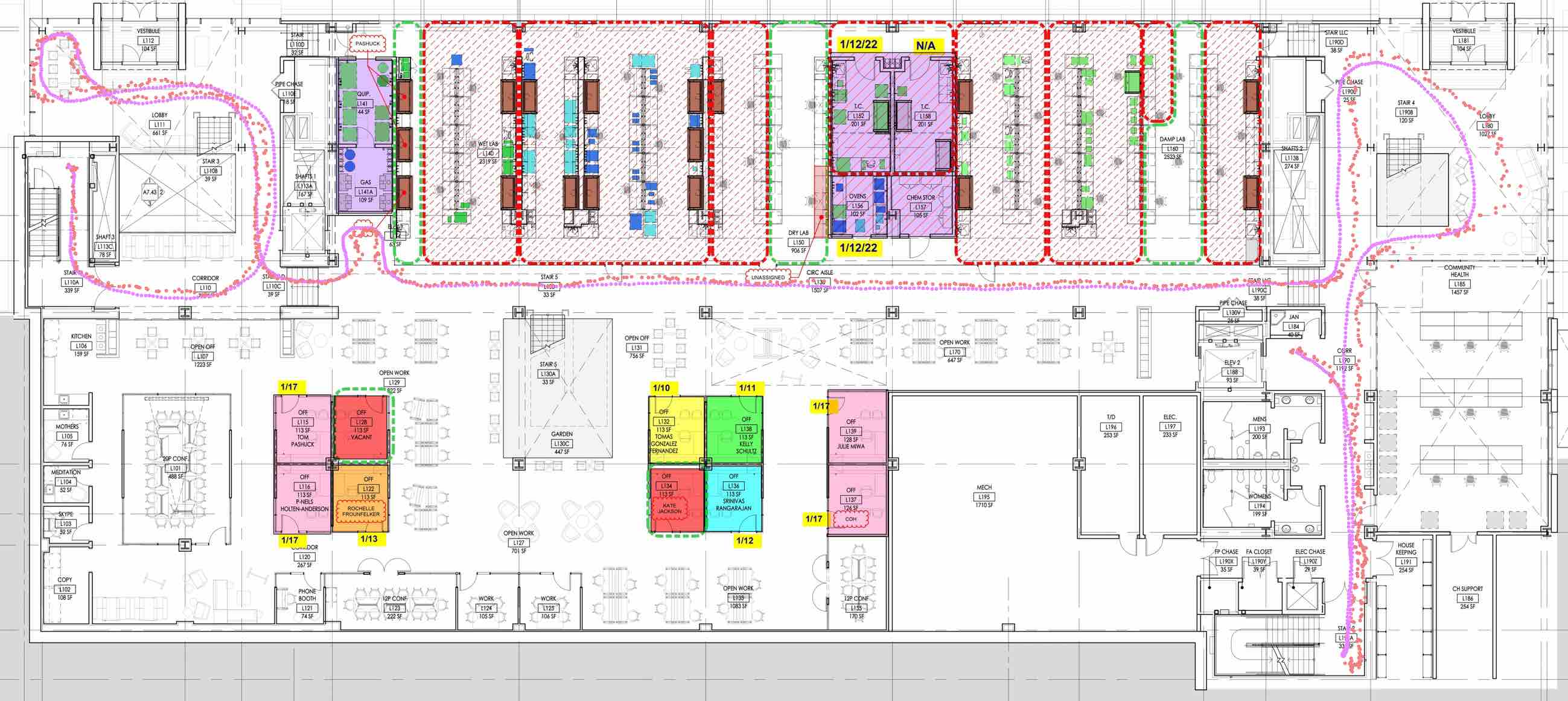}
        \caption{Lower Level: 2024/05/14-07:49:16}
    \end{subfigure}
    
    \begin{subfigure}[b]{0.45\linewidth}
        \includegraphics[width=\linewidth]{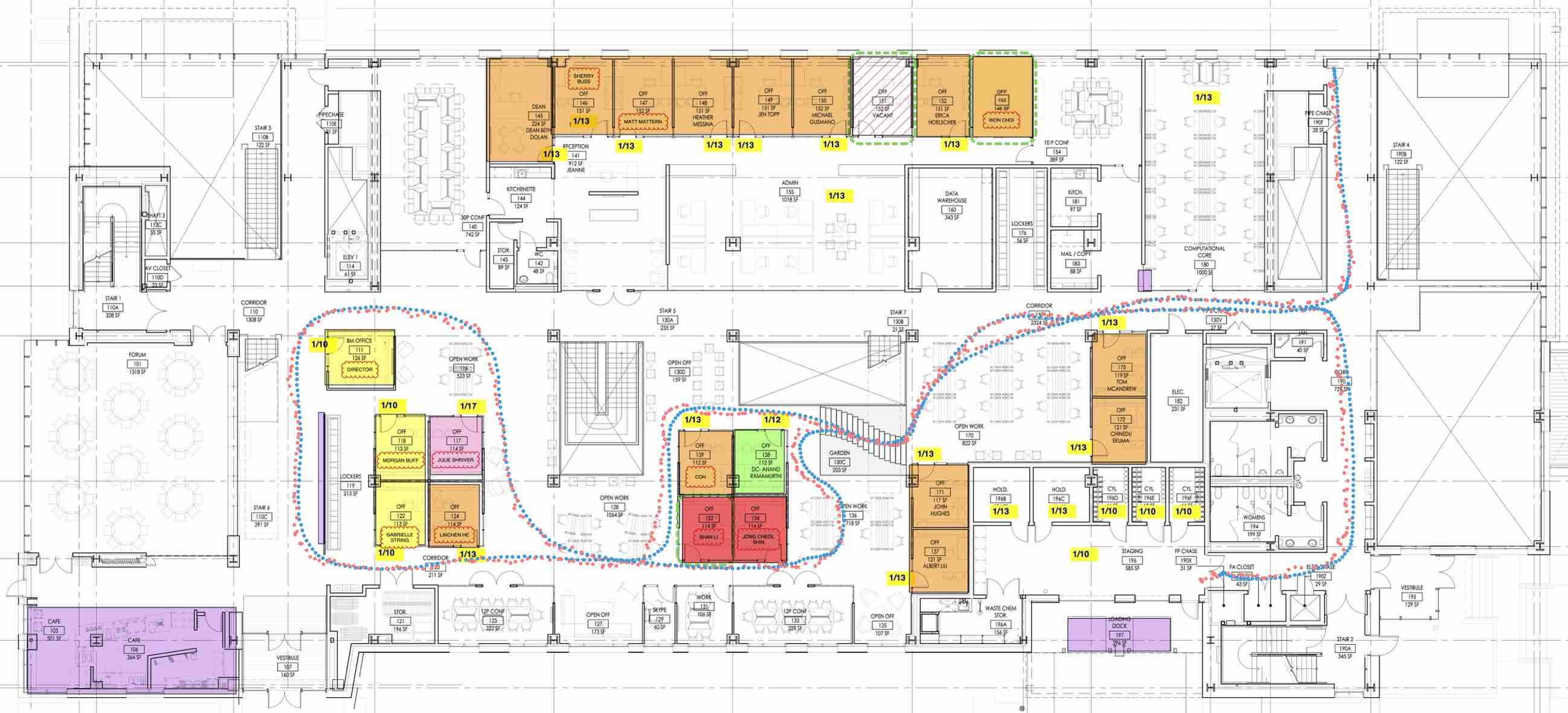}
        \caption{Level 1: 2024/04/08-11:55:52}
    \end{subfigure}\quad
    \begin{subfigure}[b]{0.45\linewidth}
        \includegraphics[width=\linewidth]{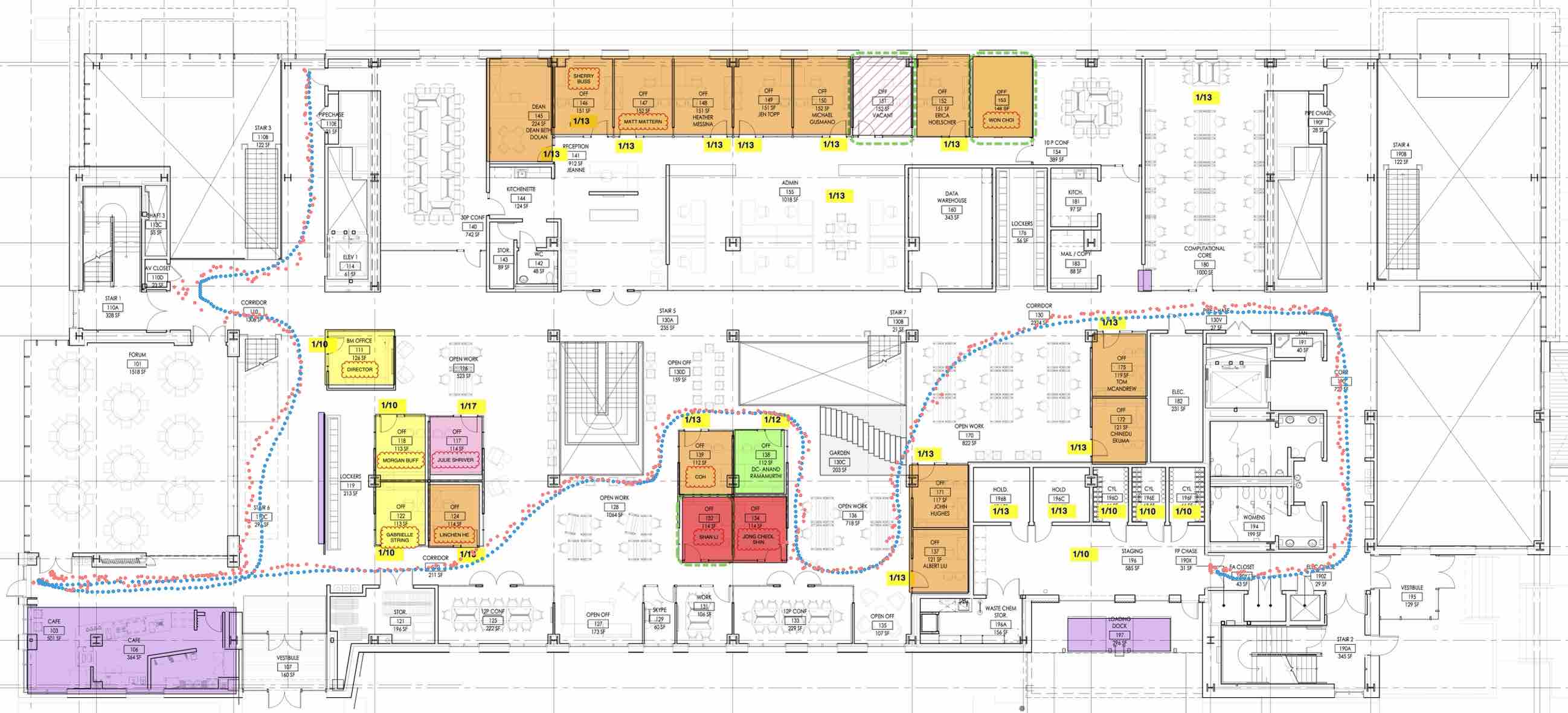}
        \caption{Level 1: 2024/04/17-00:17:14}
    \end{subfigure} \\
    \begin{subfigure}[b]{0.45\linewidth}
        \includegraphics[width=\linewidth]{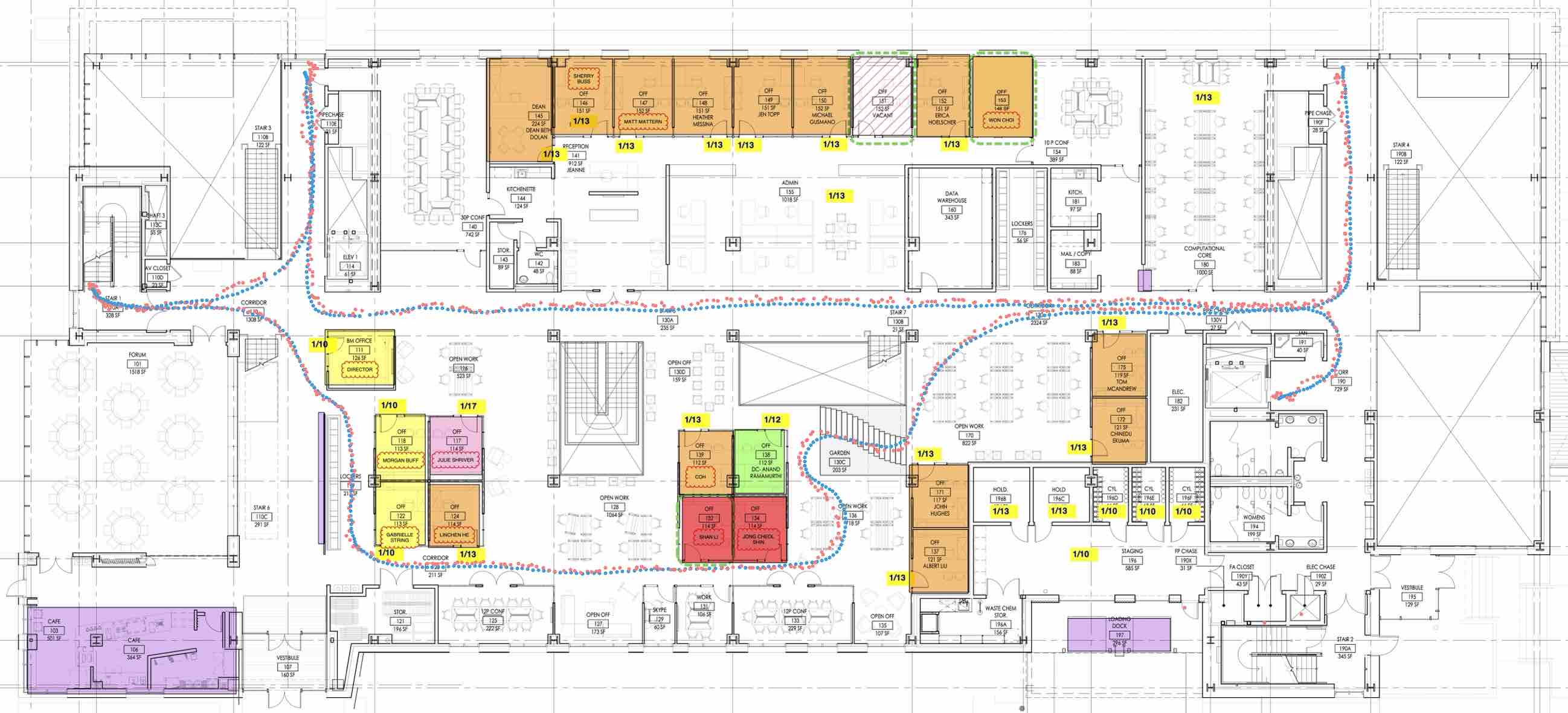}
        \caption{Level 1: 2024/04/24-21:24:18}
    \end{subfigure} \quad
    \begin{subfigure}[b]{0.45\linewidth}
        \includegraphics[width=\linewidth]{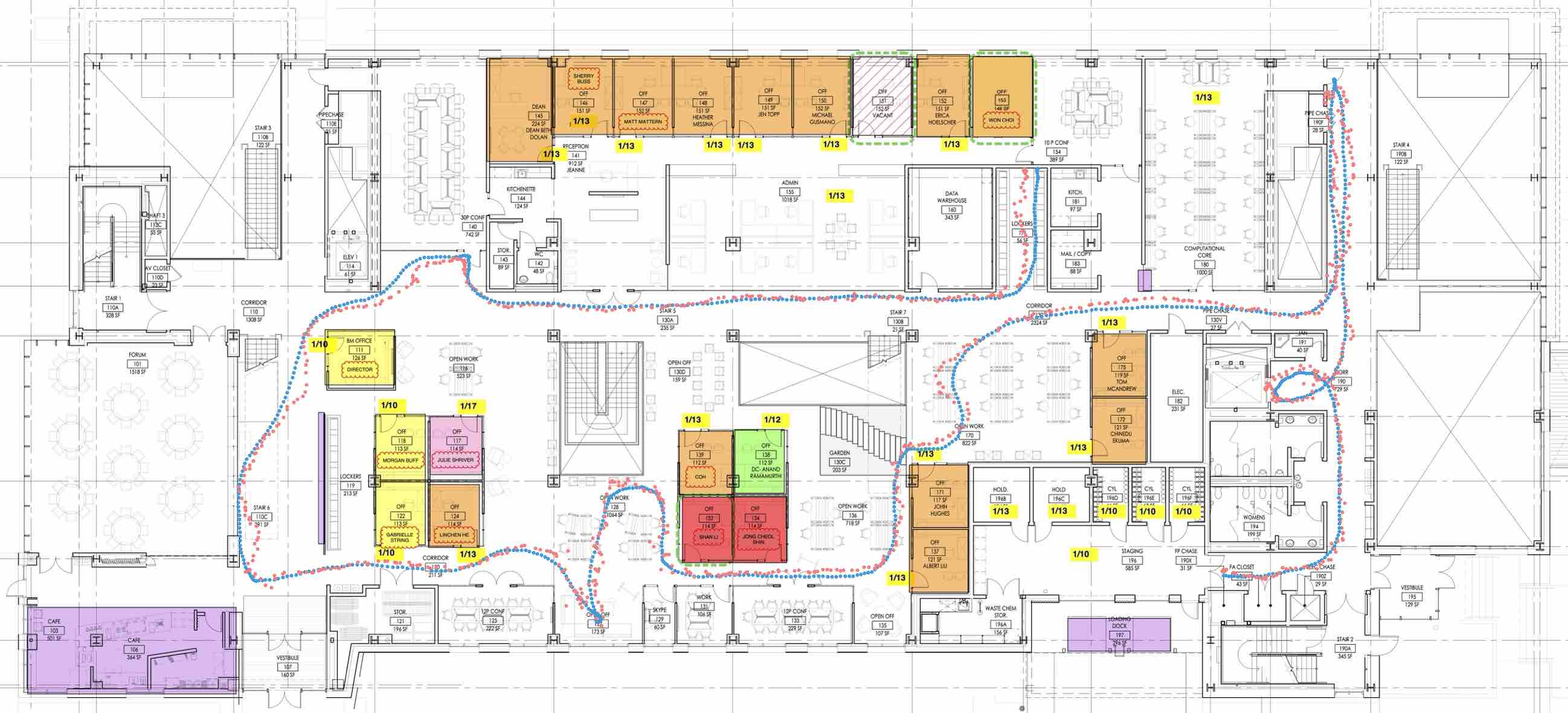}
        \caption{Level 1: 2024/05/14-07:58:49}
    \end{subfigure}

\end{figure}
\clearpage
\begin{figure}[h]\ContinuedFloat
    \centering
    \begin{subfigure}[b]{0.45\linewidth}
        \includegraphics[width=\linewidth]{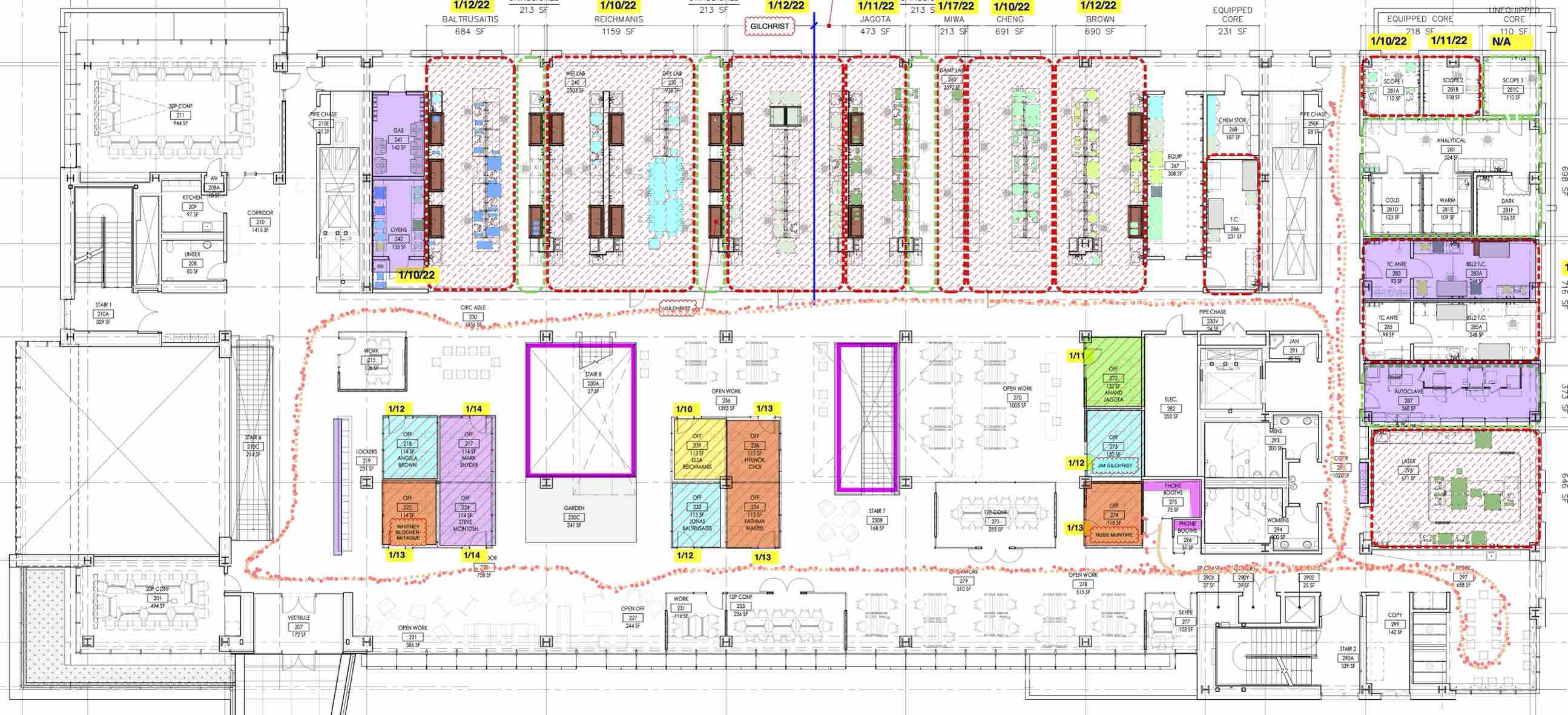}
        \caption{Level 2: 2024/04/08-12:03:48}
    \end{subfigure}\quad
    \begin{subfigure}[b]{0.45\linewidth}
        \includegraphics[width=\linewidth]{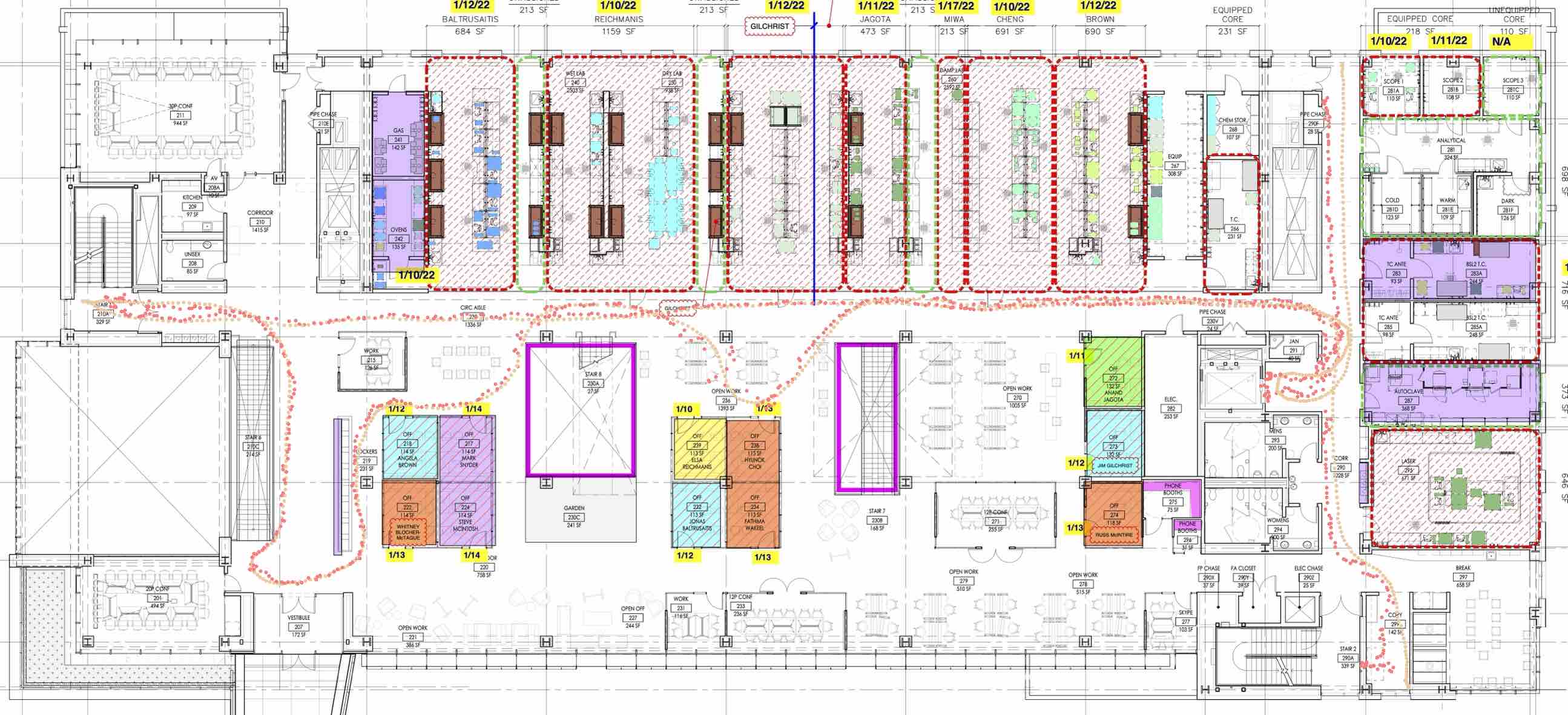}
        \caption{Level 2: 2024/04/18-17:29:21}
    \end{subfigure} \\
    \begin{subfigure}[b]{0.45\linewidth}
        \includegraphics[width=\linewidth]{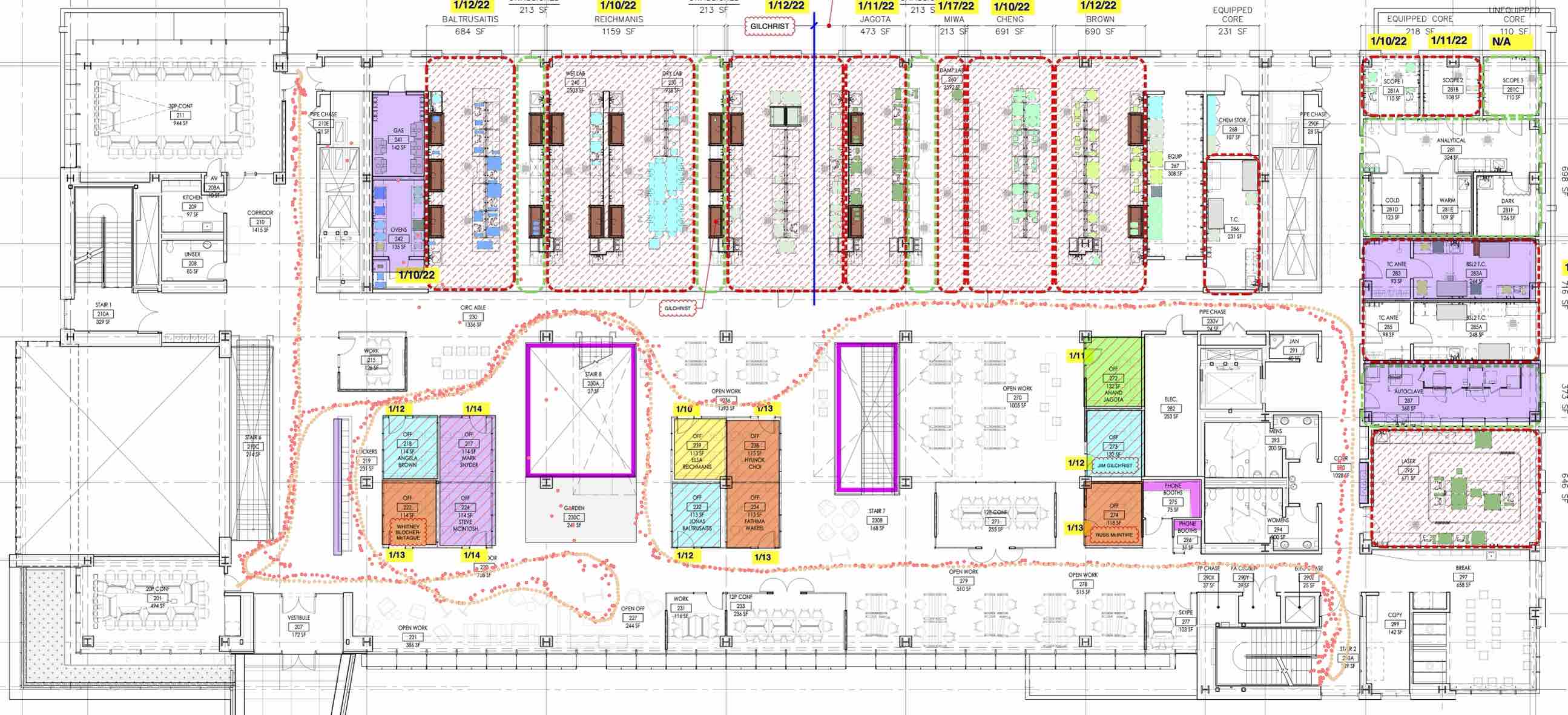}
        \caption{Level 2: 2024/05/14-09:16:56}
    \end{subfigure} \quad
    \begin{subfigure}[b]{0.45\linewidth}
        \includegraphics[width=\linewidth]{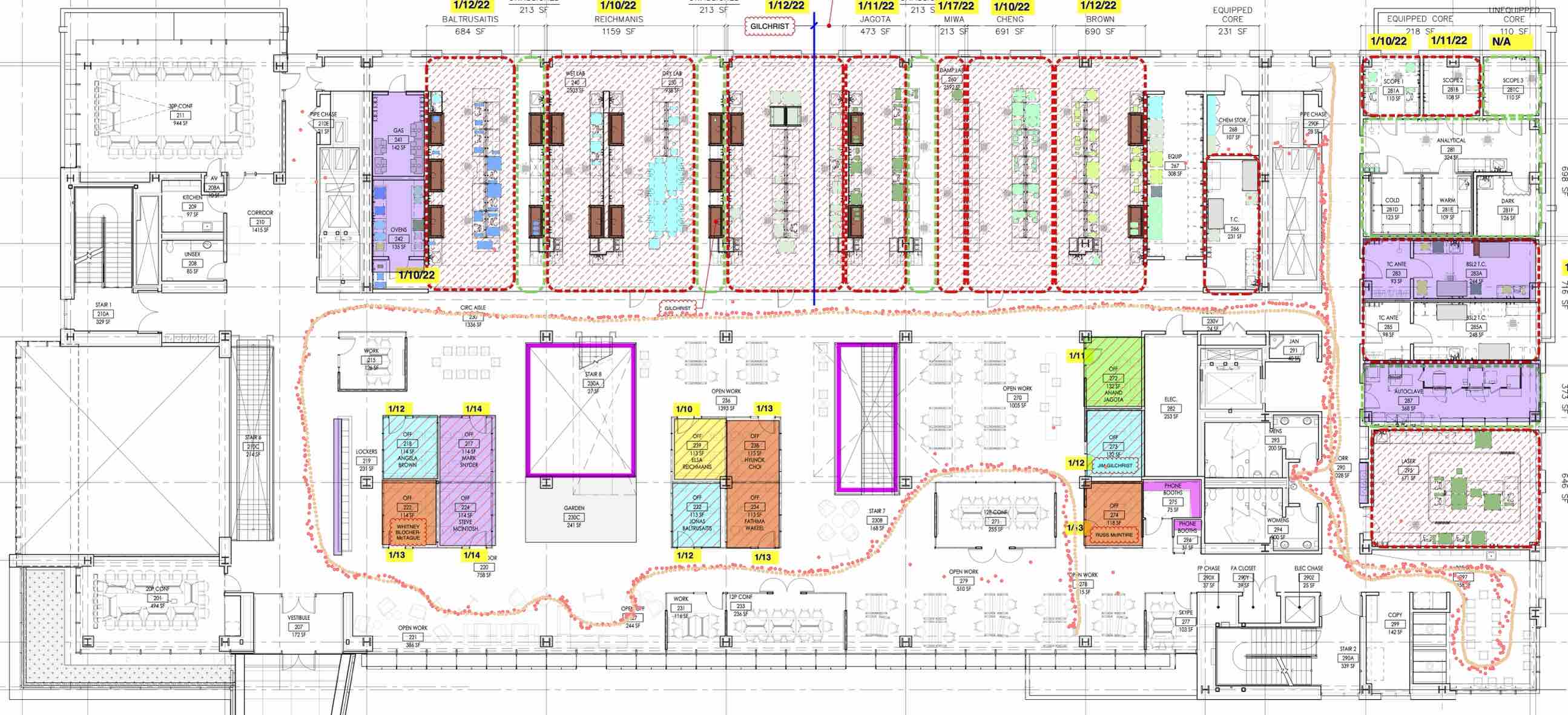}
        \caption{Level 2: 2024/05/29-18:55:34}
    \end{subfigure}
    \caption{We randomly selected four videos from each floor in the test dataset to visualize the predictions of camera poses. Predictions are marked with red dots, while ground truths for each floor are color-coded as follows: Basement in green, Lower Level in purple, Level 1 in blue, and Level 2 in orange.}
    \label{fig:pred_path}
\end{figure}


\end{document}